\documentclass[lettersize,journal]{IEEEtran}

\input{header.sty}

\begin{document}

\title{VibrantLeaves: A principled parametric image generator for training deep restoration models}
% \title{Bridging the gap between synthetic and natural images in deep image restoration}

\author{Raphaël Achddou$^1$, Yann Gousseau$^2$, Saïd Ladjal$^2$, Sabine Süsstrunk$^1$~\IEEEmembership{Fellow,~IEEE,}
        % <-this % stops a space
\thanks{$^1$ School of Computer and Communication Sciences,
EPFL, 1015 Lausanne, Switzerland (e-mail: raphael.achddou@epfl.ch).}
\thanks{$^2$ LTCI, Télécom Paris Institut Polytechnique de Paris, 19 Place Marguerite Perey 91200 Palaiseau.}
% <-this % stops a space
}

\markboth{Journal of \LaTeX\ Class Files,~Vol.~14, No.~8, August~2021}%
{Shell \MakeLowercase{\textit{et al.}}: A Sample Article Using IEEEtran.cls for IEEE Journals}


\maketitle
\begin{bibunit}[IEEEtran]
\begin{abstract}
In this paper, we introduce a synthetic image generator relying on a few simple principles, specifically focusing on \textit{geometric modeling, textures, and a simple modeling of image acquisition}. These principles, integrated into the classical Dead Leaves model, allow for the creation of high-quality training sets for image restoration tasks. Standard image denoising and super-resolution networks trained on these datasets achieve performance comparable to those trained on natural image datasets.
The motivation behind this approach stems from the limitations of Deep Neural Networks in image restoration tasks. Despite their impressive performance, these networks are often poorly understood and prone to biases inherited from standard natural image training sets. To mitigate these issues, we emphasize the need for a better control over training sets, particularly through the use of synthetic and abstract datasets.
Furthermore, our work includes a detailed analysis of the principles considered, identifying which image properties are necessary for maintaining high performance, thus taking a first step towards explainability. 
Besides, we show that neural networks trained with our synthetic sets naturally inherit robustness to the invariance properties of these sets, namely rotation and scale.

\end{abstract}


\section{Introduction}

% \begin{figure}[tp]
%     \centering
% \begin{tabular}{c|c}
% Dead Leaves\cite{achddou2021synthetic} & \textbf{VibrantLeaves (Ours)}\\
% \hline
% \includegraphics[width=0.4\linewidth]{images/dead_leaves_first_figure/im_69809436.png} &\includegraphics[width=0.4\linewidth]{images/dead_leaves_first_figure/im_first_page.png} \\
% \includegraphics[width=0.4\linewidth]{images/dead_leaves_first_figure/im_69808226.png} & \includegraphics[width=0.4\linewidth]{images/dead_leaves_first_figure/im_first_page_2.png}\\
% \end{tabular}
%     \caption{(Left) Images sampled from the original dead leaves image model used in\cite{achddou2021synthetic}. (Right) Images sampled from our improved dead leaves model \DLplus, which contains several additions regarding the modeling of geometry, textures, physical depth, and self-similarity, each presented in Section~\ref{sec:advances}.
%     These images are samples from the training set used for training image denoising NNs, which results are shown in Figure~\ref{fig:teaser}}\label{fig:dl_1stpage}
% \end{figure}

%Image restoration problems, which are famously ill-posed, are a central topic in image processing.
Advancements in image restoration techniques have always been closely tied to the development of image models, due to the inherently ill-posed nature of image restoration problems.
% Statistical image models is at the 
% Image restoration problems  been a fruitful source of advances in image modeling, in particular because of their ill-posedness.  
Researchers first derived mathematical priors, translated into either regularization-based optimization algorithms~\cite{wiener1949extrapolation,DavidL.Donoho1994,rudin1992nonlinear} or direct processing methods~\cite{buades2005non,Dabov2007a}.
With the advent of deep learning, these classical methods were largely outperformed by deep neural networks (NN).
These networks are trained on large datasets of clean and corrupted image pairs by minimizing a loss function, typically the Mean-Square Error~\cite{zhang2018ffdnet,zhang2021plug}.
Although these methods have significantly improved the performances on most image restoration tasks, they suffer from several limitations.
% First, the inner workings of image restoration NNs are difficult to understand. The architecture and the datasets 
Neural networks are indeed notoriously bad at generalizing to unseen modalities, be it changes of the image domain or of the type of distortions \cite{plotz2017benchmarking}. 
Moreover, restoration neural networks tend to hallucinate details that are not present in the original image \cite{goujon2024learning,el2022bigprior}. Indeed, training image datasets are strongly biased, which may be reflected in restoration results or pose privacy issues.
Overall, the combination of large network capacity and biased training datasets makes these models difficult to interpret and prone to reproducing training artefacts.

\begin{figure*}
    \centering
    \includegraphics[width=\linewidth]{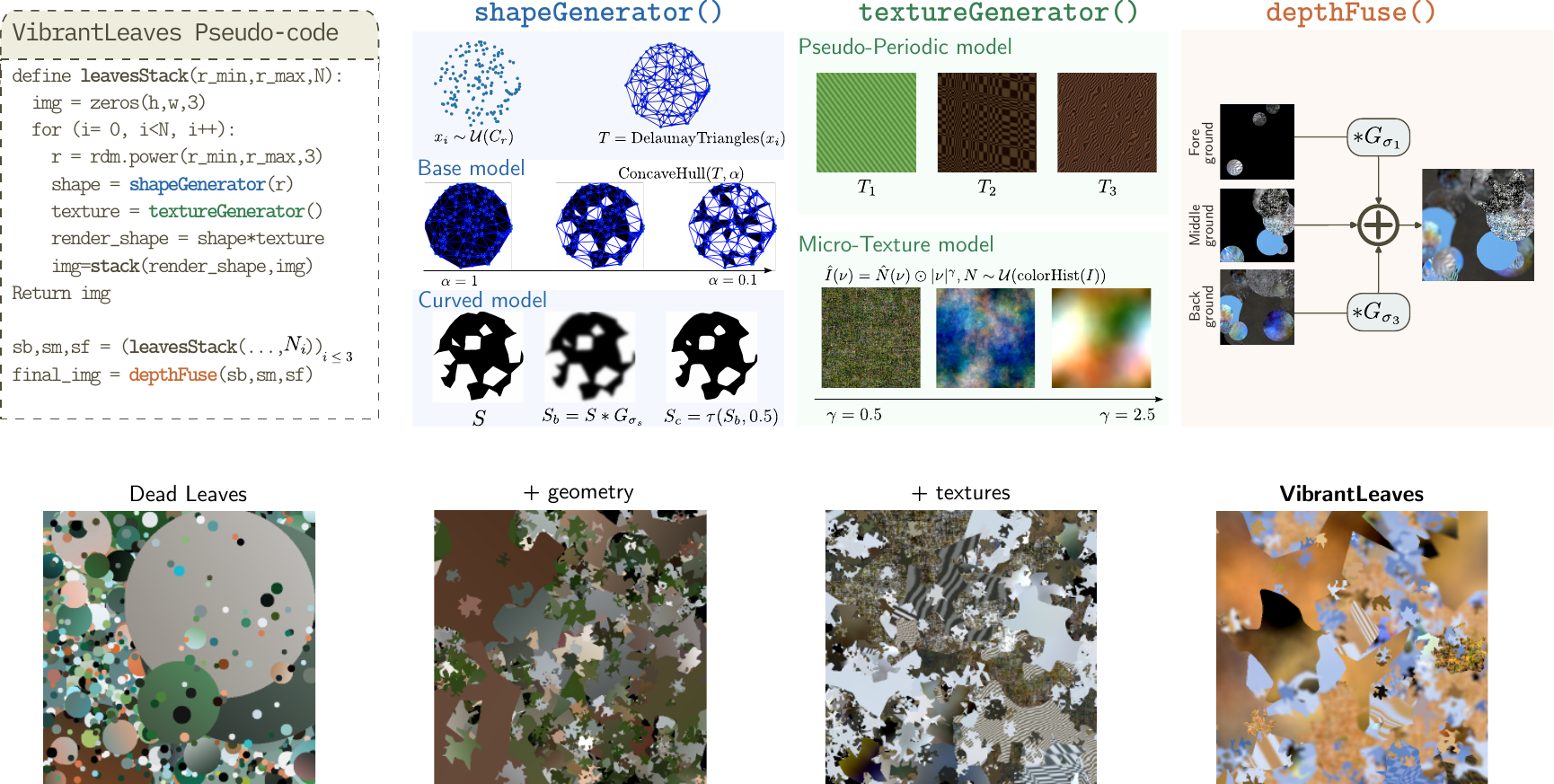}
    \caption{Graphical summary of the VibrantLeaves model. We integrate three key properties in an occlusion-based model: (1) \textbf{Geometry}, through a random shape generator based on concave hulls, (2) \textbf{Textures} with an exemplar-free parametric texture generator of both pseudo-periodic and micro textures, (3) \textbf{Depth} with a depth-of-field generator and perspective effects. The generated images match natural images statistics much better than other statistical image models, while remaining abstract and free of real image dataset biases.}
    \label{fig:teaser}
\end{figure*}

One approach toward interpretability is to replace standard training datasets with synthetic images generated by a compact parametric model.
This approach was first proposed in~\cite{achddou2021synthetic} and extended in~\cite{achddou_cviu}, where training images are synthesized from a \textit{Dead Leaves} (DL) image model.
Such images, first introduced to model porous media~\cite{matheron1968modele}, have since been proposed as an effective statistical model for natural images, closely matching several of their statistical properties~\cite{alvarez1999size,lee2001occlusion}.
Some of these properties, such as the Laplacian distribution of the image gradient, are also enforced by classical regularization priors.
%Here, they are directly encoded in the training images.
By training NNs with such images, the corresponding statistical properties are implicitly encoded in the trained network via the training sets. Even though this approach showed promising results, the performance gap with networks trained on natural images remained significant. As shown in Figure~\ref{fig:teaser}, images sampled from the classical DL model lack key properties of natural images, such as realistic textures, complex object shapes or depth-related visual cues.

In this paper, we propose a new principled image generator, called \textbf{\DLplus}, which incorporates crucial properties of natural images: (1) \textit{Geometry}: we generate complex shapes capable of reproducing arbitrary local curvature; (2) \textit{Textures}: we propose a simple parametric, exemplar-free texture generator producing pseudo-periodic and stochastic textures; (3) \textit{Depth}: we integrate a depth-of-field simulator along with perspective tilting to reproduce important perceptual depth cues. The code is available on \href{https://github.com/rachddou/DeadLeavesPlus}{Github} for reproducibility.

The corresponding model significantly increases the statistical diversity and visual richness of the generated images, while still relying on a small set of parameters.
These improvements lead to a significant increase in performance of image restoration networks trained on \DLplus~images, achieving performance nearly on par with networks trained on natural images, with only an average $0.72$\,dB gap in PSNR for image denoising. We further confirm these gains through super-resolution experiments, as well as deblurring and inpainting experiments in a Plug-and-Play setting~\cite{venkatakrishnan2013plug}.
%Indeed, out-performing standard baselines on traditional benchmarks datasets, which match with the training data distribution perfectly, seems unrealistic.
%Nonetheless, 
Moreover, the modular framework of \DLplus~sheds light on the image properties that drive restoration performance, contributing to more interpretable image restoration models.
We also show that networks trained on \DLplus~images naturally inherit robustness to the invariance properties of the generating model.

In short, we can summarize our contribution as follows:
\begin{enumerate}
    \item We propose \DLplus~(VL), a principled image generator based on the Dead Leaves model, which incorporates essential properties of natural images, such as \textit{geometry, textures, and depth}.
    \item We show that standard image denoising and super-resolution NNs trained on VL~images achieve performance almost on par with NNs trained on natural images. Furthermore, using the PnP setup, we apply it to a wide range of linear inverse problems, including inpainting and deblurring.
    %, with only a $0.7$dB gap in PSNR.
    \item As a step toward interpretability, we provide a careful analysis of the image properties required to reach good image restoration performance, through ablation studies.
    \item We
    %analyze the invariances of image restoration models, showing
    show that networks trained with VL images naturally inherit robustness to the rotation and scale invariance properties of the generating model.
\end{enumerate}
% \begin{itemize}
%     \item \DLplus: an improved version of Dead Leaves focusing on essential properties of natural images, matching natural images statistics and geometry, 
%     \item synthetically trained neural networks for both Image Denoising and Single-Image Super-Resolution with only a $0.5\sim 0.7$dB gap in PSNR with models trained on natural images,
%     \item A study of the invariances of image restoration models, showing that training with \DLplus~images leads to better robustness,
%     \item A careful analysis of the image properties required to reach good image restoration performance.
% \end{itemize}

\section{Related Works}
\subsection{Image restoration: prior-based vs learning based approaches}
The goal of image restoration is to retrieve an estimate of a ground-truth image $x$ from altered measurements $y$, related by:
\begin{align}
    y = \Phi(x)+ \eta
    \label{eq:ip}
\end{align}
where $\Phi$, called the forward operator, models the acquisition system and $\eta$ is typically additive white Gaussian noise.
Retrieving $\hat{x}$ is an ill-posed problem. In the Bayesian framework, the Maximum-a-Posteriori (MAP) estimator is:
\begin{align}
    \hat{x}_{MAP}   &= \arg \max_x p(x|y) \\
    &= \arg \min_x \left[ \frac{(\Phi(x)-y)^2}{2\sigma^2} -\log(p(x))  \right] \\
    &= \arg \min_x f(x) + g(x)
    \label{eq:MAP}
\end{align}
where $\sigma$ is the noise standard deviation, $f$ is the data fidelity term, and $g$ is the prior term — unknown in practice.
To derive tractable algorithms from this formulation, researchers developed analytical priors imposing regularity such as smoothness \cite{wiener1949extrapolation,chu1998edge} or sparsity \cite{donoho1998minimax,rudin1992nonlinear,Yu2011}. Non-local methods \cite{Dabov2007a,lebrun2013nonlocal,buades2005non} similarly exploit self-similarity as a regularity principle, though they can't be expressed in this formalism.

The focus then shifted to deep learning, which surpassed prior-based approaches on all benchmarks \cite{zhang2021plug,zhang2018ffdnet}, and continues to improve with increasingly complex architectures such as Transformers \cite{zamir2022restormer} and Diffusion models \cite{saharia2022image}.
Deep learning methods learn a direct mapping minimizing the MSE on pairs of clean and corrupted images, making the prior implicit in the network parameters and hard to interpret. One can view this implicit prior $\rho_{NN}(x)$ as follows: 
\begin{align}
    \rho_{NN}(x) \simeq \mathcal{F}(\text{dataset} \times \text{architecture} \times \text{loss function},x )
\end{align}
where $\mathcal{F}$ entangles dataset, architecture, and loss function — hence the ``black-box'' qualifier. Despite this opacity, such implicit priors can be used in Plug-and-Play \cite{venkatakrishnan2013plug} or stochastic solvers \cite{kadkhodaie2021stochastic}.

A recent trend consists in injecting prior knowledge through architecture design: MWCNN \cite{liu2018multi} operates in the wavelet domain for sparsity, and non-local NNs exploit self-similarity in the feature space \cite{cruz2018nonlocality,valsesia2020deep}. Additional perceptual loss functions can further constrain the learned mapping \cite{johnson2016perceptual,sajjadi2017enhancenet}. However, little effort has been devoted to controlling the properties of the training data itself for image restoration tasks.

\subsection{Synthetic images for training deep learning models}

Using synthetic data to train NNs is standard practice across many fields, driven by the difficulty and cost of acquiring sufficiently diverse real-world data.

The importance of realism depends on the task. For low-level vision, abstract datasets suffice: Flying Chairs and AutoFlow achieve strong performance for optical flow estimation despite depicting only flying objects on plain backgrounds \cite{dosovitskiy2015flownet,sun2021autoflow}. For high-level tasks — semantic segmentation \cite{chen2019learning}, instance segmentation \cite{ward2018deep}, object detection \cite{hinterstoisser2019annotation}, and classification \cite{frid2018synthetic} — photo-realism and semantic content are more critical. Datasets for these tasks are drawn from rendering engines \cite{ros2016synthia,gan2020threedworld,greff2022kubric,zheng2023pointodyssey}, video games \cite{richter2016playing}, or generative models \cite{zhang2021datasetgan,shmelkov2018good,sariyildiz2023fake}, whose primary goal is to produce annotated data at scale. However, such generators are either highly engineered or highly parametrized, and the implicit prior learned from them remains hard to interpret and prone to bias.

To address this, a recent trend consists in training NNs on abstract images drawn from simple random processes. FractalDB \cite{kataoka2020pre} pretrains classification NNs on binary images of random fractals, later extended to Transformers \cite{takashima2023visual} and segmentation \cite{shinoda2023segrcdb}; however, it lacks key natural image properties such as color, background, and occlusion \cite{anderson2022improving}. Achddou et al. \cite{achddou2021synthetic} proposed training image restoration NNs on Dead Leaves images, which match the color histogram, Fourier spectrum, and gradient statistics of natural images \cite{ruderman1997origins,lee2001occlusion} (see Section~\ref{sec:background} for background). In parallel, Baradad et al. \cite{baradad2021learning} benchmarked synthetic datasets for image classification, finding textured dead leaves and constrained Style-GAN images most effective. Madhusudana \cite{madhusudana2021revisiting} further augments texture realism by sampling from real-world texture datasets, at the cost of introducing recognition bias.

% , \cleardoublepage, \pagebreak, 
\section{
%Incorporation of natural image properties in the DL model
The \DLplus~model}\label{sec:advances}
In this section, we introduce \DLplus, a new synthetic model for training image restoration networks, incorporating simple geometric modeling, textures and depth clues. 
The proposed algorithm  is built upon the Dead Leaves (DL) model, which is extensively described in \cite{gousseau2007modeling}. An introduction to the model is provided in the supplementary material.

%In all that follows, some key components of the original method remain the same:
%\begin{itemize}
 %   \item \textbf{Color Sampling:} for a single VL image, colors are sampled in the histogram of a single randomly picked natural image from Waterloo dB. This procedure, justified in \cite{achddou2021synthetic}, leads to better color rendering than a uniform sampling of the RGB cube. Note that sampling the global color histogram of a large database leads to poor results, as shown in \cite{achddou2023learning}, as it destroys local color correlations.
  %  \item \textbf{Size Distribution:} the object's size distribution is still a power law  $f(r)=C*\mathbf{1}_{[r_{\min},r_{\min}]}(r)*r^{-\alpha}$, with $\alpha = 3$ to maintain scale invariance up to the minimal radius,
   % \item \textbf{Post-Processing:} we keep a down-scaling operator, in order to remove aliasing caused by binary shapes.
%\end{itemize} 
\subsection{Geometry}\label{sec:geometry}
A limitation of the synthetic dataset proposed in \cite{achddou_cviu} is that dead leaves images are obtained using only disks. 
While disks are easily generated and guarantee rotation invariance, they are insufficiently expressive to capture the shape diversity of natural and man-made objects.
Common natural structures, such as straight boundaries, corners, holes, or arbitrary local curvatures, cannot arise from disk superimposition.
% A simple way to solve that would be to include rectangular shapes in the DL model, potentially augmented with rotations, to maintain rotation invariance.
% However, only adding rectangles will not solve the diversity issue pointed out earlier. Natural objects come in all sorts of shapes which can not be solely modelled by disks or rectangles.

Therefore, we propose a random shape generator based on random polygon algorithms~\cite{auer1996heuristics,tomas2003generating} and on the computation of a concave hull~\cite{edelsbrunner1983shape} (also called $\alpha$-shape) of a set of points.

Our generation algorithm consists of the following steps:
\begin{enumerate}
    \item Sample $n$ points 
    %$(x_i)_{i\leq n},\text{ st } 
   $ x_i \sim \mathcal{U}(D(0,r))$, with $\mathcal{U}$ the uniform distribution and $D(0,r)$ the disk of radius $r$,
    \item Compute the associated Delaunay triangulation, 
    %$\text{DelaunayTriangles}((x_i)_{i\leq n}) = 
    %$\mathbf{T} = \{ T_k\}_{k\leq K}$,
    \item \textbf{Concave hull:} Let $e_{\min},e_{\max}$ denote the smallest and largest edge lengths of the triangulation. For a given $\alpha \in [0,1]$, remove triangles whose longest edge exceeds $e_{\min} +\alpha(e_{\max} -e_{\min})$, while preserving connectivity of all points $(x_i)_{i\leq n}$,
    %all triangles $\{ T_k\}_{k\leq K}$.
    \item Rasterize the obtained polygon $S$
    %$S \in [0,1]^{(2r+1,2r+1)} $,
    \item Apply Gaussian blur $S_b = S *G_{\sigma_s}$ 
    %with $\sigma_s \sim \mathcal{U}[0,\sigma_{s \max}] $, 
    %Yann : c'est précisé dans l'appendice
    and threshold the resulting image to get $S_c = \tau(S_b, 1/2)$, where $\tau$ is the thresholding operator.
\end{enumerate}

\begin{figure}[htp]
    \centering
    \begin{tabular}{cccc}
        \scriptsize
         \includegraphics[width = 0.22\linewidth]{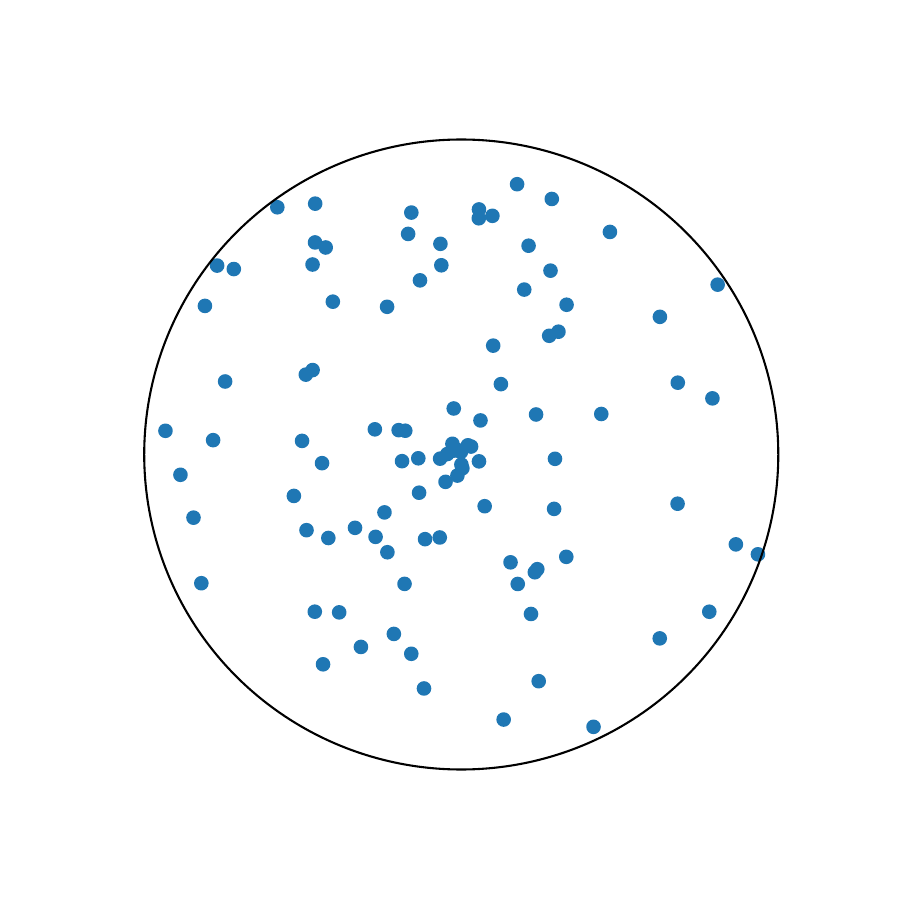}&\includegraphics[width = 0.22\linewidth]{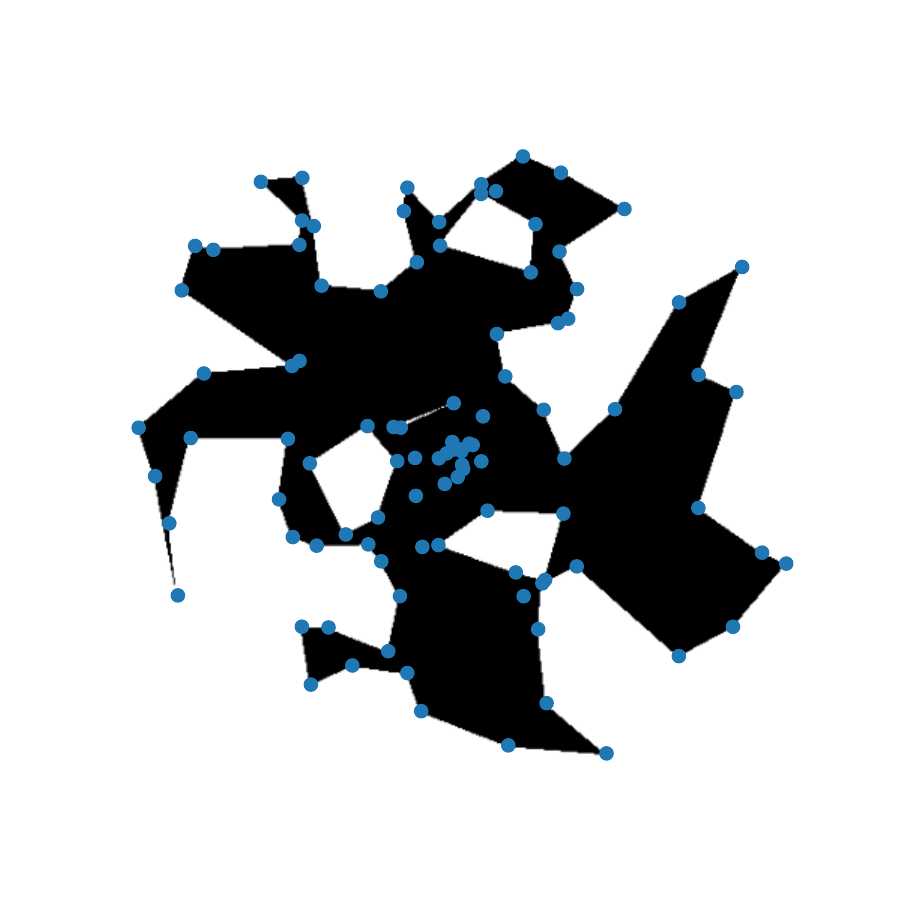}&\includegraphics[width = 0.22\linewidth]{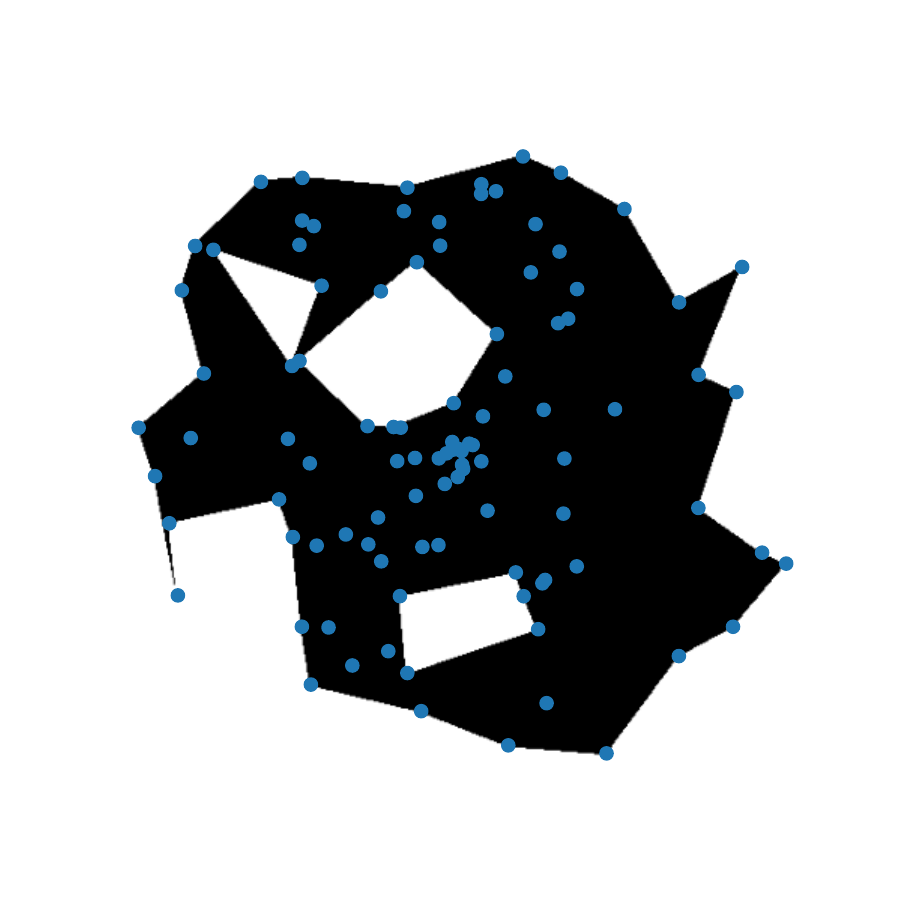}&\includegraphics[width = 0.22\linewidth]{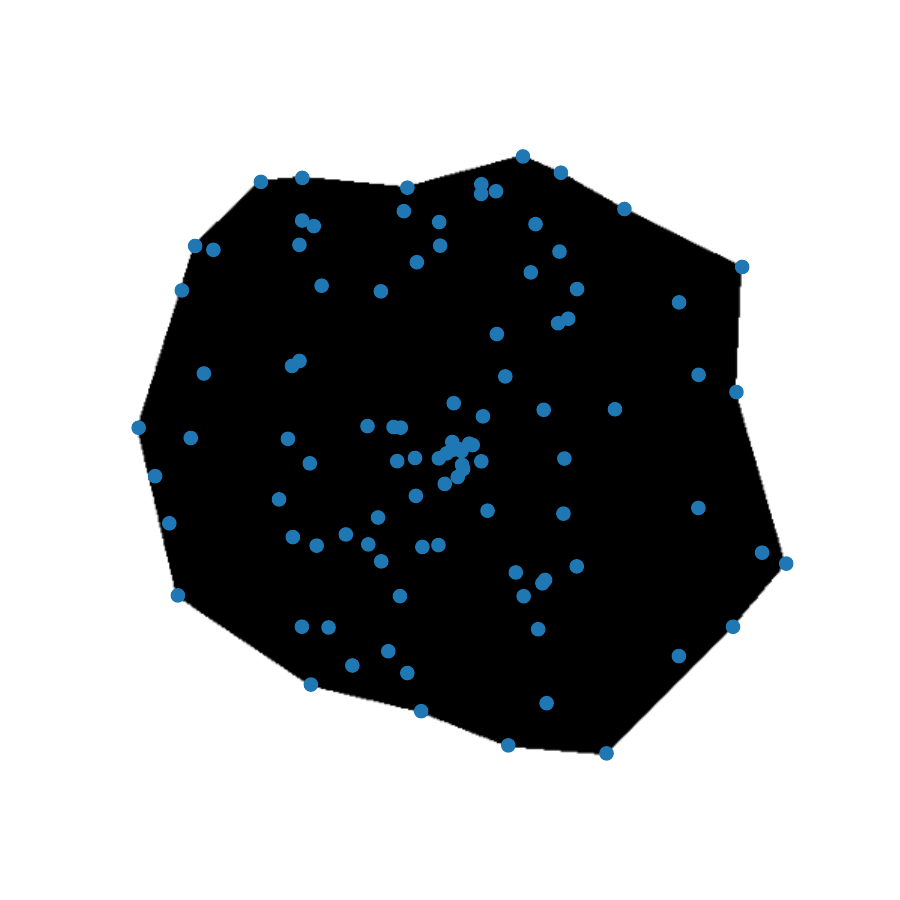}  \\
         $ x_i\sim \mathcal{U}(D_1)$ & $\alpha = 0.2$ & $\alpha = 0.4$ & $\alpha = 0.6$ \\
    \end{tabular}
    \caption{Random shape generation. We start by sampling $n$ points (here $n=100$) uniformly in a disk (see first picture), which ensures rotation invariance. Then, we compute the concave hull of $(x_i)_{i\leq n}$ for parameter $\alpha$. As shown here, the larger $\alpha$, the closer the concave hull is to the convex hull.}\label{fig:shape_gen}
\end{figure}

We now justify our design choices.
First, we sample points uniformly in a disk to guarantee rotation invariance.
%, one of the key properties of DL images.
Second, we use a concave hull algorithm from the GEOS library~\cite{GEOS}, which iteratively removes triangles from the Delaunay triangulation according to the procedure above,
%$\{ T_k\}_{k\leq K}$
starting from boundary triangles.
%, whose longest edge is larger than $e_{\min} +\alpha\times(e_{\max} -e_{\min} )$.
This also allows holes to be created by removing interior triangles.
%which are not in the boundary of the shape. 
%Since there is a strict constraint to maintain all points in the final shape, the algorithm converges to a minimal shape as $\alpha \rightarrow 0$.
%yann : pas compris 
%said: Je propose d'enlever la phrase sur la convergence , le lecteur risque de se demander si alpha evolue pendant l'algorithme de gneration d'une forme...
%yann : ok je supprime
\begin{figure}[htp]
    \centering
    \includegraphics[width=0.9\linewidth]{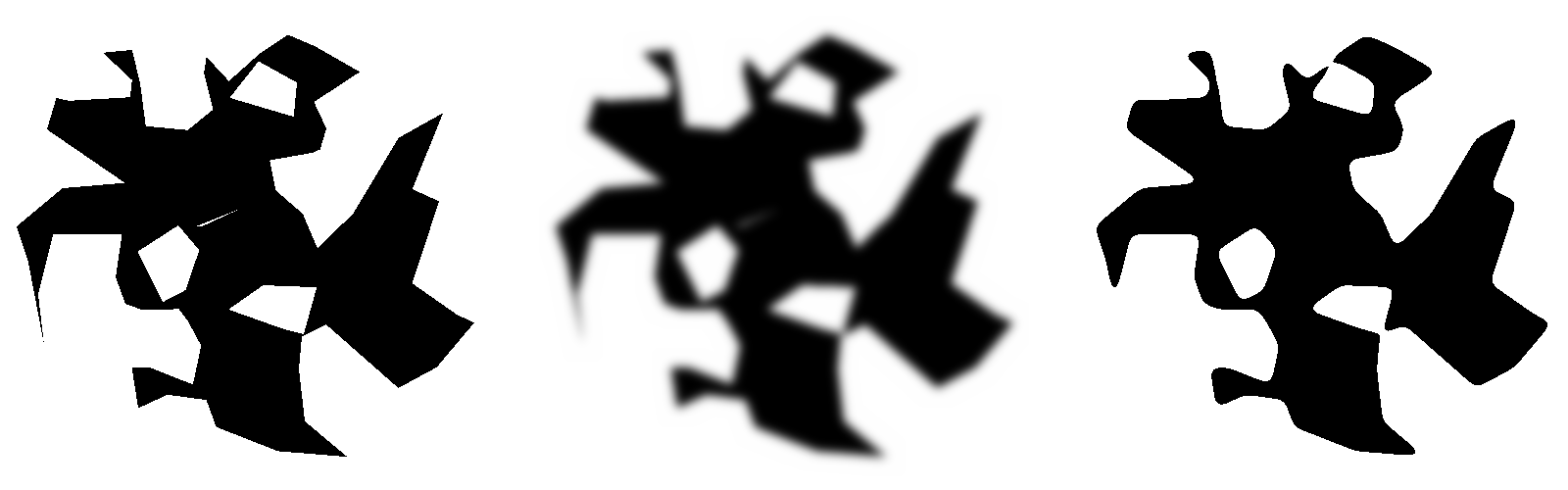}
    \caption{Gaussian smoothing of a random concave hull. From left to right: original shape, shape after Gaussian blurring, and final shape, obtained after thresholding.}\label{fig:gaussian_smoothing}
\end{figure}

Third, to obtain a wide variety of local curvatures, we smooth these random shapes using a rough approximation of the mean curvature motion. We convolve the binary shape mask with a Gaussian kernel $G_{\sigma_s}$ and threshold the blurred image at level $0.5$, as shown in Figure~\ref{fig:gaussian_smoothing}.

% An interesting benefit of this technique is that the algorithm is able to create a wide variety of curvatures, as may be found in natural objects.

\begin{figure}[ht]
    \centering
    \includegraphics[width = \columnwidth]{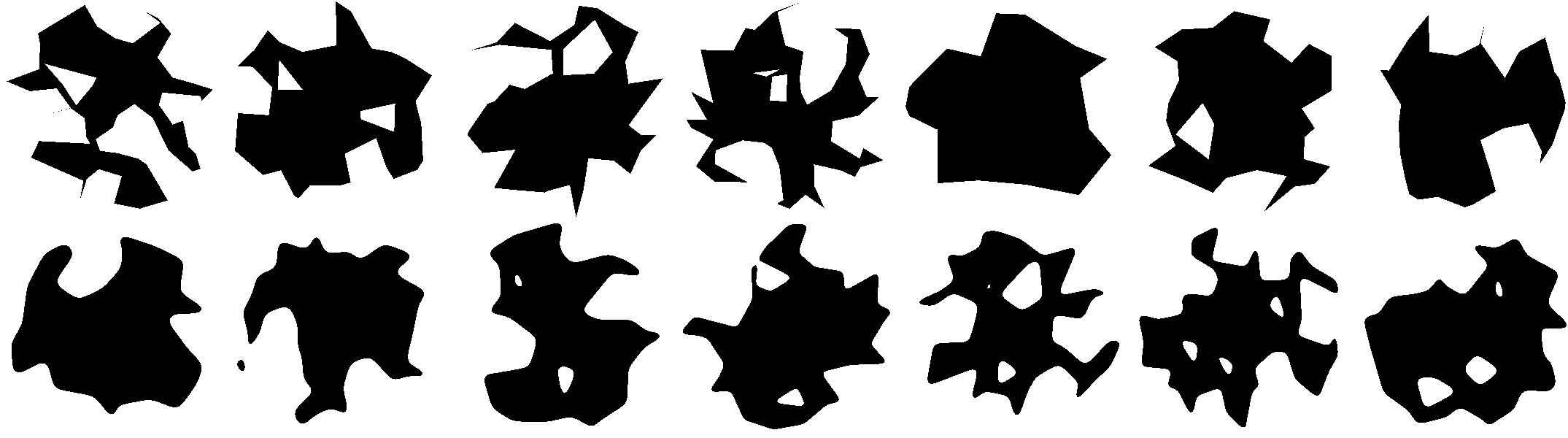}
    \caption{Samples from our shape generator. First row: sharp polygons. Second row: shapes after Gaussian smoothing.}\label{fig:shapes_examples}
\end{figure}

The proposed algorithm produces concave shapes, acute angles, straight boundaries, holes, and a wide variety of curvatures, as can be seen in Figure~\ref{fig:shapes_examples}. It relies only on four parameters: $(\alpha,\sigma_{s},n,r)$, whose values are specified in the supplementary material.
The resulting diversity of shapes would be difficult to match using, e.g., the starred polygons introduced in~\cite{auer1996heuristics}.

% We first sample $n$ points uniformly at random in the unit disc $D_1$ to maintain rotation invariance, one of the key properties of DL images. 
% We then compute the $\alpha\text{-shape}$\cite{edelsbrunner1983shape} (also called concave hull) associated with this set of points. 
% The parameter $\alpha \in (0,1)$ acts as a regularity parameter :  the larger the $\alpha$, the closer the shape is from the convex hull, which is regular and similar to the unit disc for large values of $n$, as illustrated in  Figure~\ref{fig:shape_gen}.
% This approach generates relatively complex shapes from only two parameters, possibly including holes and irregular branches. 

These shapes are mixed with circles and rectangles in the final VL model. For all shapes, the size distribution is chosen as a power law $f(r)=C\cdot\mathbf{1}_{[r_{\min},r_{\max}]}(r)\cdot r^{-\gamma}$, with $\gamma = 3$. This choice is consistent with natural image statistics and ensures scale invariance up to the minimal radius $r_{\min}$. The proportions of each shape type and all other hyperparameter choices are detailed in supplementary Section~\ref{sec:sup_implem}.

%Overall, the proposed generator allows for a better diversity of object shapes, which better matches with the diversity observed in natural images.

\subsection{Textures}
The synthetic dataset introduced in \cite{achddou_cviu} is obtained from a dead leaves model where each shape is colored uniformly, resulting in flat surfaces. This is a severe limitation for modeling natural images, where most objects are textured.

Though texture synthesis has been extensively studied,
%in the last 40 years\cite{heeger1995pyramid,portilla2000parametric,efros1999texture,wei2000fast},
most methods aim to reproduce the visual appearance of an existing exemplar, by either matching its marginal statistics~\cite{portilla2000parametric,gatys2015texture} or by patch-based copy-and-paste~\cite{efros1999texture,wei2000fast}. Our purpose is different: we seek a compact, exemplar-free parametric texture model.
We focus on two essential aspects of visual textures: repetitions and randomness, and propose two distinct texture generators — a pseudo-periodic pattern generator and a random micro-texture generator — which we combine into a two-scale texture model.

\begin{figure*}[b]
    \centering
    
    \resizebox{\linewidth}{!}{
    \begin{tabular}{ccggccgg}
         \multicolumn{2}{c}{Sinusoidal textures} & 
         \multicolumn{2}{g}{Clipped sinusoid} & \multicolumn{2}{c}{Random periods} & \multicolumn{2}{g}{Warped textures} \\ 

         \includegraphics[width = 0.13\textwidth]{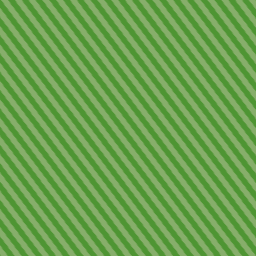} 
         & \includegraphics[width = 0.13\textwidth]{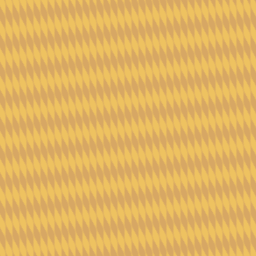} 
         & \includegraphics[width = 0.13\textwidth]{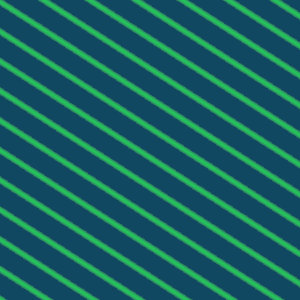}
         & \includegraphics[width = 0.13\textwidth]{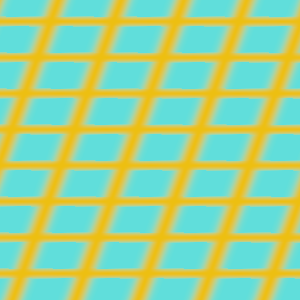}
         & \includegraphics[width = 0.13\textwidth]{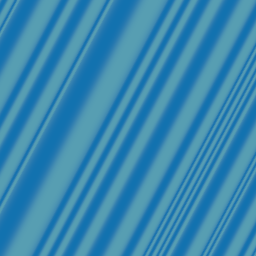} 
         & \includegraphics[width = 0.13\textwidth]{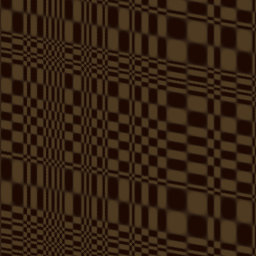} 
         & \includegraphics[width = 0.13\textwidth]{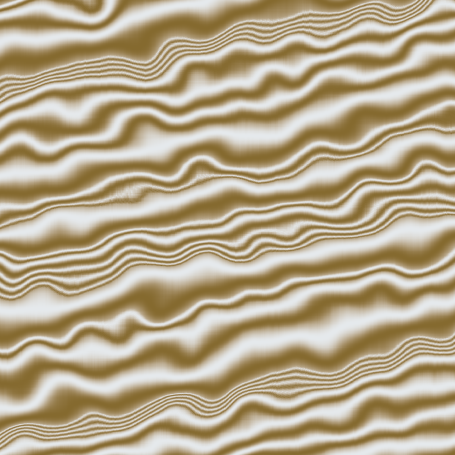}
         & \includegraphics[width = 0.13\textwidth]{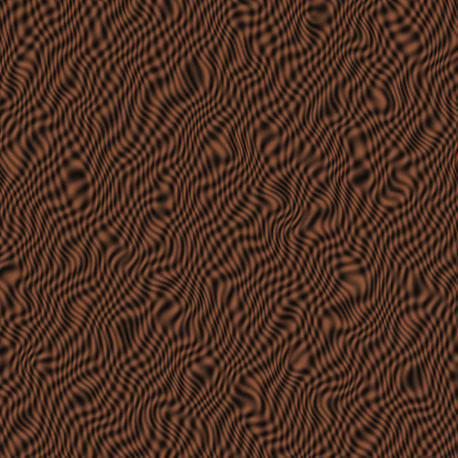} \\ 
    \end{tabular}}
    \caption{Pseudo-periodic texture samples from our texture model in either 1 or 2 dimensions.
    %Natural images are redundant, and this is particularly true for textures.
    We start from simple sinusoids and enrich them by: (1) thresholding the interpolation fields, (2) stacking sinusoids of random periods, and (3) randomly distorting texture maps with Gaussian vector fields.}\label{fig:sin_textures}
\end{figure*}

\begin{figure*}[ht]
    \centering
\resizebox{\linewidth}{!}{
    \begin{tabular}{cgcgcgc}
        $\gamma = 0.5$& $\gamma = 0.833$& $\gamma = 1.166$& $\gamma = 1.5$& $\gamma = 1.833$& $\gamma = 2.166$& $\gamma = 2.5$\\ 
     
         \includegraphics[width = 0.133\textwidth]{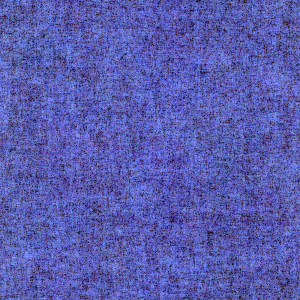} & \includegraphics[width = 0.133\textwidth]{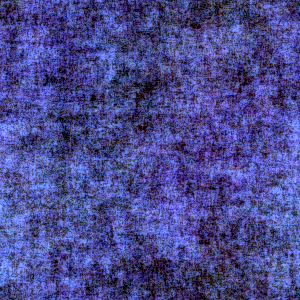} & \includegraphics[width = 0.133\textwidth]{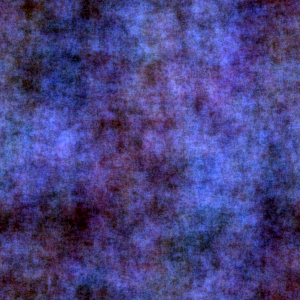} &  
         \includegraphics[width = 0.133\textwidth]{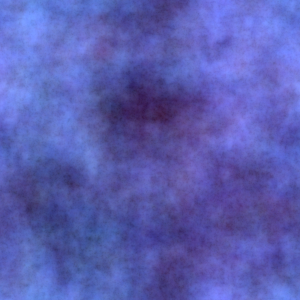} & 
         \includegraphics[width = 0.133\textwidth]{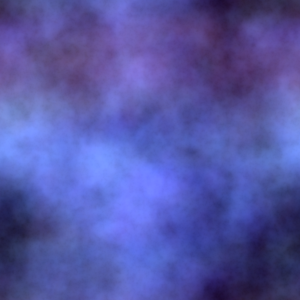}&
        \includegraphics[width = 0.133\textwidth]{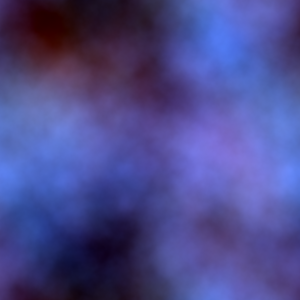}&
        \includegraphics[width = 0.133\textwidth]{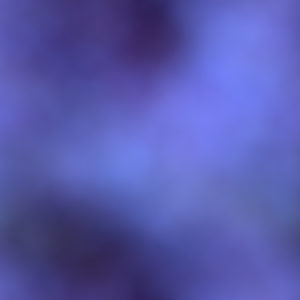}
    \end{tabular}
}
    \caption{Samples from our micro-texture model, all obtained from the same color histogram. These images are made with colored noise with an increasing slope $\gamma$ in the power spectrum from left to right. The smaller $\gamma$, the richer the high-frequency content; the larger $\gamma$, the smoother the texture.}\label{fig:micro_textures}
\end{figure*}

\subsubsection{Pseudo-periodic patterns}\label{sec:macrotext}
These patterns are common in natural images (zebra fur, honeycombs, etc.) and urban scenes (brick walls, windows, etc.). To mimic them, we use pseudo-periodic interpolation fields between two randomly sampled colors $(c_1,c_2)$, with interpolation performed in CIELAB — a perceptually uniform color space. Colors are sampled from the color histogram of a single randomly chosen natural image. As justified in~\cite{achddou2021synthetic}, this yields better color rendering than uniform sampling (which is unnatural) or direct sampling from a global histogram (which lacks coherence~\cite{achddou2023learning}). Further coherence could be obtained by modeling second-order dependencies between colors, but this would entail significantly heavier modeling.

We start with simple sine functions 
%and  define $T_{0,\omega}(x,y) = 
$x \mapsto\sin( 2\pi x/t)$, where $t \sim \mathcal{U}[T_{\min}, T_{\max}]$. 
We then apply a sigmoid transform of varying slope to sinusoids,
%$S_{\omega}$,
to create sharper transitions and enrich the harmonic content of the patterns.
%as sinusoids are soft, especially with smaller $\omega$.
We write the resulting field as follows: $S_{t}(x,y) = \sigma_{\lambda}(\text{sin}(2\pi x/t))$, where $(x,y)\in \mathbb{R}^2$ is the position and $\sigma_{\lambda}$ is a logit function of growth rate $\lambda$, linearly rescaled so that it ranges in $[0,1]$.
%Note that the field only depend on the $x$ coordinate.
To obtain random orientations, we rotate this field by a randomly sampled angle
%sampled uniformly
%Yann : je pense que cette precision doit aller dans l'annexe
in $[0,\pi/4]$.
To obtain a 2-dimensional field, we repeat the same process along the $y$ axis and multiply the result by the previously obtained sinusoidal field. The rendered image is the linear interpolation $I(x,y) = S_{t}(x,y)\cdot c_1 + (1-S_{t}(x,y))\cdot c_2$.
Examples of such textures are shown in the first column of Figure~\ref{fig:sin_textures}.

Next, we clip the periodic functions $S_{t}(x,y)$ with a randomly sampled value $\tau \in [0,1]$. This produces grid-like interpolation fields and thin lines, commonly found in urban and natural environments, as shown in the second column of Figure~\ref{fig:sin_textures}.

%Such sinusoids are too regular to account for natural and urban patterns for the following reasons: 
%\begin{itemize}
 %   \item the interpolation field is equally distributed between the two colors, 
  %  \item the period is constant across the interpolation field,
   % \item they have no curvature. 
%\end{itemize}     

%We introduce simple additions to correct these defects.
%First, we threshold the periodic functions $S_{t}(x,y)$ with a random picked value $\tau \in [-1,1]$. This allows to create grid-like interpolation fields and thin lines, commonly found in urban and natural environments. }

% Then, we spatially vary the frequency of these sinusoids as follows. 
To further enrich our pseudo-periodic patterns, we replace the aforementioned sinusoids with the function $F(x,y)$ described as follows.
We sample a sequence of $k$ periods $(t_i)_{i \leq k} \sim \mathcal{U}[T_{\min},T_{\max}]$ with total length $T = \sum t_i$, and tile this pattern to obtain a global period of $T$.
Writing $\tilde{x} = x \text{ mod } T$, the resulting 1-dimensional oscillatory field is:
\begin{align}
    F(x,y) = \begin{cases}
    \text{sin}(\frac{2 \pi}{t_0}\tilde{x}), & \text{if $\tilde{x} \in[0,t_0]$}.\\ 
    \text{sin}(\frac{2 \pi}{t_i}\tilde{x}), & \text{if $\tilde{x} \in \left (\sum_{j = 0}^{i-1}t_j,\sum_{j = 0}^{i}t_j\right ]$}.
    \end{cases}
\end{align}
As before, a sigmoid transform $\sigma_{\lambda}$ and clipping are applied to the resulting field.

Finally, we randomly distort these interpolation fields using random vector fields from an atmospheric turbulence generator~\cite{meinhardt2014implementation}.
Specifically, a random vector field $M \in \mathbb{R}^{2 \times H \times W}$ is generated as white Gaussian noise ($M_{i,j} \sim \mathcal{N}(0,1)$), blurred with a Gaussian kernel of standard deviation $s$ for regularity, and rescaled by a parameter $t$ to control the displacement range.
%Both parameters are randomly sampled in a range specified in the supplementary material. 
Examples of the obtained texture maps are given in the fourth column of Figure~\ref{fig:sin_textures}.

\subsubsection{Random micro-textures}\label{sec:microtext}
Pseudo-periodic textures, as presented in the previous section, correspond to the random repetition of textons at a macroscopic scale.
When the texton size falls below pixel resolution, individual textons are not resolved, giving rise to micro-textures — very common in natural images (grass, sand, clouds from afar) — which can be efficiently modeled as correlated Gaussian noise fields~\cite{galerne2010random}.

In terms of statistics, such micro-textures are described by their covariance (equivalently, their power spectrum under the stationarity assumption) and color marginals.
This has been exploited for exemplar-based synthesis by randomizing the phase of an exemplar's power spectrum~\cite{galerne2010random}.
Inspired by this, we propose a parametric, exemplar-free micro-texture generator based on the well-known average power spectrum decay of natural images, $|\hat{I}(\nu)|^2 \propto 1/|\nu|^{\gamma}$ (where $\nu$ is the spatial frequency)~\cite{ruderman1994statistics,burton1987color}.

%Our generation of micro-textures starts by sampling white noise from the color histogram of a randomly chosen natural image.

We first generate a white noise image by sampling colors from the color histogram of a randomly chosen natural image; as noted above, this significantly outperforms uniform color sampling, which produces unrealistic colors~\cite{achddou2021synthetic}.
We then impose a linear decay of the power spectrum in the log-domain with slope $\gamma$, giving $|\hat{I}(\nu)|^2 \simeq 1/|\nu|^{\gamma}$.
The spectrum model is isotropic to preserve rotation invariance.
As shown in Figure~\ref{fig:micro_textures}, the larger $\gamma$, the smoother the texture.
% Visually, the obtained micro-texture can remind the reader of patterns observed in natural images such as plants, minerals or clouds.
% With this addition to the DL model, we hope that the learnt denoiser should better tackle such textures.

% \begin{itemize}
%     \item Limit of textures in the current model: Textures only arise from the appearance of tiny disks / cannot produce natural-looking textures.
%     \item Definition of textures "a pattern that repeats itself with random modifications at different scales": a notion of repetitions and a notion of randomness and scale.
%     \item Texture 1: sinusoidal textures / explain warping: present examples of textures in natures that are very repetitive
%     \item Texture 2: micro textures: colored noise with appropriate color model
%     \item Texture 3: multi-texture objects : morphing textures in object
% \end{itemize}

\subsubsection{A micro-macro texture model}
\begin{figure}[hb]
    \centering
    \resizebox{1.0\columnwidth}{!}{
    \begin{tabular}{cccg}
        Text.1 & Text.2 & Mix. mask & Fused texture\\ 
         \includegraphics[width=0.23\linewidth]{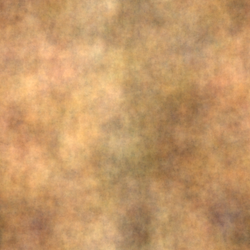}& 
         \includegraphics[width=0.23\linewidth]{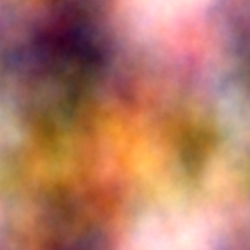}&
         \includegraphics[width=0.23\linewidth]{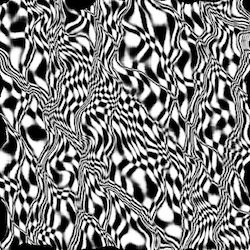}& 
           \includegraphics[width=0.23\linewidth]{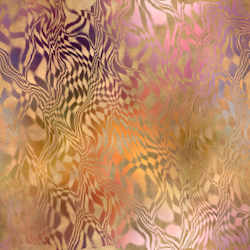}
    \end{tabular}}
    \caption{Macro-Micro texture model.}\label{fig:biscale_text}
\end{figure}

The two previous models focus on either macroscopic oscillating patterns or micro-textures, whereas natural objects often exhibit both simultaneously. For instance, pebbles can be seen as a repetition of objects of similar shape and size, each characterized by a distinct micro-texture.
We therefore merge them into a two-scale model: we generate two micro-texture maps $(T_1,T_2)$ as in Section~\ref{sec:microtext} and a pseudo-periodic interpolation mask as in Section~\ref{sec:macrotext}, then blend the maps as $T_{\text{blended}} = \beta T_1 + (1-\beta)T_2$ in CIELAB\@.
This produces textures at multiple scales, as illustrated in Figure~\ref{fig:biscale_text}.
The proportions of each texture type and all other hyperparameter choices are detailed in supplementary Section~\ref{sec:sup_implem}.

% \begin{itemize}
%     \item explain the motivation : objects can have different textures
%     \item textures have different scale: pebbles have micro textures on there surface but pebbles are also repeated in a semi-random semi repetitive mannners.
%     \item idea: combine both macroscopic repetitive patterns and microtextures into a single model.
%     \item How? take a macroscopic sinusoidal field as an interpolation mask
%     \item generate two micro-texure maps and combine them with this technique -> creates anisotropic textures.
% \end{itemize}

% \begin{figure}[b]
%     \centering
%     \includegraphics[width=\linewidth]{images/depth/depth_of_field_2.png}
%     \caption{Diagram of the depth-of-field algorithm. After generating three DL stack (background, middle-ground and foreground), we fuse them by applying blur kernels $G_{\sigma_1},G_{\sigma_2}$ respectively to the background and foreground. 
%     % While the application of this blur amounts to a convolution for the background, it is a bit more complicated for the foreground blur, as it needs to account for the pixels of the combination of the background and middle-ground. Using the binary masks defined by the DL stack, we are able to combine the images successfully.
%     }\label{fig:depth_of_field}
% \end{figure}

\subsection{Depth}
% Photographs are the projection of a 3-dimensional word on a 2-dimensional sensor. 
% Therefore, the aspect of a photograph greatly depends on the position the objects with respect to the imaging device as well as the device's optical properties.
Physical depth can be perceived in several ways in photographs. Occlusions occur as objects block light-paths from the object to the sensor. Limited depth-of-field causes blur when the aperture of the lens is wide open. Finally, perspective and vanishing points are other depth clues, especially important in urban environments.
While the occlusion phenomenon is already properly modeled in classical DL images, perspective and depth-of-field are not.
In this section, we propose two additions to better render these properties.

\subsubsection{Depth-of field}\label{sec:dof}
In order to recreate this aspect of natural images, a substantial amount of techniques exist in the Computer Graphics litterature\cite{demers2004depth}.
For accurate modeling, renderers model the physical response of a lense and use ray tracing.
However, this process comes with high computational costs, and requires to model our scene in 3D. Since our goal is to develop simple synthetic models with relatively few parameters, we keep the $2D+t$ approach offered by the dead leaves models and disregard such approaches. 
%As explained in Section~\ref{sec:background}, DL images are the result of a tesselation of the 2D-image plan, which doesn't concur with a 3D-modeling. 

Instead, we use a simpler model, initially proposed for 3D scene rendering, which amounts to split the z-axis in layers and apply different blur kernels for different layers, before combining them in a single fused image\cite{scofield1992212}.
When 3D objects spread over multiple layers, this approach may create artifacts at layers boundaries. Nonetheless, since we consider 2-dimensional shapes that are perfectly parallel to the projection plane, we can use this approach without creating any artifacts. 

% We illustrate this approach in Figure~\ref{fig:depth_of_field}. 

Consider a foreground image $F$ with  support $M$, in front of a background image $B$. We assume that the foreground is affected by blur $g_{_F}$ and the background by blur $g_{_B}$ so that the resulting image is approximately given by the formula (which also defines the $\oplus$ operation)~:

$$
(B,g_{_B})\oplus(F,g_{_F},M):=(B\ast g_{_B})\times (1-M\ast g_{_B}\ast g_{_F})+(F\ast g_{_F})
$$
The main feature of this equation, compared to simple convolutions and masking, is the multiplication part that accounts for the smooth extinction of the background rather than a sharp masking. It should be noted that this formula is exact when the background is in focus ($g_{_B}=\delta$). When the foreground is in focus then the exact version should be
$$
B\ast_v g_{_B}+F (=F\ast \delta)
$$
where the operation $\ast_v$ is a variable convolution against, not merely $g_{_B}$, but a variable version of $g_{_B}$ that accounts for the occlusion of the camera lens by the mask $M$ when viewed from the background. Instead of this complex operation we chose to use a smooth extinction that is more tractable. 

We consider a stack of three planes $I_1$, $I_2$ and $I_3$ with blurs $G_{\sigma_1}$, $\delta$ (in focus) and $G_{\sigma_3}$ respectively and masks $M_1$ and $M_2$ (the background $I_3$ has a trivial mask). Then the final stacking is given by~:
$$
\left[(I_3,G_{\sigma_3})\oplus(I_2,\delta,M_2),\delta\right]\oplus(I_1,G_{\sigma_1},M_1)
$$

Since the expected benefit of this feature of the model is to obtain realistic transitions (thus involving two objects only) we did not find it useful to consider more than three planes.

\begin{figure}[ht]
    \centering
    \includegraphics[width=\linewidth]{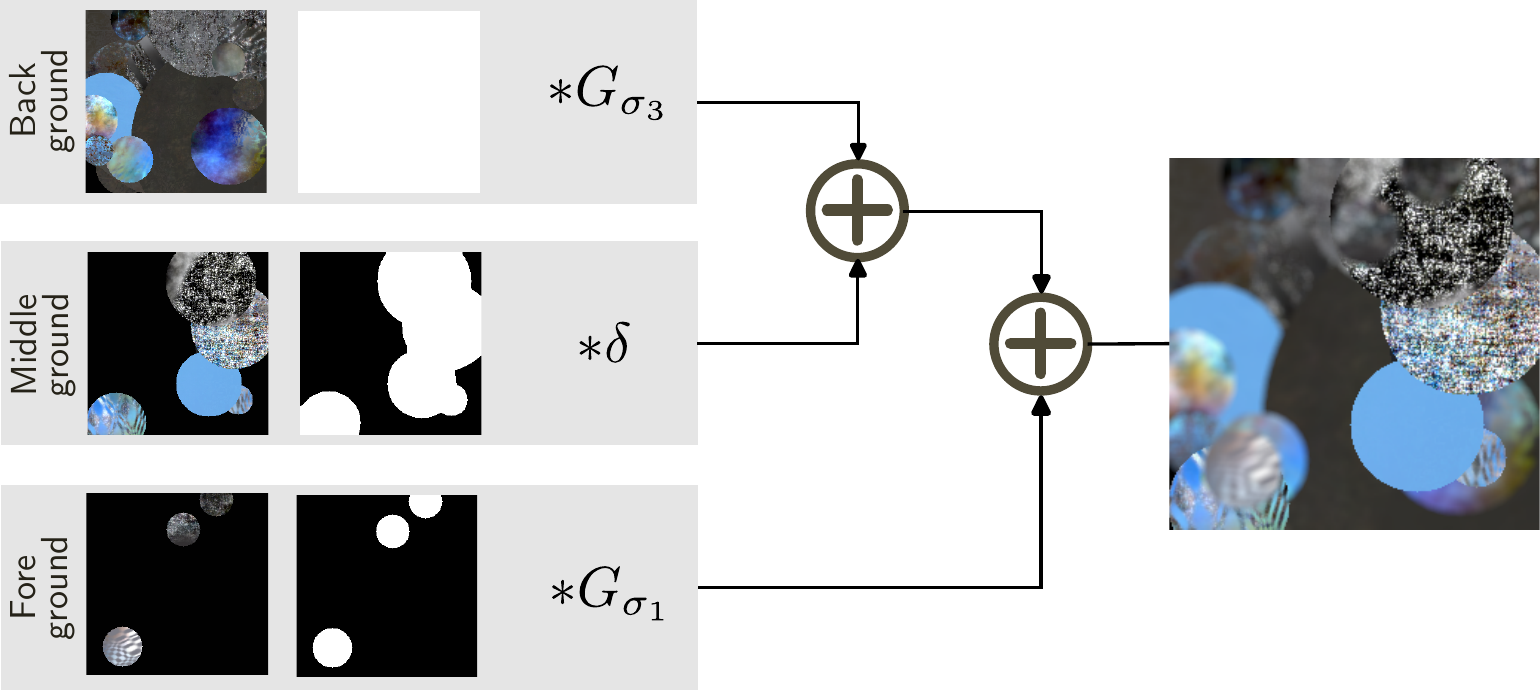}
    \caption{Depth-of-field diagram}
    \label{fig:depth}
\end{figure}

% Here, we only suppose a decomposition of the z-axis into 3 plans. 
% % We first generate 3 DL stacks: a background, which covers the whole image plan, a middle ground which covers $25\sim50$ \% of the image plan, and a foreground which covers $10 \sim 25$ \% of the image plan, all generated with the same hyper parameters.
% We first generate 3 DL stacks: a background, a middle ground, and a foreground.
% We then convolve the background with a blur kernel $G_{\sigma_1}$ and superimpose the middleground (which we suppose is in focus) on it.
% Given a background image $I_1$ and a middleground image $I_2$ characterized by a mask $M_2$, this addition is formulated as:
% \begin{align}
%     \hat{I}_2 = I_1 \oplus (I_2,M_2) (i,j)= I_1(i,j).(1-M_2(i,j))+ I_2(i,j)
% \end{align}
% where $(i,j)$ are the pixel's coordinates.
% Combining this image with the foreground is not as straightforward as it requires to take $\hat{I}_2$ into account as background information.
% Therefore we start by adding the foreground $I_3$ to $\hat{I}_2$. We blur the resulting image as well as the foreground's binary mask with another blur kernel $G_{\sigma_2}$, leading to a blurred image and mixing mask $(\hat{I_3},\hat{M_3})$. 
% We combine the obtained image with $\hat{I}_2$ by addition: 
% \begin{align}
%     I_{\text{fused}} = \hat{I}_2 \oplus (\hat{I_3}.\hat{M_3},\hat{M_3})
% \end{align}

\subsubsection{Perspective}\label{sec:perspective}
The second property related to depth that we incorporate is perspective.
Perspective is particulary visible in images when parallel lines converge to a single point called a \textit{vanishing point}.
Instead of operating at the shape level,
%which are 2-dimensional
 we directly modify our texture synthesis algorithm to recreate this effect.
 We apply a perspective homographic transform to texture maps so that the corners coordinates of the original map $p_i \in \mathbb{N}^{2}, i \in[0,3]$ are transformed into new corner coordinates $\tilde{p_i}$. 
The coefficients of the homography matrix are obtained as the solution of an ordinary least square problem.

\section{Examples and statistical properties}

% \begin{algorithm}[htp]\label{algo1}\caption{VL generation algorithm}
% \footnotesize
% \SetKwInOut{Input}{Parameters}
% \SetKwInOut{Output}{Output}

% \Function{LeavesStack ($r_{\min},r_{\max},\alpha$,$p$,colors)}{
% \Comment{$p\in [0,1]$ corresponds to the expected coverage of the image plane}
% \var{stack}, \var{mask} $\assign$ \FuncCall{zeros}{$w,w,3$},\FuncCall{zeros}{$w,w$};\\
% \While{$||\var{mask}||/w^2 < p$}
% {
%      $x,y \assign \mathcal{U}[0,w]^2$ \\
%     \var{shape\_mask} $\assign$  \FuncCall{shapeGenerator}{$r_{\min},r_{\max},\alpha$}; \\
%     \var{texture} $\assign$ \FuncCall{textureGenerator}{\var{colors}};\\
% x    \var{mask} $\assign$ \FuncCall{MaskUpdate}{\var{mask},\var{shape\_mask},$x$,$y$};\\
%     \var{stack} $\assign$ \FuncCall{StackUpdate}{\var{stack},$x$,$y$,\var{shape\_mask},\\ \var{texture}};
% }
% \Output{\var{stack}, \var{mask}} 
% }
% \Function{VL ($r_{\min},r_{\max},\alpha$)}{
% \Comment{This function merges background, middleground and foreground DL stacks.}
% \var{colors} $\assign$  \FuncCall{SampleNatHistogram}{};\\
% \var{b}, \var{b\_m} $\assign$ \FuncCall{LeavesStack}{$r_{\min},r_{\max},\alpha$,1,\var{colors}};\\
% \var{m}, \var{m\_m} $\assign$ \FuncCall{LeavesStack}{$r_{\min},r_{\max},\alpha$,1/2,\var{colors}};\\
% \var{f}, \var{f\_m} $\assign$ \FuncCall{LeavesStack}{$r_{\min},r_{\max},\alpha$,1/4,\var{colors}};\\
% \var{merged} $\assign$ \FuncCall{MergeStacks}{\var{b}, \var{b\_m},\var{m}, \var{m\_m},\var{f}, \var{f\_m}};\\
% \Output{\var{merged}} 
% }
% \end{algorithm}

\begin{algorithm}[htp]\label{algo1}\caption{VL generation algorithm}
\footnotesize
\SetKwInOut{Input}{Parameters}
\SetKwInOut{Output}{Output}

\Function{LeavesStack ($p$,$w$)}{
\Comment{$p\in [0,1]$ corresponds to the expected coverage of the image plane}
\var{stack}, \var{mask} $\assign$ \FuncCall{zeros}{$w,w,3$},\FuncCall{zeros}{$w,w$};\\
\While{$||\var{mask}||/w^2 < p$}
{
     $x,y \assign \mathcal{U}[0,w]^2$ \\
    \var{shape\_mask} $\assign$  \FuncCall{shapeGenerator}{}; \\
    \var{texture} $\assign$ \FuncCall{textureGenerator}{};\\
    \var{mask} $\assign$ \FuncCall{MaskUpdate}{\var{shape\_mask},$x$,$y$};\\
    \var{stack} $\assign$ \FuncCall{StackUpdate}{$x$,$y$,\var{shape\_mask},\\ \var{texture}};
}
\Output{\var{stack}, \var{mask}} 
}
\Function{VL ()}{
\Comment{This function merges background, middleground and foreground DL stacks.}
\var{colors} $\assign$  \FuncCall{SampleNatHistogram}{};\\
$s_b$ $s_{b,\text{mask}}$ $\assign$ \FuncCall{LeavesStack}{1,512};\\
$s_m$ $s_{m,\text{mask}}$ $\assign$ \FuncCall{LeavesStack}{1/2,512};\\
$s_f$ $s_{f,\text{mask}}$ $\assign$ \FuncCall{LeavesStack}{1/4,512};\\
\var{merged} $\assign$ \FuncCall{depthFuse}{$s_b,s_{b,\text{mask}}, s_m,s_{m,\text{mask}},s_f,s_{f,\text{mask}}$};\\
\var{output} $\assign$ \FuncCall{downScale}{\var{merged},2};\\
\Output{\var{output}} 
}
\end{algorithm}

Algorithm 1 summarizes the \DLplus~formation. More implementations details are provided in the supplementary Section~\ref{sec:sup_implem}, with a detailed presentation of the hyper-parameters choices. We remind the reader that the code is accessible via the following \href{https://github.com/rachddou/DeadLeavesPlus}{link}.
Next, we present some samples of the VL model as well as a statistical evaluation of other synthetic image datasets.

\begin{figure*}[h]
    \centering
    \includegraphics[width = 0.23\textwidth]{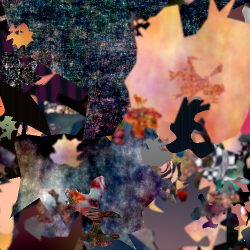}
    \hfill
    \includegraphics[width = 0.23\textwidth]{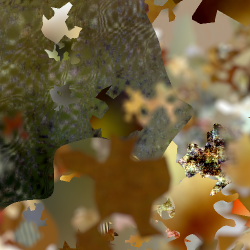}
    \hfill
    \includegraphics[width = 0.23\textwidth]{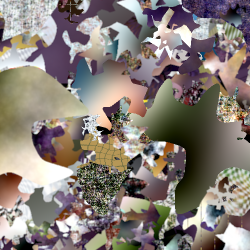}
    \hfill
    \includegraphics[width = 0.23\textwidth]{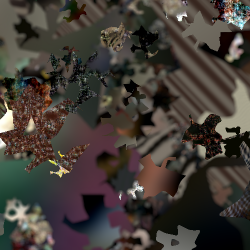}
    \caption{Examples of samples from the \DLplus~model, which integrates modeling for \textit{geometry, textures, and depth.}}
    \label{fig:examples}
\end{figure*}

\subsection{Samples of the VL model} 
The VL model has the ability to render geometry, textures, and depth, and produces images with a much more natural aspect than classical DL. Examples of VL samples are provided in Figure~\ref{fig:examples}. Additional examples are provided in the supplementary material.

\subsection{Statistical validation}
Not only do the generated images better match natural images perceptually, they also reproduce their second-order statistics more faithfully.
As explained in~\cite{alvarez1999size,lee2001occlusion}, DL images tend to match two important second-order statistics: the image gradient distribution and the power spectrum profile. We now verify that these statistics, along with others, are reproduced even better by the VL model than by the standard DL model. These measurements are provided to illustrate the gain in statistical similarity to natural images offered by the proposed model, and not as a standalone performance criterion. This is because obtaining higher similarity scores is a separate and rather straightforward problem — for instance by using large generative models (incompatible with our compactness goal) or by directly optimizing the model's parameters to match these statistics.

Regarding the image gradient, Figure~\ref{fig:stats} shows a closer match to natural images for the VL model than for DL\@.
The gradient histogram of DL images shows a strong peak at zero, attributable to the piecewise constant regions (the leaves). The addition of textures in the VL model mitigates this peak at zero, as shown in the linear-scale view of Figure~\ref{fig:stats}. Moreover, the gradient histogram of natural images is very close to that of VL images for intermediate gradient magnitudes, i.e., $\|\nabla I\|\in[0.1,0.6]$.
Numerically, the Kullback-Leibler (KL) divergence between the gradient distributions of synthetic and natural images decreases from $0.28$ to $0.006$ for DL and VL images respectively (see Table~\ref{tab:similaritymetrics}).

\begin{figure*}[b]
    \centering
    \begin{tikzpicture}
      \begin{axis}[
          height=0.25\textheight, % Scale the plot to \linewidth
          width = 0.4\textwidth,
          xlabel= $||\nabla(I)||_2$, % Set the labels
          ylabel= \# of occurences,
          % x tick label style={rotate=90,anchor=east}
          x post scale=1.25
        ]
        \addplot[mark=none,thick, color=black] table[x=x,y=y,col sep=comma] {csv_files/gradient_natural.csv};
        \addplot[mark=none, thick,color = blue] table[x=x,y=y,col sep=comma] {csv_files/gradient_complete.csv};
        \addplot[mark=none,thick, color=red] table[x=x,y=y,col sep=comma] {csv_files/gradient_og_dl.csv};
        
        \coordinate (subplot) at (0.35,40000);
      \end{axis}
    \begin{axis}[
            height=0.2\textheight,
            at=(subplot),
            % title = log-linear representation,
            % every axis title/.style={below right,at={(0,1)},
            ymode=log,
            grid=major,
            grid style={dashed,gray!30},
            legend style={at={(0.7,0.95)},anchor=north},
            axis background/.style={fill=gray!10},
            % log ticks with fixed point,
            style={font=\footnotesize}% Set the labels
    ]
        \addplot[mark=none,thick, color=black] table[x=x,y=y,col sep=comma] {csv_files/gradient_natural.csv};
        \addplot[mark=none, thick,color = blue] table[x=x,y=y,col sep=comma] {csv_files/gradient_complete.csv};
        \addplot[mark=none,thick, color=red] table[x=x,y=y,col sep=comma] {csv_files/gradient_og_dl.csv};
        \node[draw=black,fill=white,anchor=south west] at (rel axis cs:0,0) {\textit{log-linear representation}};
        \legend{\footnotesize Natural , \footnotesize VL,\footnotesize DL}
\end{axis}
    \end{tikzpicture}
    \hfill
    \begin{tikzpicture}[spy using outlines=
        {rectangle, magnification=2, connect spies}]
          \begin{axis}[
            height=0.25\textheight, % Scale the plot to \linewidth
            width = 0.4\textwidth,
            x post scale=1.25,
              ymode=log,
              xmode=log,
              grid=major,
              legend style={at={(0.1,0.1)},anchor=south west},
              grid style={dashed,gray!30},
              xlabel= log-frequency(cycle/pixels), % Set the labels
              ylabel= $\text{log}(|\hat{I}(\nu)|)$,
              % x tick label style={rotate=90,anchor=east}
            ]
            \addplot[mark=none,thick, color=black] table[x=x,y=nat,col sep=comma] {csv_files/spectrum_complete.csv};
            \addplot[mark=none, thick,color = blue] table[x=x,y=dl_++,col sep=comma] {csv_files/spectrum_complete.csv};
            \addplot[mark=none,thick, color=red] table[x=x,y=dl_og,col sep=comma] {csv_files/spectrum_complete.csv};
        \legend{\footnotesize Natural , \footnotesize VL,\footnotesize DL}
        \coordinate (spypoint) at (0.3,15);
        \coordinate (magnifyglass) at (0.25,1200);
          \end{axis}
          \spy [blue,dashed, size=2cm] on (spypoint)
       in node[fill=white] at (magnifyglass);
       
        \end{tikzpicture}

    \caption{Comparison of second-order statistics for DL (red), VL (\textcolor{blue}{blue}) and Natural Images (black).
    \textit{(Left).} Histograms of the image gradient $||\nabla(I)||_2$ estimated on 1K patches of size $(500 \times 500)$ randomly drawn from each datasets. The grey plot represents the same quantities in a log-linear representation for better visualization of the behavior at higher gradient values.
    \textit{(Right).} Average 1D power spectrum $(|\hat{I}(\nu)|)$ in a log-log representation for each datasets.}\label{fig:stats}
\end{figure*}

Similar conclusions hold for the power spectrum profile (Figure~\ref{fig:stats}). The spectrum is reduced to 1D by radial averaging, exploiting rotational invariance, and displayed on a log-log scale to illustrate the $1/|\nu|^{\gamma}$ decay. The spectral decay of VL images closely matches that of natural images; a linear regression in the log-log domain gives $\gamma_{\text{Nat}} = 1.44$, $\gamma_{\DLplus} = 1.41$, and $\gamma_{\text{DL}} = 1.79$ (see Table~\ref{tab:similaritymetrics}).

\begin{table*}[t]
    \centering
    \resizebox{1.0\textwidth}{!}{
    \begin{tabular}{gcccccc}

    \rowcolor{gray!10}Dataset&
    DL\cite{achddou2021synthetic}&
    ClevR\cite{johnson2017clevr}&
    GTA-V\cite{richter2016playing}&
    FractalDB\cite{kataoka2020pre}&
    DL-textured\cite{baradad2021learning}&
    \textbf{VL}\\
    \midrule
    
    Samples&
    \raisebox{-0.5\height}{\includegraphics[width = 0.1\linewidth]{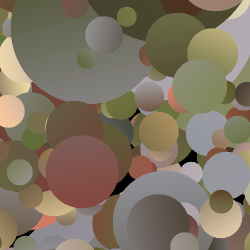}}
    &
    \raisebox{-0.5\height}{\includegraphics[width = 0.1\linewidth]{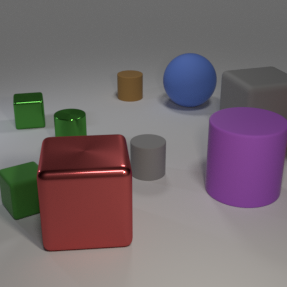}}& 
    \raisebox{-0.5\height}{\includegraphics[width = 0.1\linewidth]{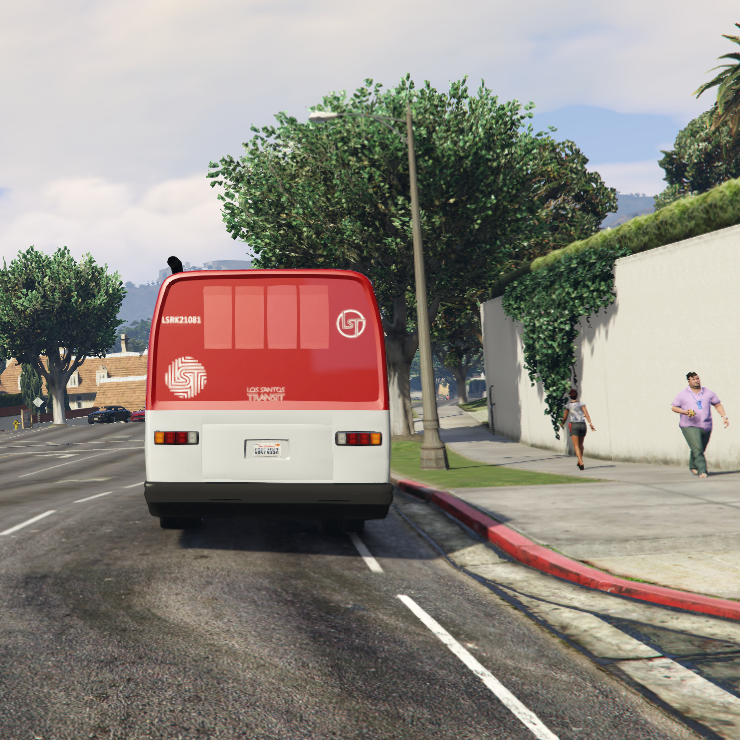}}&
    \raisebox{-0.5\height}{\includegraphics[width = 0.1\linewidth]{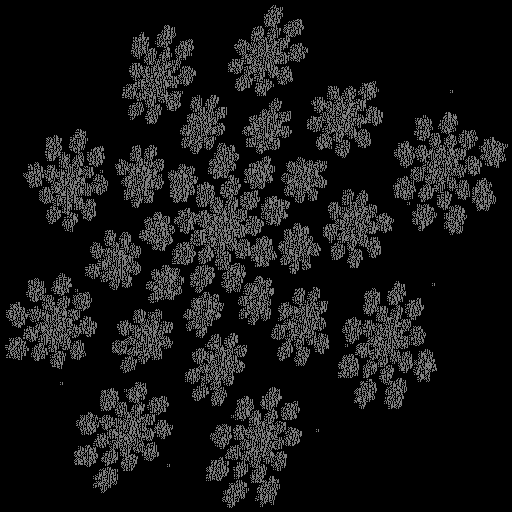}}& 
    \raisebox{-0.5\height}{\includegraphics[width = 0.1\linewidth]{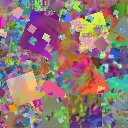}}&
    \raisebox{-0.5\height}{\includegraphics[width = 0.1\linewidth]{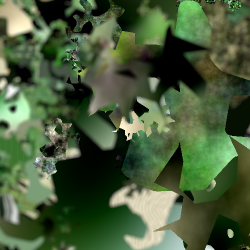}}
    \\
    \midrule

    \textit{KL-Gradient} $\downarrow$ &
    0.286&
    0.517&
    \second{0.015}&
    1.917&
    0.228&
    \textbf{0.006}
    \\
    \midrule

    \makecell{$\gamma_{\text{Spectrum}}$  
    $(\gamma_{\textit{Nat}} = 1.43)$ \\ $(R^2)$
    }&
    \makecell{1.73 \\ {\footnotesize \it (0.992)}}&
    \makecell{1.67 \\ {\footnotesize  \it (0.992)}}&
    \makecell{\second{1.49}\\ {\footnotesize  \it (0.982)}}&
    \makecell{0.51\\ {\footnotesize  \it (0.584)}}&
    \makecell{0.99\\ {\footnotesize  \it (0.98)}}&
    \makecell{\textbf{1.41}\\ {\footnotesize  \it(0.995)}}
    \\
    \midrule

    \textit{FID} $\downarrow$ &
    318&
    217&
    \textbf{186}&
    342&
    312&
    \second{193}
    \\
    \bottomrule

    \end{tabular}}
    \caption{Comparison of image naturalness metrics for different synthetic datasets. We report the FID, the KL divergence of the gradient distribution with respect to WaterlooDB, and the power spectrum slope $\gamma$ (with linear regression score $R^2$). VL achieves the best gradient-KL and power spectrum slope; its FID is second only to GTA-V.}\label{tab:similaritymetrics}
\end{table*}

\subsection{Comparison with other synthetic image datasets}

We compare the statistical similarity of the VL model to natural images with that of other synthetic datasets, using the three metrics introduced above. We consider:

 \begin{itemize}
    \item \textbf{CleVR}~\cite{johnson2017clevr}: abstract scenes of 3D volumes,
    \item \textbf{FractalDB}~\cite{kataoka2020pre}: binary fractal images for pre-training high-level vision models,
    \item \textbf{Textured DL}~\cite{baradad2021learning}: textured dead leaves, also used for pretraining classification models,
    \item \textbf{GTA-V}~\cite{richter2016playing}: photorealistic images of an urban video-game environment.
 \end{itemize}
 
% \begin{figure}[h]
%     \centering
%     \includegraphics[width = 0.23\linewidth]{images/figure_datasets/clevR.png}
%     \hfill
%     \includegraphics[width = 0.23\linewidth]{images/figure_datasets/gta.png}
%     \hfill
%     \includegraphics[width = 0.23\linewidth]{images/figure_datasets/fractalDB.png}
%     \hfill
%     \includegraphics[width = 0.23\linewidth]{images/figure_datasets/dl_text.jpg}

%     \caption{Examples of the other synthetic image datasets. From left to right: ClevR\cite{johnson2017clevr}, GTA-V\cite{richter2016playing}, FractalDB\cite{kataoka2020pre}, Textured DL\cite{baradad2021learning}}\label{fig:synth_datasets}
% \end{figure}

Like our model, the first three consist of abstract images with no semantic content; GTA-V was designed for photorealism and includes rich semantic information. These datasets have primarily been used for high-level tasks such as classification and segmentation.

We report naturalness metrics in Table~\ref{tab:similaritymetrics}. In addition to gradient-KL and power spectrum slope, we compute the Fréchet Inception Distance (FID) with respect to WaterlooDB\@. This metric compares deep feature representations from an Inception-V3 network pre-trained on ImageNet and is known to correlate well with human perception.

As shown in Table~\ref{tab:similaritymetrics}, VL achieves better gradient-KL and power spectrum profiles than all other synthetic datasets. While these statistics may matter less for image analysis tasks, they are critical for restoration, since preserving fine-scale structure is a key challenge. In terms of FID, VL outperforms the other abstract datasets by a large margin and remains close to GTA-V, despite the latter's photorealism and semantic content.

%Nonetheless, these datasets still show a significant statistical gap with natural images, which can be shortened with appropriate modeling.
In the next section, we report the advantages of the VL model for image restoration tasks.

\begin{table*}[hb]
\centering
\resizebox{\linewidth}{!}{
\setlength{\tabcolsep}{4pt}
\begin{tabular}{lccccccccccbb}
\toprule
& \multicolumn{2}{c}{Kodak24}
& \multicolumn{2}{c}{CBSD68}
& \multicolumn{2}{c}{McMaster}
& \multicolumn{2}{c}{Urban100}
& \multicolumn{2}{c}{Bokeh}
& \multicolumn{2}{b}{Average} \\
\cmidrule(lr){2-3}\cmidrule(lr){4-5}\cmidrule(lr){6-7}\cmidrule(lr){8-9}\cmidrule(lr){10-11}\cmidrule(lr){12-13}
& $\sigma=25$ & $\sigma=50$
& $\sigma=25$ & $\sigma=50$
& $\sigma=25$ & $\sigma=50$
& $\sigma=25$ & $\sigma=50$
& $\sigma=25$ & $\sigma=50$
& $\sigma=25$ & $\sigma=50$ \\
\cmidrule(lr){2-2}\cmidrule(lr){3-3}\cmidrule(lr){4-4}\cmidrule(lr){5-5}\cmidrule(lr){6-6}\cmidrule(lr){7-7}\cmidrule(lr){8-8}\cmidrule(lr){9-9}\cmidrule(lr){10-10}\cmidrule(lr){11-11}\cmidrule(lr){12-12}\cmidrule(lr){13-13}
Method
& PSNR / {\tiny LPIPS} & PSNR / {\tiny LPIPS}
& PSNR / {\tiny LPIPS} & PSNR / {\tiny LPIPS}
& PSNR / {\tiny LPIPS} & PSNR / {\tiny LPIPS}
& PSNR / {\tiny LPIPS} & PSNR / {\tiny LPIPS}
& PSNR / {\tiny LPIPS} & PSNR / {\tiny LPIPS}
& PSNR / {\tiny LPIPS} & PSNR / {\tiny LPIPS} \\
\midrule
Input
& \textit{20.43} / {\tiny \textit{--}} & \textit{14.86} / {\tiny \textit{--}}
& \textit{20.53} / {\tiny \textit{--}} & \textit{15.01} / {\tiny \textit{--}}
& \textit{20.91} / {\tiny \textit{--}} & \textit{15.38} / {\tiny \textit{--}}
& \textit{20.63} / {\tiny \textit{--}} & \textit{15.11} / {\tiny \textit{--}}
& \textit{20.46} / {\tiny \textit{--}} & \textit{14.95} / {\tiny \textit{--}}
& \textit{20.59} / {\tiny \textit{--}} & \textit{15.06} / {\tiny \textit{--}} \\

\midrule
$\text{DRUNet}_{\text{Fractal}}$ \cite{kataoka2020pre}
& 17.32 / {\tiny 0.6211} & 17.06 / {\tiny 0.7126}
& 17.05 / {\tiny 0.6319} & 16.66 / {\tiny 0.6968}
& 15.57 / {\tiny 0.6055} & 15.19 / {\tiny 0.6972}
& 16.20 / {\tiny 0.5319} & 15.93 / {\tiny 0.6037}
& 19.49 / {\tiny 0.7028} & 18.47 / {\tiny 0.8543}
& 17.13 / {\tiny 0.6186} & 16.66 / {\tiny 0.7129} \\

$\text{DRUNet}_{\text{ClevR}}$ \cite{johnson2017clevr}
& 30.42 / {\tiny 0.2447} & 27.71 / {\tiny 0.3506}
& 29.45 / {\tiny 0.2490} & 26.64 / {\tiny 0.3729}
& 30.98 / {\tiny 0.1429} & 28.15 / {\tiny 0.2230}
& 29.43 / {\tiny 0.1150} & 26.05 / {\tiny 0.2122}
& 37.13 / {\tiny 0.1534} & 33.91 / {\tiny 0.2206}
& 31.48 / {\tiny 0.1810} & 28.49 / {\tiny 0.2759} \\

$\text{DRUNet}_{\text{DL}}$ \cite{achddou2021synthetic}
& 30.95 / {\tiny 0.1559} & 28.09 / {\tiny 0.2549}
& 30.20 / {\tiny 0.1323} & 27.18 / {\tiny 0.2461}
& 31.25 / {\tiny 0.0951} & 28.32 / {\tiny 0.1621}
& 28.94 / {\tiny 0.0600} & 26.05 / {\tiny 0.1425}
& 36.66 / {\tiny 0.1766} & 33.76 / {\tiny 0.2499}
& 31.60 / {\tiny 0.1240} & 28.68 / {\tiny 0.2111} \\

$\text{DRUNet}_{\text{DL-Text}}$ \cite{baradad2021learning}
& 31.14 / {\tiny 0.1562} & 28.11 / {\tiny 0.2667}
& 30.35 / {\tiny 0.1369} & 27.18 / {\tiny 0.2560}
& 31.33 / {\tiny 0.1022} & 28.31 / {\tiny 0.1745}
& 29.26 / {\tiny 0.0626} & 25.81 / {\tiny 0.1441}
& 37.19 / {\tiny 0.1255} & 33.90 / {\tiny 0.2065}
& 31.85 / {\tiny 0.1167} & 28.66 / {\tiny 0.2096} \\

$\text{DRUNet}_{\text{GTAV}}$ \cite{richter2016playing}
& 32.14 / {\tiny 0.129} & \second{29.20} / {\tiny 0.215}
& 31.14 / {\tiny 0.116} & \second{28.06} / {\tiny 0.216}
& 32.43 / {\tiny 0.0968} & 29.47 / {\tiny 0.1653}
& 31.17 / {\tiny \second{0.0446}} & 27.90 / {\tiny 0.0974}
& 38.59 / {\tiny 0.1050} & 35.71 / {\tiny \second{0.1460}}
& 33.09 / {\tiny 0.0983} & 30.07 / {\tiny 0.1679} \\

$\text{DRUNet}_{\text{VL}}$
& \second{32.16} / {\tiny 0.1275} & 29.16 / {\tiny \second{0.2145}}
& \second{31.21} / {\tiny 0.1166} & \second{28.06} / {\tiny 0.2172}
& \second{32.63} / {\tiny \second{0.0885}} & \second{29.59} / {\tiny 0.1545}
& 31.27 / {\tiny \best{0.0425}} & 27.94 / {\tiny \second{0.0921}}
& \second{38.70} / {\tiny 0.1143} & \second{35.78} / {\tiny 0.1567}
& \second{33.19} / {\tiny \second{0.0979}} & \second{30.11} / {\tiny \second{0.1670}} \\

\rowcolor{gray!20}
$\text{DRUNet}_{\text{Nat}}$
& \best{32.89} / {\tiny \best{0.1078}} & \best{29.86} / {\tiny \best{0.1851}}
& \best{31.69} / {\tiny \best{0.0976}} & \best{28.51} / {\tiny \best{0.1834}}
& \best{33.14} / {\tiny \best{0.0770}} & \best{30.08} / {\tiny \best{0.1338}}
& \best{32.60} / {\tiny 0.0503} & \best{29.60} / {\tiny \best{0.0907}}
& \best{39.21} / {\tiny \best{0.1013}} & \best{36.31} / {\tiny \best{0.1425}}
& \cellcolor{blue!30!gray!20}\best{33.91} / {\tiny \best{0.0868}} & \cellcolor{blue!30!gray!20}\best{30.86} / {\tiny \best{0.1471}} \\
\midrule
$\text{FFDNet}_{\text{Fractal}}$ \cite{kataoka2020pre}
& 18.08 / {\tiny 0.6239} & 17.08 / {\tiny 0.7889}
& 17.98 / {\tiny 0.6147} & 17.01 / {\tiny 0.7521}
& 16.82 / {\tiny 0.5808} & 15.94 / {\tiny 0.7278}
& 17.44 / {\tiny 0.4307} & 16.50 / {\tiny 0.6259}
& 19.11 / {\tiny 0.8823} & 18.08 / {\tiny 1.0507}
& 17.89 / {\tiny 0.6265} & 16.92 / {\tiny 0.7891} \\

$\text{FFDNet}_{\text{ClevR}}$ \cite{johnson2017clevr}
& 30.14 / {\tiny 0.2265} & 27.50 / {\tiny 0.3357}
& 29.31 / {\tiny 0.2308} & 26.58 / {\tiny 0.3536}
& 30.05 / {\tiny 0.1425} & 27.47 / {\tiny 0.2218}
& 27.90 / {\tiny 0.1141} & 25.09 / {\tiny 0.2097}
& 36.24 / {\tiny 0.1574} & 33.40 / {\tiny 0.2154}
& 30.73 / {\tiny 0.1743} & 28.01 / {\tiny 0.2672} \\

$\text{FFDNet}_{\text{DL}}$ \cite{achddou2021synthetic}
& 30.91 / {\tiny 0.1508} & 28.02 / {\tiny 0.2577}
& 30.23 / {\tiny 0.1284} & 27.19 / {\tiny 0.2396}
& 31.10 / {\tiny 0.0905} & 28.18 / {\tiny 0.1656}
& 29.21 / {\tiny 0.0659} & 25.79 / {\tiny 0.1560}
& 36.36 / {\tiny 0.1779} & 33.45 / {\tiny 0.2703}
& 31.56 / {\tiny 0.1227} & 28.52 / {\tiny 0.2178} \\

$\text{FFDNet}_{\text{DL-Text}}$ \cite{baradad2021learning}
& 30.43 / {\tiny 0.2340} & 27.14 / {\tiny 0.3915}
& 29.76 / {\tiny 0.2044} & 26.41 / {\tiny 0.3637}
& 30.59 / {\tiny 0.1323} & 27.37 / {\tiny 0.2461}
& 28.41 / {\tiny 0.1074} & 24.85 / {\tiny 0.2419}
& 35.71 / {\tiny 0.1973} & 31.42 / {\tiny 0.3791}
& 30.98 / {\tiny 0.1751} & 27.44  / {\tiny 0.3245} \\

$\text{FFDNet}_{\text{GTAV}}$ \cite{richter2016playing}
& 31.46 / {\tiny 0.1390} & 28.52 / {\tiny 0.2337}
& 30.56 / {\tiny 0.1230} & 27.55 / {\tiny 0.2283}
& 31.30 / {\tiny 0.0991} & 28.47 / {\tiny 0.1664}
& 29.68 / {\tiny 0.0592} & 26.48 / {\tiny 0.1288}
& 37.39 / {\tiny 0.1111} & 34.29 / {\tiny 0.1815}
& 32.08 / {\tiny 0.1063} & 29.06 / {\tiny 0.1877} \\

$\text{FFDNet}_{\text{VL}}$
& 31.72 / {\tiny 0.1445} & 28.61 / {\tiny 0.2507}
& 30.85 / {\tiny 0.1311} & 27.68 / {\tiny 0.2419}
& 31.85 / {\tiny 0.0990} & 28.78 / {\tiny 0.1704}
& 30.18 / {\tiny 0.0537} & 26.73 / {\tiny 0.1235}
& 37.85 / {\tiny 0.1177} & 34.43 / {\tiny 0.1859}
& 32.49 / {\tiny 0.1092} & 29.24 / {\tiny 0.1945} \\

\rowcolor{gray!20}
$\text{FFDNet}_{\text{Nat}}$
& 32.13 / {\tiny \second{0.1238}} & 28.98 / {\tiny 0.2175}
& \second{31.21} / {\tiny \second{0.1135}} & 27.96 / {\tiny \second{0.2126}}
& 32.35 / {\tiny 0.0900} & 29.18 / {\tiny \second{0.1539}}
& \second{31.40} / {\tiny 0.0639} & \second{28.05} / {\tiny 0.1168}
& 38.28 / {\tiny \second{0.1048}} & 35.05 / {\tiny 0.1482}
& \cellcolor{blue!30!gray!20}33.07 / {\tiny 0.0992} & \cellcolor{blue!30!gray!20}29.84 / {\tiny 0.1698} \\

\bottomrule
\end{tabular}
}

\caption{Image denoising results. We report PSNR~($\uparrow$) and LPIPS~($\downarrow$) of two denoising networks (DRUNet\cite{zhang2021plug} and FFDNet\cite{zhang2018ffdnet}) trained on either natural images or synthetic images (FractalDB~\cite{kataoka2020pre}, ClevR~\cite{johnson2017clevr}, GTAV~\cite{richter2016playing}, Dead Leaves~\cite{achddou2021synthetic}, DL-Text~\cite{baradad2021learning}, or \DLplus). The models are tested on several image denoising benchmarks. Best results are in \best{bold} and second results are \second{underlined}. }
\label{tab:denoising}
\end{table*}

\section{Deep Image Restoration Experiments}\label{sec:experiments}
In \cite{achddou_cviu}, classic DL images were used to train image denoising and super-resolution NNs, obtaining promising but sub-par performance compared to state-of-the-art (SOTA) models trained on natural images.
In this section, we show that the new VL model
%in terms of physical and statistical modeling
leads to significant gains in image restoration performance, closing the gap with models trained on natural images across standard architectures for denoising and single-image super-resolution. 
Additional experiments on image inpainting and image deblurring are provided in Section~\ref{sec:supp_restoration}.

Next, we show how the denoising networks can be leveraged as an implicit prior for solving linear inverse problems in Plug-and-Play algorithms.

\vspace{-0.3cm}

\subsection{Additive White Gaussian Noise Removal}\label{sec:denoising}
\noindent \textbf{Training Details.}
We train two network architectures: FFDNet \cite{zhang2018ffdnet}, a lightweight fully convolutional model, and DRUNet \cite{zhang2021plug}, the current state of the art in image denoising.

For both architectures, we assess the impact of the training data distribution on the denoising performance. Thus, we train multiple versions of the same network on the individual synthetic datasets presented in Table~\ref{tab:similaritymetrics}, i.e., FractalDB\cite{kataoka2020pre}, ClevR\cite{johnson2017clevr}, GTAV\cite{richter2016playing}, Dead Leaves \cite{achddou2021synthetic}, and DL-textured\cite{baradad2021learning}, as well as on natural images. Our natural baselines are trained on a dataset composed of Waterloo DB \cite{ma2017waterloo}, DIV2K \cite{Agustsson_2017_CVPR_Workshops} and Flickr database \cite{lim2017enhanced}, totaling roughly 9K images, following the procedure presented in the original DRUNet paper~\cite{zhang2021plug}.

To ensure a similar volume of training data across all experiments, we synthesize $10K$ images of size $(500\times500)$ with our VibrantLeaves model and extract 500k patches of size $(128\times128)$ from each dataset, as proposed in \cite{tassano2019analysis}.
% We generated 500k patches of size $(128\times128)$ with the Dead-Leaves ++ and the original Dead-Leaves models and extracted the same amount of images from the GTA-V dataset.
% For the synthetic training, for both the DL and VL models, we generated 500k patches of size $(128\times128)$ with the VL model and the classic DL models.
We follow the training procedures of \cite{zhang2018ffdnet} and \cite{zhang2021plug}, minimizing the $\mathcal{L}_2$ and $\mathcal{L}_1$ losses respectively, with an ADAM optimizer and a learning rate decaying from $10^{-4}$ to $10^{-6}$.

\noindent \textbf{Testing datasets.}
We test our models on several classic image denoising benchmarks: Kodak24\cite{franzen1999kodak}, CBSD68\cite{martin2001database}, McMaster\cite{zhang2011color} and Urban100\cite{huang2015single}. The first three test sets mix natural and man-made environments, while Urban100 consists of urban architectural photographs exhibiting repetitive patterns at different scales.
To evaluate the importance of depth-of-field modeling, we also test our models on the Bokeh validation set\cite{ignatov2020rendering}, comprising images with shallow depth-of-field.

\begin{figure*}[t]
    \centering
    \resizebox{\textwidth}{!}{
    \setlength{\tabcolsep}{1pt}
    \begin{tabular}{ccccc}
    Ground Truth& Noisy $\sigma = 25$ & $\text{DRUNet}_{\text{DL}}$ & $\Dvl$ & $\Dnat$ \\
    \Plotimagespy[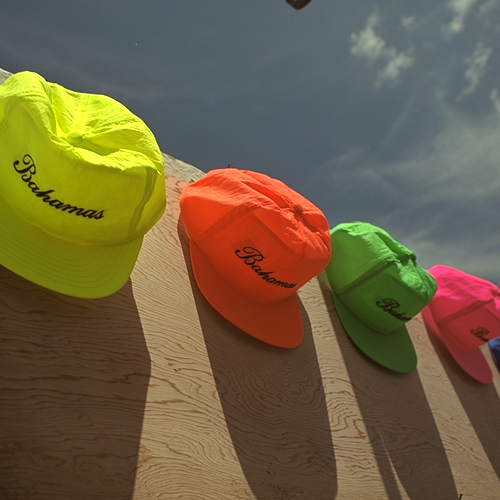]{2.2}{0.65}{0.18\textwidth}{0.4}[4]
    &
    \Plotimagespy[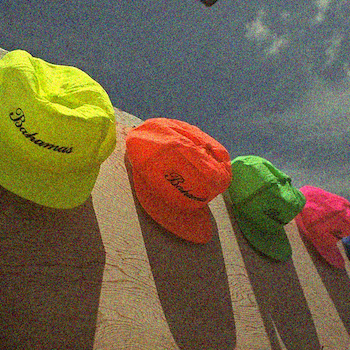]{2.2}{0.65}{0.18\textwidth}{0.4}[4]
    &
    \Plotimagespy[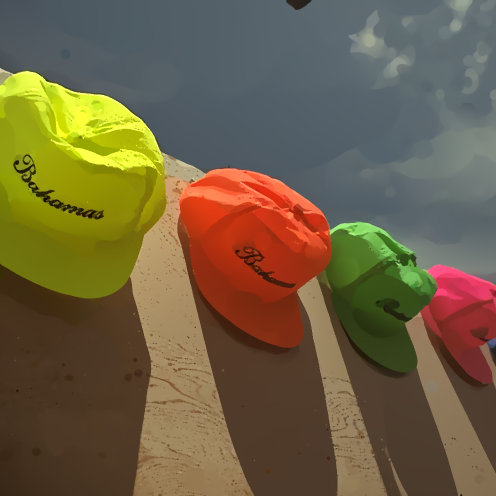]{2.2}{0.65}{0.18\textwidth}{0.4}[4]
    &
    \Plotimagespy[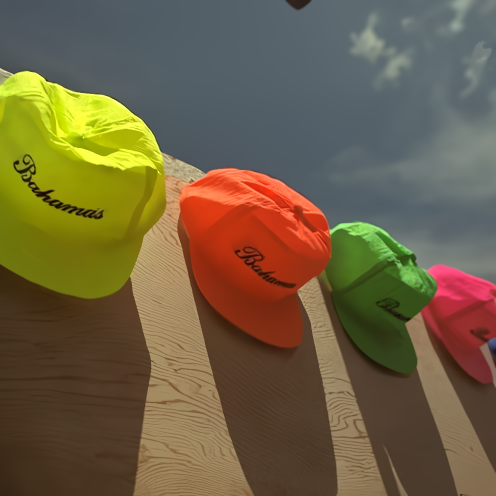]{2.2}{0.65}{0.18\textwidth}{0.4}[4]
    &
    \Plotimagespy[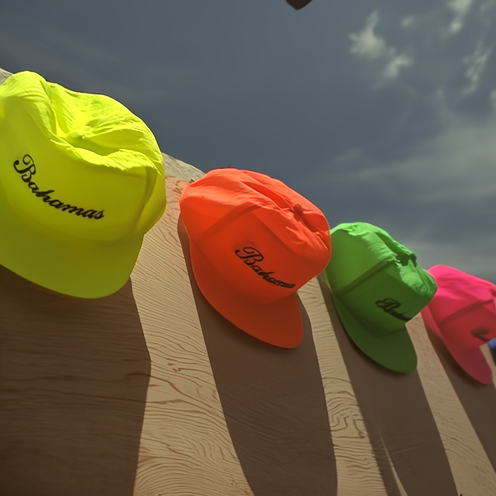]{2.2}{0.65}{0.18\textwidth}{0.4}[4]
    \\
    \Plotimagespy[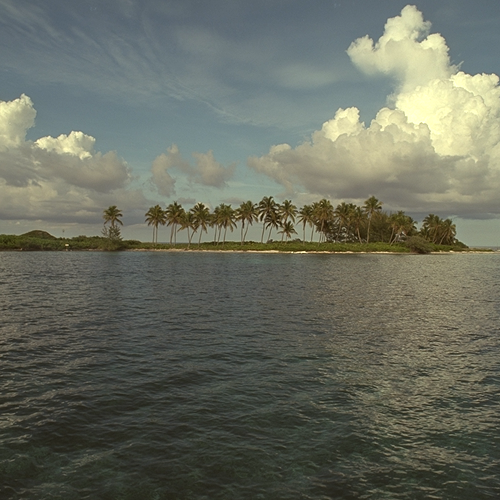]{2.7}{2.35}{0.18\textwidth}{0.4}[4]
    &
    \Plotimagespy[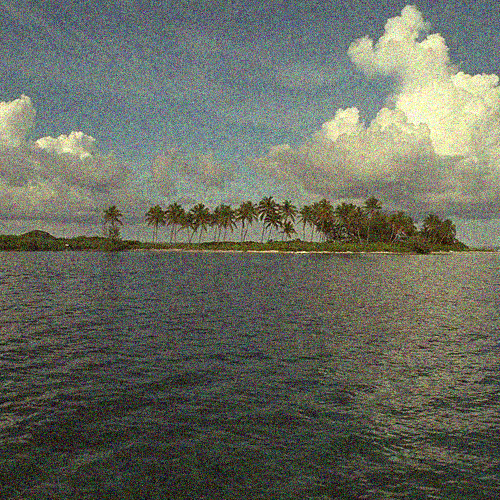]{2.7}{2.35}{0.18\textwidth}{0.4}[4] 
    &
    \Plotimagespy[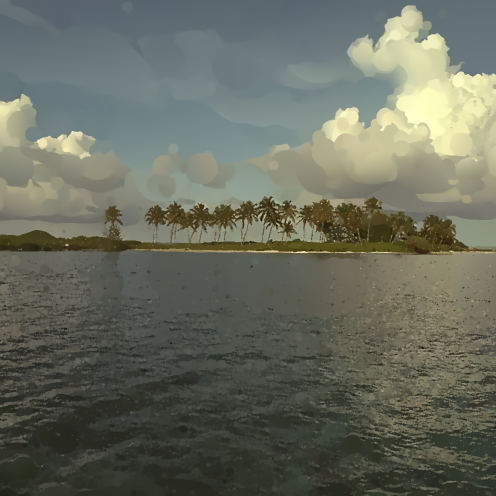]{2.7}{2.35}{0.18\textwidth}{0.4}[4] 
    &
    \Plotimagespy[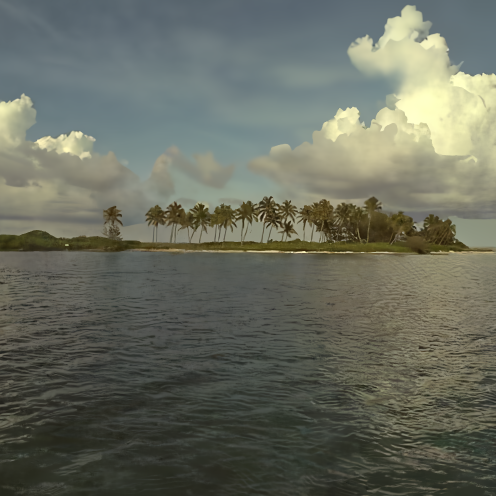]{2.7}{2.35}{0.18\textwidth}{0.4}[4] 
    &
    \Plotimagespy[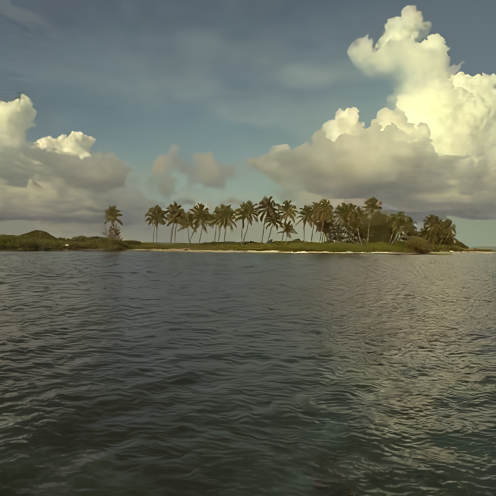]{2.7}{2.35}{0.18\textwidth}{0.4}[4] 
    \end{tabular}}

\caption{Visual results for AWGN denoising (best viewed on a digital screen). We compare the same DRUNet architecture trained either on Dead Leaves \cite{achddou2021synthetic}, \DLplus, or natural images.}\label{fig:denoising}
\end{figure*}

\noindent \textbf{AWGN removal results.}
We report numerical results in Table~\ref{tab:denoising} and visual comparisons in Figure~\ref{fig:denoising}.
For both architectures, VL training substantially improves over classic DL, reducing the average PSNR gap with natural-image-trained networks from $2.2$\,dB to $0.72$\,dB.
Moreover, VL images compare favorably to other synthetic training sets: the margin over abstract procedural models is significant (more than 1.4\,dB over DL-text). Compared to GTA5 images --- which comprise realistic urban scenes rendered from a 3D world model with semantic information and design biases --- \DLplus~achieves comparable denoising performance. Instead, the benefits of \DLplus~over GTA5 lie in its compactness, controllability, and reduced semantic content, which are a step towards better interpretability.
Note that the average PSNR gap is partly explained by lower performance on Urban100: this dataset contains perfectly repetitive content that does not exactly match our semi-periodic texture model, which introduces random phase and amplitude variations.

This quantitative improvement correlates with the visual quality of the denoised results shown in Figure~\ref{fig:denoising}.
The examples show much better restoration of semi-periodic textures (first example) and micro-textures (second example).
The model trained on DL tends to combine small circles to reconstruct complex structures from the original image.
We further investigate these training-induced biases in Section~\ref{sec:synthetic_bias}, where we probe each model at an extreme noise level to make the hallucinations inherited from each training distribution explicit.
A more in-depth analysis of each component of the VL model is presented in Section~\ref{sec:ablations}.

Interestingly, switching FFDNet for DRUNet when training on DL images yields only marginal improvements, whereas the architectural gain is greater for other training sets, including VL.
This suggests that performance is upper-bounded when training with the DL model, which we interpret as a lack of expressiveness of this image prior.
In contrast, we do not observe this saturation with the VL model, suggesting that VL images are sufficiently expressive to fully exploit the capacity of state-of-the-art architectures.

In order to assess generalization ability of synthetic training, we also tested our denoising models on out-of-distribution datasets comprising medical images and satellite images. The results, reported in Section~\ref{sec:supp_cross}, exhibit smaller PSNR gaps than for natural image testsets. 

\subsection{Single-Image Super-resolution (SISR)}

\noindent \textbf{Training Details.}
We use the Swin-IR lightweight model~\cite{liang2021Swinir} for SISR at scales $\times 2$ and $\times 4$. We train these networks with images from either the Dead Leaves model \cite{achddou2021synthetic} or the \DLplus~model. More precisely, we generate 10K patches of size $(256\times256)$ and downscale them using MATLAB's bicubic downsampler (standard in SISR benchmarks) at scales $\times 2$ and $\times 4$. We train these networks following the optimization schedule and code provided by the authors of~\cite{liang2021Swinir}.  

\noindent \textbf{Testing datasets.}
We test our models on several classic SISR benchmarks: Set5\cite{franzen1999kodak}, Set14\cite{zeyde2012single}, and the DIV2K\cite{Agustsson_2017_CVPR_Workshops} validation set. Note that the DIV2K training set was used for training the original SwinIR model.

\noindent \textbf{SISR results.} We report numerical results in Table~\ref{tab:sisr} and visual results in Figure~\ref{fig:superres}.
In terms of PSNR, the gap between $\text{Swin-IR}_{\text{VL}}$ and $\text{Swin-IR}_{\text{Nat}}$ averages 0.75\,dB for scale $\times 2$ and 0.60\,dB for scale $\times 4$, while the corresponding gap for $\text{Swin-IR}_{\text{DL}}$ is 2.28\,dB and 1.51\,dB respectively.
\begin{table}[h]
    \centering
\resizebox{\columnwidth}{!}
{\setlength{\tabcolsep}{6pt}
\begin{tabular}{llcccb}
\toprule
Scale & Method
& Set5 \cite{franzen1999kodak}
& Set14 \cite{zeyde2012single}
& DIV2K \cite{Agustsson_2017_CVPR_Workshops}
& \textit{Average} \\
\cmidrule(lr){3-3}\cmidrule(lr){4-4}\cmidrule(lr){5-5}\cmidrule(lr){6-6}
& & PSNR / {\tiny LPIPS} & PSNR / {\tiny LPIPS} & PSNR / {\tiny LPIPS} & PSNR / {\tiny LPIPS} \\
\midrule
\rowcolor{gray!20} \multirow{3}{*}{x2}
& $\text{Swin-IR}_{\text{Nat}}$ & \best{38.14} / {\tiny \second{0.0553}} & \best{33.67} / {\tiny \second{0.0869}} & \best{36.51} / {\tiny \second{0.0823}} & \textit{\best{36.11} / {\tiny \second{0.0748}}} \\
& $\text{Swin-IR}_{\text{VL}}$  & \second{37.39} / {\tiny 0.0560} & \second{33.19} / {\tiny 0.0910} & \second{35.51} / {\tiny 0.0951} & \textit{\second{35.36} / {\tiny 0.0807}} \\
& $\text{Swin-IR}_{\text{DL}}$  & 35.27 / {\tiny \best{0.0362}} & 32.03 / {\tiny \best{0.0703}} & 34.19 / {\tiny \best{0.0737}} & \textit{33.83 / {\tiny \best{0.0601}}} \\
\midrule
\rowcolor{gray!20} \multirow{3}{*}{x4}
& $\text{Swin-IR}_{\text{Nat}}$ & \best{32.34} / {\tiny \second{0.1750}} & \best{28.44} / {\tiny \second{0.2823}} & \best{30.65} / {\tiny 0.2650} & \textit{\best{30.48} / {\tiny \second{0.2408}}} \\
& $\text{Swin-IR}_{\text{VL}}$  & \second{31.65} / {\tiny 0.1835} & \second{27.92} / {\tiny 0.2919} & \second{30.08} / {\tiny 0.2914} & \textit{\second{29.88} / {\tiny 0.2556}} \\
& $\text{Swin-IR}_{\text{DL}}$  & 30.27 / {\tiny \best{0.1502}} & 27.39 / {\tiny \best{0.2512}} & 29.25 / {\tiny \best{0.2489}} & \textit{28.97 / {\tiny \best{0.2168}}} \\
\bottomrule
\end{tabular}}
\caption{Results for Single-Image Super-Resolution. The models are tested on several SISR benchmarks. Best results are in \best{bold} and second results are \second{underlined}.}
    \label{tab:sisr}
\end{table}

Note that the gap is larger on the DIV2K test set, which is very similar to the training set of $\text{Swin-IR}_{\text{Nat}}$.
These performance gains translate to better image quality, as illustrated in Figure~\ref{fig:superres}. Both examples show that $\text{Swin-IR}_{\text{VL}}$ corrects two major defects of $\text{Swin-IR}_{\text{DL}}$: staircasing artifacts (first example) and the inability to restore line structures (second example). The results of $\text{Swin-IR}_{\text{VL}}$ appear almost as sharp as those of $\text{Swin-IR}_{\text{Nat}}$.

Despite a PSNR gap of 2.28\,dB ($\times 2$) and 1.51\,dB ($\times 4$) behind $\text{Swin-IR}_{\text{Nat}}$, $\text{Swin-IR}_{\text{DL}}$ achieves the lowest LPIPS scores across all benchmarks. LPIPS indeed rewards outputs with sharp, high-frequency content, which the model trained on Dead Leaves images reproduces due to the abundance of sharp discontinuities in the training set. Consequently, LPIPS does not penalise the staircasing artifacts visible in Figure~\ref{fig:superres}, and these scores should be interpreted with caution.
% without having ever seen natural images during training.
\begin{figure*}[ht]
        \centering
        \resizebox{\textwidth}{!}{
        \setlength{\tabcolsep}{1pt}
        \begin{tabular}{ccccc}
            High Res.& Low Res. $\downarrow 4$ & $\text{Swin-IR}_{\text{DL}}$ & $\text{Swin-IR}_{\text{VL}}$ & $\text{Swin-IR}_{\text{Nat}}$ \\
        \Plotimagespy[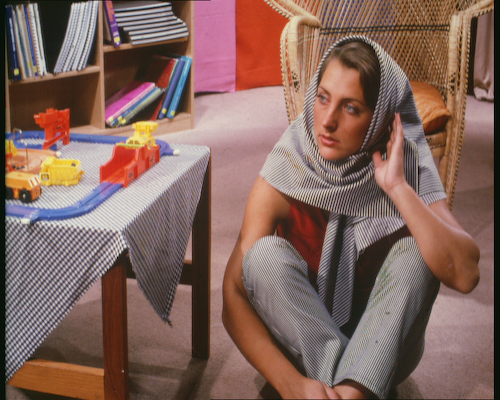]{2.2}{1.9}{0.18\textwidth}{0.4}[4]
        &
        \Plotimagespy[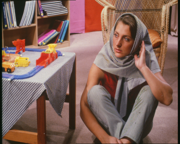]{2.2}{1.9}{0.18\textwidth}{0.4}[4]
        %{}
        &
        \Plotimagespy[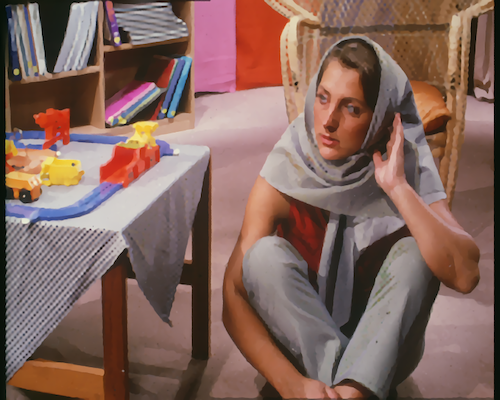]{2.2}{1.9}{0.18\textwidth}{0.4}[4]
        %{Swin-IR(DL)}
        &
        \Plotimagespy[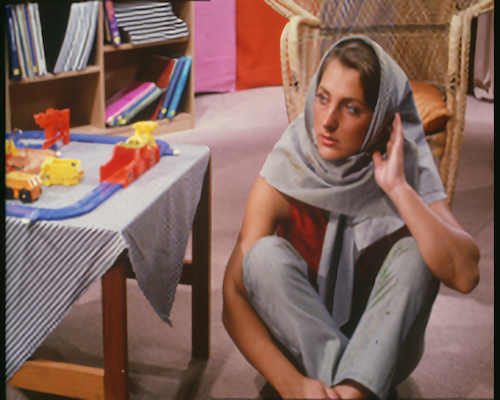]{2.2}{1.9}{0.18\textwidth}{0.4}[4]
        %{Swin-IR(VL)}
        &
        \Plotimagespy[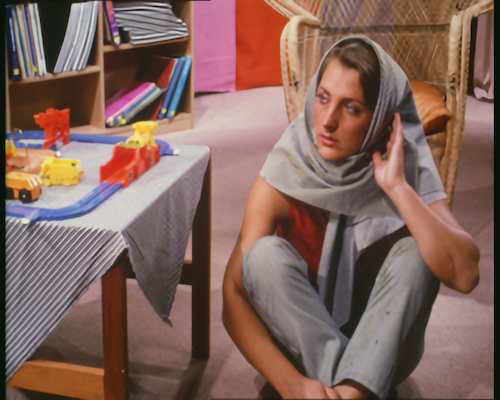]{2.2}{1.9}{0.18\textwidth}{0.4}[4]
        %{Swin-IR(Nat)}
        \\
        \Plotimagespy[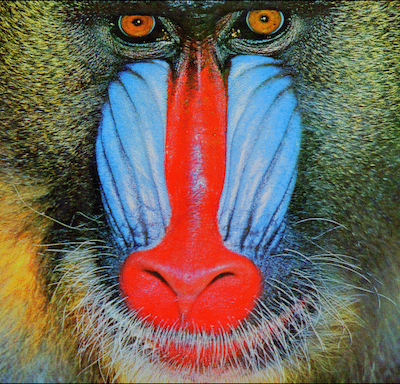]{2.2}{0.6}{0.18\textwidth}{0.4}[4]  
        &
        \Plotimagespy[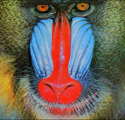]{2.2}{0.6}{0.18\textwidth}{0.4}[4]  
        &
        \Plotimagespy[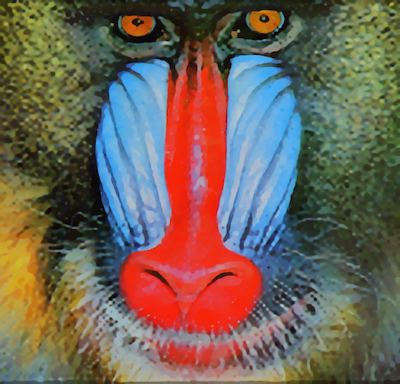]{2.2}{0.6}{0.18\textwidth}{0.4}[4] 
        &
        \Plotimagespy[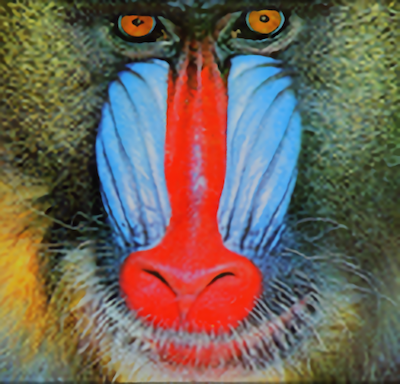]{2.2}{0.6}{0.18\textwidth}{0.4}[4]  
        &
        \Plotimagespy[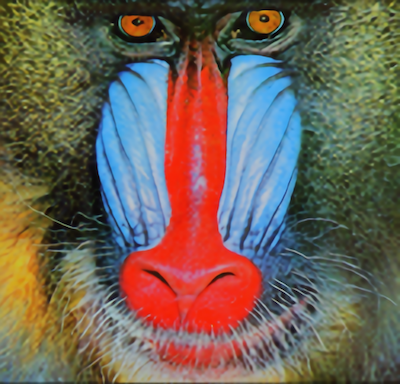]{2.2}{0.6}{0.18\textwidth}{0.4}[4] 
        \end{tabular}}

    \caption{Visual results for Single-Image-Super-Resolution (best viewed on a digital screen). We compare the same SwinIR architecture trained either on Dead Leaves \cite{achddou2021synthetic}, \DLplus, or natural images. The results suggest that Swin-IR(DL) creates staircasing artifacts (first row) and cannot recreate linear objects and textures (second row).}\label{fig:superres}
\end{figure*}

\subsection{Plug-and-Play for Inverse Problems}
\noindent \textbf{Background on Plug-and-Play (PnP) methods}
Many image restoration problems can be formulated with the linear inverse problem setup presented in Eq.~\eqref{eq:ip}, for which $\Phi(x) = Ax$ where $A$ is assumed to be known. Optimization-based algorithms for inverse problems can be derived from the MAP estimator formulation Eq.~\eqref{eq:MAP}, by replacing the term $g(x)$ with a regularization term \cite{rudin1992nonlinear,DavidL.Donoho1994}. Among these optimization algorithms, variable splitting descent methods such as HQS or ADMM have emerged as practical solvers. The problem can be rewritten as $\arg \min f(x)+g(z) \text{ s.t } z=x$, where the algorithms alternate between a descent step on $f$ and a descent step on $g$.
For such algorithms, Venkatakrishnan et al. \cite{venkatakrishnan2013plug} first showed that the descent step on the prior $g$ is equivalent to a denoising step.
Thus, any denoising algorithm, and in particular deep denoisers \cite{zhang2021plug}, can be used to replace this descent step, while the other step admits a closed-form solution (at least for Gaussian noise).
This allows any pre-trained denoiser to solve a wide variety of image restoration tasks without retraining. This is because deep denoisers encode richer regularity models than a hand-crafted $g(x)$.

We assess the quality of our VL image generator by training a denoiser on VL images and plugging it into a PnP setup. We compare results using either $\text{DRUNet}_{\text{VL}}$ or $\text{DRUNet}_{\text{Nat}}$ as the prior, for both deblurring and inpainting. We use the DPIR algorithm~\cite{zhang2021plug}, based on HQS with an ad-hoc noise scheduler, as our PnP solver.
\\
\noindent \textbf{Image Deblurring.}
The associated inverse problem takes the form:
\begin{equation}
    y = k * x + n,
\end{equation}
where $k$ is a motion blur kernel from the kernel set of Levin et al.~\cite{levin2009understanding}, and $n \sim \mathcal{N}(0,\sigma^2)$ with $\sigma = 7/255$. We test three of the proposed kernels on Set14.
We report PSNR results for deblurring in the first 3 columns of Table~\ref{tab:pnp}. The gap between $\text{DPIR}_{\text{VL}}$ and $\text{DPIR}_{\text{Nat}}$ is in the range $[0.53, 0.60]$\,dB, while the gap with $\text{DPIR}_{\text{DL}}$ is in the range $[1.98, 2.38]$\,dB. 
In terms of LPIPS, $\text{DPIR}_{\text{VL}}$ closely tracks $\text{DPIR}_{\text{Nat}}$ (gap of $[0.008, 0.014]$), both substantially outperforming $\text{DPIR}_{\text{DL}}$. Visually, $\text{DPIR}_{\text{DL}}$ produces disk-like artifacts, while $\text{DPIR}_{\text{VL}}$ correctly restores textures and object boundaries.
\\

\begin{table}[hb]
    \centering
    \caption{Plug-and-Play numerical results. Tests are performed on the dataset  Set14\cite{zeyde2012single}, with a Gaussian noise of std $\sigma = 7$ for deblurring. Best results are in\textbf{ bold}, second are \second{underlined}.}
    \resizebox{\columnwidth}{!}{
    \begin{tabular}{rcccgg}
    \toprule
         \textit{Problem}& \multicolumn{3}{c}{Deblurring} & \multicolumn{2}{g}{Inpainting} \\
         \cmidrule(lr){2-4}\cmidrule(lr){5-6}
         \textit{Prior / level}& $k_0$ & $k_1$ & $k_2$ & $p = 50\%$ & $p = 70\%$ \\
         \cmidrule(lr){2-2}\cmidrule(lr){3-3}\cmidrule(lr){4-4}\cmidrule(lr){5-5}\cmidrule(lr){6-6}
         & PSNR / {\tiny LPIPS} & PSNR / {\tiny LPIPS} & PSNR / {\tiny LPIPS} & PSNR / {\tiny LPIPS} & PSNR / {\tiny LPIPS} \\
         \midrule
         $\text{DPIR}_{\text{DL}}$& 28.02 / {\tiny 0.1942} & 27.88 / {\tiny 0.1976} & 27.64 / {\tiny 0.2093} & 29.58 / {\tiny 0.1002} & 26.33 / {\tiny 0.1686}\\
         $\text{DPIR}_{\text{VL}}$& \second{29.56} / {\tiny 0.1769} & \second{29.28} / {\tiny 0.1886} & \second{29.49} / {\tiny 0.1950} & \second{31.03} / {\tiny 0.0756} & \second{26.89} / {\tiny 0.1510}\\
         $\text{DPIR}_{\text{Nat}}$& \best{30.16} / {\tiny 0.1630} & \best{29.86} / {\tiny 0.1762} & \best{30.02} / {\tiny 0.1848} & \best{31.64} / {\tiny 0.0700} & \best{27.29} / \best{\tiny 0.1395}\\
         \bottomrule
    \end{tabular}}
    
    \label{tab:pnp}
\end{table}

\begin{figure}[hb]
    \centering
    \resizebox{\columnwidth}{!}
    {
    \setlength{\tabcolsep}{1pt}
    \begin{tabular}{cccccc}
    \toprule 
    $x$&
    $y$&
    $\text{DPIR}_{\text{DL}}$
    &
    $\text{DPIR}_{\text{VL}}$
    &
    $\text{DPIR}_{\text{Nat}}$ \\
    \midrule 
        \Plotimagespy[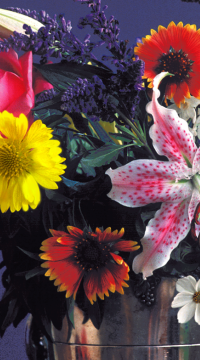]{1.1}{1.68}{0.18\columnwidth}{0.75}[3]&

    \Plotimagespy[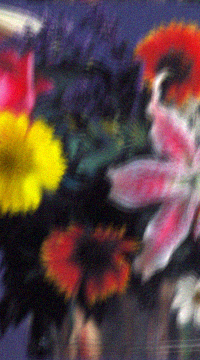]{1.1}{1.68}{0.18\columnwidth}{0.75}[4]&
    \Plotimagespy[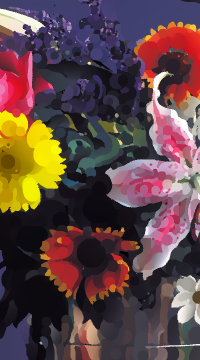]{1.1}{1.68}{0.18\columnwidth}{0.75}[3]&
    \Plotimagespy[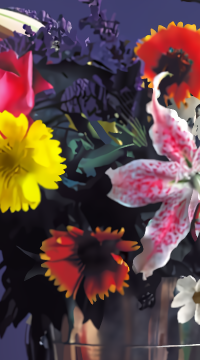]{1.1}{1.68}{0.18\columnwidth}{0.75}[3]&
    \Plotimagespy[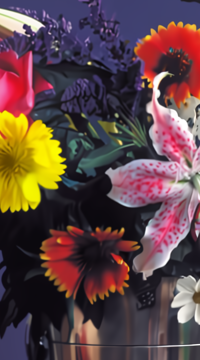]{1.1}{1.68}{0.18\columnwidth}{0.75}[3]\\
    \multicolumn{5}{c}{Deblurring} \\
    \midrule
    \Plotimagespy[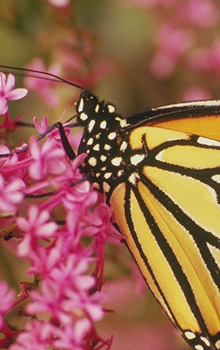]{0.35}{1.35}{0.18\columnwidth}{0.5}[3]&

    \Plotimagespy[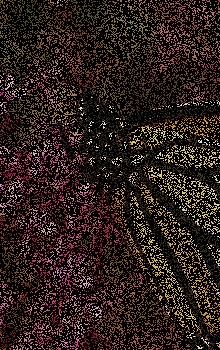]{0.35}{1.35}{0.18\columnwidth}{0.5}[3]&
    \Plotimagespy[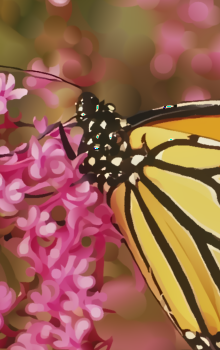]{0.35}{1.35}{0.18\columnwidth}{0.5}[3]&
    \Plotimagespy[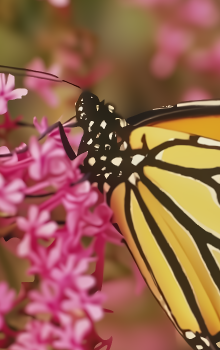]{0.35}{1.35}{0.18\columnwidth}{0.5}[3]&
    \Plotimagespy[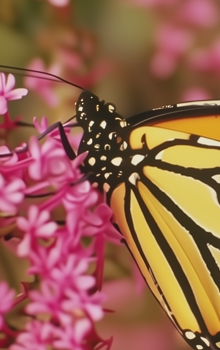]{0.35}{1.35}{0.18\columnwidth}{0.5}[3] \\
    \multicolumn{5}{c}{Inpainting} \\
    \bottomrule
    \end{tabular}}
    \caption{Plug-and-Play Deblurring and Inpainting. We compare three different priors in the DPIR framework: $\text{DRUNet}_{\text{DL}}$, $\text{DRUNet}_{\text{VL}}$ and $\text{DRUNet}_{\text{Nat}}$. Perceptually, results from $\text{DRUNet}_{\text{VL}}$ and $\text{DRUNet}_{\text{Nat}}$ are very similar. Best viewed zoomed in.}    
    \label{fig:placeholder}
\end{figure}

\noindent \textbf{Image Inpainting.}
In this case the linear degradation operator zeroes out $p\%$ of pixels, leaving the others unchanged. The problem is then to recover the missing pixels (we do not consider additional additive noise). Results are reported in Table~\ref{tab:pnp}: VL training achieves a PSNR gap of $[0.40, 0.61]$\,dB with respect to natural training. 
The LPIPS gap between $\text{DPIR}_{\text{VL}}$ and $\text{DPIR}_{\text{Nat}}$ remains small ($[0.006, 0.012]$), confirming perceptual similarity despite the PSNR difference. Visually, $\text{DPIR}_{\text{VL}}$ and $\text{DPIR}_{\text{Nat}}$ produce perceptually similar reconstructions in the inpainted regions, as illustrated in Figure~\ref{fig:placeholder}.

\section{Invariances and Ablation Studies}
\subsection{Invariance analysis of the trained models}\label{sec:invariances}

In the previous section, we have shown that natural images can be replaced by the proposed VL synthetic images for training restoration NNs, with limited performance loss. In this section, we show that training on VL images also yields better generalization ability of the resulting networks.  
%Despite the advances in modeling  natural images properties in a synthetic image generator, a performance gap remains when testing our models on the same distribution of images.
Many studies have indeed shown that neural networks are often over-specialized to the training distribution \cite{plotz2017benchmarking,hein2019relu}, leading to poor generalization capability to slightly out-of-distribution samples.
Such samples can be obtained by distorting the test data with operations such as contrast manipulation, rotation, or downscaling.
In order to deal with these issues, one could either augment training data with these distortions or develop neural architectures that are invariant to these specific distortions \cite{cohen2016group}.
However these solutions fall short when the model is confronted with unseen distortions.

Conversely, the VL model is intrinsically invariant to some transformations including scaling, rotation and contrast distortions.
In the following experiments, we show that these invariance properties are inherited by the NNs trained on VL images, contrary to NNs trained on natural images, which lack such properties. 
Note that bias-free denoising NNs~\cite{kadkhodaie2021stochastic} such as the original DRUNet architecture, i.e., architectures with no additive bias terms, are 1-homogeneous and therefore exactly invariant to a global \textit{scalar} rescaling of pixel intensity, i.e., a basic form of contrast invariance.

This scalar-intensity invariance is distinct from robustness to the more complex, nonlinear S-curve contrast transformation used in Section~\ref{sec:additional_robustness}, which 1-homogeneity does not guarantee; we assess our model's robustness to this distortion along with other non-geometric ones such as blur and hue shift.
%Our models show better robustness to several distortions.

\subsubsection{Protocol}
We compare $\Dnat$ and $\Dvl$ on Kodak24 images first distorted at various levels and then corrupted with noise at $\sigma = 25$.
For each type of distortion, we test our denoising NNs across a range of distortion levels.
To assess how distortion affects denoising performance, we define for each denoiser $f$ the PSNR difference induced by distortion $g$ at level $\beta$:
$$\Delta_{f}(g,\beta) = \text{PSNR}(f(g(X,\beta),25)) - \text{PSNR}(f(X,25)) ,$$
where $g(\cdot,\beta)$ is the distortion function and $X$ are the Kodak images.
Note that $\Delta_f \leq 0$ in general, since distortion degrades denoising performance; a value closer to $0$ indicates greater robustness.
When comparing two denoisers $(f_1,f_2)$, $\Delta_{f_1}>\Delta_{f_2}$ means that model $f_1$ is more robust to the tested distortion than model $f_2$.
%Having $\Delta_{\text{DRUNet}_{\text{Nat}}}(g,\beta) < \Delta_{\text{DRUNet}_{\text{VL}}}(g,\beta)$ means that our version of DRUNet is more robust to distortion $g$ at level $\beta$.
We therefore also report the evolution of $$\mathbf{\Delta}(g,\beta) = \Delta_{\Dnat}(g,\beta) - \Delta_{\Dvl}(g,\beta)$$ with respect to $\beta$, where a positive value indicates that $\Dnat$ is more sensitive to the distortion than $\Dvl$.

The distortions we consider are the following:
\begin{itemize}
    % \item \textit{Contrast:} $\tilde{I} = 0.5 +\frac{(I-I_{\min})}{\beta(I_{\max}-I_{\min})}$
    \item \textit{Rotation:} $g_{\text{rot}}(X,\theta)$, implemented using a linear interpolation, which results in a small loss of high frequency,
    %= \text{Rotate}(I,\theta)$, which is implemented in openCV. The rotation involves 
    %a linear interpolation, which results in a small loss of high frequencies,
    \item \textit{Scale:} $g_{\text{scale}}(X,\gamma)  = \text{Bicubic}\downarrow_{\gamma}\left[G_{\sigma(\gamma)}*X\right]$, where $\sigma(\gamma)$ is  a linear function that sets the standard deviation of the gaussian kernel depending on the downscaling factor, to ensure no aliasing.
\end{itemize}

\subsubsection{Results}
In Figure~\ref{fig:invariances}, we report $\Delta_{f}(g,\beta)$ for both distortions in separate plots. In each plot, we report $\Delta_{\Dnat}(g,\beta)$ in black and $\Delta_{\Dvl}(g,\beta)$ in red. We also report $\mathbf{\Delta}(g,\beta)$ for each distortion.

For both distortions, the red curve is always above the black curve, meaning that $\Dvl$ is more robust to these distortions.

$\mathbf{\Delta}(g_{\text{scale}},\gamma)$ increases monotonically with the distortion level, consistent with the VL model's exact scale invariance.
On the other hand, $\mathbf{\Delta}(g_{\text{rot}},\theta)$ has a profile symmetric around $\theta = 45^{\circ}$ and reaches its maximum there. This is because $\Dnat$ was trained with images rotated by $\theta \in \{90^{\circ}, 180^{\circ}, 270^{\circ}\}$, making $g_{\text{rot}}(\cdot, 45^{\circ})$ the rotation that maximizes the distance from the training distribution.

Overall, these experiments suggest that training with VL images improves the robustness of the resulting model to geometric distortions.
Embedding invariances directly into the training data appears to be a more practical approach than modifying the network architecture \cite{cohen2016group}.
The VL model offers a principled way to achieve this, while maintaining high restoration performance.

\begin{figure*}
    \centering
    \begin{tikzpicture}
        \tikzstyle{every node}=[font=\small]
        \begin{axis}[
            width=0.35\textwidth, % Scale the plot to \linewidth
            height=0.2\textheight, % Scale the plot to \linewidth
            grid=major, 
            grid style={dashed,gray!30}, % Set the labels
            xlabel= \footnotesize{rotation angle $\theta$ in degrees},
            ylabel = \footnotesize{$\Delta_{f}(g_{\text{rot}},\theta)$},
            xtick={5,45,85},
            label style={font=\footnotesize},
            tick label style={font=\footnotesize},
            y tick label style={/pgf/number format/.cd, fixed, fixed zerofill,precision=2}
            % x tick label style={rotate=90,anchor=east}
          ]
          \addplot[mark=none,thick, color=black] table[x=angle,y=psnr_nat,col sep=comma] {csv_files/rotation_1.csv};
          \addplot[mark=none,thick, color=red] table[x=angle,y=psnr_vl,col sep=comma] {csv_files/rotation_1.csv};
          \coordinate (subplot) at (86,1.8);
        \end{axis}
        
    \begin{axis}[
            height=0.15\textheight,
            at=(subplot),
            % title = \footnotesize{$\mathbf{\Delta}(g_{\text{rot}},\theta)$},
            grid=major,
            grid style={dashed,gray!30},
            legend style={at={(0.7,0.95)},anchor=north},
            axis background/.style={fill=gray!10},
            yticklabel style={
                    /pgf/number format/fixed,
                    /pgf/number format/precision=1,
                    /pgf/number format/fixed zerofill
                },
            label style={font=\tiny},
            tick label style={font=\tiny},
            ytick={0,0.15},
            xtick={0,45,90},
            % ylabel = \footnotesize{$\mathbf{\Delta}(g_{\text{rot}},\theta)$},
            % log ticks with fixed point,
            style={font=\tiny}% Set the labels
    ]
        \addplot[mark=none,thick, color=black] table[x=x,y=y,col sep=comma] {csv_files/gap_rotation.csv};
        \node[draw=black,fill=white,anchor=south west] at (rel axis cs:0,0) {\scriptsize{$\mathbf{\Delta}(g_{\text{rot}},\theta)$}};
        % \node[draw=black,fill=white,anchor=south west] at (rel axis cs:0,0) {;
        \end{axis}
    \end{tikzpicture}
    \hfill
    \begin{tikzpicture}
        \tikzstyle{every node}=[font=\small]
        \begin{axis}[
            legend pos=outer north east,
            % (so the legend looks a bit better)
            legend cell align=left,
            % (moved this common key here)
            smooth,
            width=0.4\textwidth, % Scale the plot to \linewidth
            height=0.2\textheight, % Scale the plot to \linewidth
            grid=major, 
            grid style={dashed,gray!30}, % Set the labels
            xlabel= \footnotesize{Scaling factor $\gamma$},
            ylabel = \footnotesize{$\Delta_{f}(g_{\text{scale}},\gamma)$},
            % x tick label style={rotate=90,anchor=east}
            ]
            \addplot[mark=none,thick, color=black] table[x=scale,y=psnr_nat,col sep=comma] {csv_files/scale.csv};
            \addplot[mark=none,thick, color=red] table[x=scale,y=psnr_vl,col sep=comma] {csv_files/scale.csv};
            \coordinate (subplots) at (4,-0.7);
        %     \legend{
        %     \footnotesize{$\Dnat$},
        %     \footnotesize{$\Dvl$},
        % }
        \end{axis}
    \begin{axis}[
            height=0.15\textheight,
            at=(subplots),
            % title = \footnotesize{$\mathbf{\Delta}(g_{\text{rot}},\theta)$},
            grid=major,
            grid style={dashed,gray!30},
            legend style={at={(0.7,0.95)},anchor=north},
            axis background/.style={fill=gray!10},
            yticklabel style={
                    /pgf/number format/fixed,
                    /pgf/number format/precision=1,
                    /pgf/number format/fixed zerofill
                },
            label style={font=\tiny},
            tick label style={font=\tiny},
            ytick={0,0.15},
            % ylabel = \footnotesize{$\mathbf{\Delta}(g_{\text{rot}},\theta)$},
            % log ticks with fixed point,
            style={font=\tiny}% Set the labels
    ]
        \addplot[mark=none,thick, color=black] table[x=x,y=y,col sep=comma] {csv_files/gap_scale.csv};
        \node[draw=black,fill=white,anchor=south west] at (rel axis cs:0,0) {\scriptsize{$\mathbf{\Delta}(g_{\text{scale}},\gamma)$}};
        % \node[draw=black,fill=white,anchor=south west] at (rel axis cs:0,0) {;
        \end{axis}
    \end{tikzpicture}
    \caption{Invariance experiments. These plots show the PSNR difference between $\Dnat$(in black) and $\Dvl$(in red) tested at $\sigma = 25$ on slight modifications of the Kodak24 dataset: rotations with an angle  $\theta \in[5,85]$ degrees, and downscaling with an increasing factor $\gamma\in[1,4.5]$.}\label{fig:invariances}
\end{figure*}

\vspace{-0.3cm}
\subsection{Ablation Study}\label{sec:ablations}

To assess the contribution of each component of \DLplus, we perform an ablation study on the generation algorithm.
We generate 10K $(512\times 512)$ images, removing one or more properties of the proposed \DLplus model.
We consider the following settings:
\begin{enumerate}
    \Myitem{dlplus} \textbf{\DLplus}: all properties combined.
    \Myitem{depth}  \textbf{Without Depth}:  we remove depth-of-field modeling and perspective.
    \Myitem{shapes} \textbf{Without complex shapes}: images are generated with disks only.
    \Myitem{microt} \textbf{Without micro-textures }: the textures are sampled from the semi-periodic texture generator only.
    \Myitem{sin} \textbf{Without periodic textures}: the textures are sampled from our micro-texture generator only.
    \Myitem{text}  \textbf{Without textures}: the shapes are generated with constant colors.
    \Myitem{depth_text} \textbf{Without depth and textures}: the random shapes are generated without textures and without depth-of-field simulation.
    \Myitem{dead_leaves} \textbf{Dead Leaves}: the original model from \cite{achddou_cviu}.
\end{enumerate}

For each of these settings, we train a DRUNet denoising network using identical training hyperparameters. We test the trained models on three datasets: Kodak24, BokehDB and Urban100.
We report numerical evaluations in Figure~\ref{fig:ablation_numbers} and visual comparisons in the Figure~\ref{fig:ablation_visual}.

\noindent{\textbf{Depth.}} While removing depth-of-field and perspective leads to only a minor performance drop on Kodak24, there is a notable gap on BokehDB, which contains images with limited depth-of-field.
To isolate this effect from texture, we compare models \Circle{text} and \Circle{depth_text}, which share the same absence of texture modeling: removing depth-of-field simulation alone causes a gap of more than 1dB on BokehDB.
Visually, the denoised images exhibit strong artifacts in blurry regions, where the model tends to hallucinate sharp boundaries, as shown in the first row of Figure~\ref{fig:ablation_visual}.
Additionally, the performance gap between models \Circle{depth} and \Circle{dlplus} is somewhat larger for Urban100 than for Kodak24, as Urban100 mostly contains architectural scenes with strong linear perspective.

\noindent{\textbf{Textures.}}
Texture modeling contributes significantly to the performance of \DLplus.
Removing semi-periodic textures (model \Circle{sin}) leads to a notable drop on Urban100, as these scenes feature repetitive patterns that require such textures; model \Circle{sin} struggles to reproduce them, as shown in the second row of Figure~\ref{fig:ablation_visual}.
Removing micro-textures (model \Circle{microt}), on the other hand, has limited impact on quantitative metrics, but causes noticeable qualitative artifacts: grass patterns are reproduced as distorted sinusoidal structures, as visible in the third row of Figure~\ref{fig:ablation_visual}.
Removing both texture types leads to a significant performance drop (model \Circle{text}), confirming that both components contribute independently to overall performance.

\noindent{\textbf{Shapes.}} The performance gap between models \Circle{shapes} and \Circle{dlplus} is relatively small, which might suggest that shape modeling is not crucial.
However, comparing models \Circle{dead_leaves} and \Circle{depth_text} — which share the absence of texture and depth modeling, isolating the effect of shapes — reveals that complex shapes alone yield significant PSNR improvements (see Figure~\ref{fig:ablation_numbers}).
This is further supported by the larger gap between \Circle{shapes} and \Circle{dlplus} on Urban100: this dataset mostly contains straight boundaries and right-angled corners, which the disk-based generator of model \Circle{shapes} cannot reproduce.
\begin{figure*}[hbp]
    \centering
    \begin{minipage}[c]{0.8\linewidth}
        \begin{tikzpicture}
            \centering
            \begin{axis}[%
                xmin=0.8, xmax=3.5,
                ymin=-2.1, ymax=0.4,
                width=\linewidth,
                height=0.27\textheight,
                grid=major,
                grid style={dashed,gray!30},
                ylabel=PSNR gap of the ablated models,
                xticklabels={$\sigma = 25$,$\sigma = 50$,$\sigma = 25$,$\sigma = 50$,$\sigma = 25$,$\sigma = 50$},
                xtick={1,1.2,2,2.2,3,3.2},
                x tick label style={rotate=60,anchor=east,font = 
                \footnotesize},
                label style={font=\footnotesize},
                tick label style = {font=\footnotesize},
                ]
                \addplot[
                    scatter,%
                    scatter/@pre marker code/.code={%
                        \edef\temp{\noexpand\definecolor{mapped color}{rgb}{\pgfplotspointmeta}}%
                        \temp
                        \scope[draw=mapped color!50!black,fill=mapped color]%
                    },%
                    scatter/@post marker code/.code={%
                        \endscope
                    },%
                    only marks,     
                    mark=*,
                    mark size=2.5pt,
                    opacity=0.9,
                    point meta={TeX code symbolic={%
                        \edef\pgfplotspointmeta{\thisrow{RED},\thisrow{GREEN},\thisrow{BLUE}}%
                    }},
                ] 
                table {
                x   y   RED   GREEN   BLUE
                1   0  1   1   1
                1   -0.06  0.847    0.106    0.376
                1   -0.14  0.118    0.533    0.898
                1   -0.2  1    0.757    0.027
                1   -0.28  0.    0.302    0.251
                1   -0.89  0.984    0.424    0.698
                1   -1.03 0.431     0.875    0.812
                1   -1.11 0.408     0.322    0.365
                1.2   0  1   1   1
                1.2   -0.06  0.847    0.106    0.376
                1.2   -0.12  0.118    0.533    0.898
                1.2   -0.18  1    0.757    0.027
                1.2   -0.28  0.    0.302    0.251
                1.2   -0.78  0.984    0.424    0.698
                1.2   -0.82 0.431     0.875    0.812
                1.2   -1.05 0.408     0.322    0.365
                2   0  1   1   1
                2   -0.03  0.847    0.106    0.376
                2   -0.3  0.118    0.533    0.898
                2   -0.39  1    0.757    0.027
                2   -0.12  0.    0.302    0.251
                2   -0.52  0.984    0.424    0.698
                2   -1.52 0.431     0.875    0.812
                2   -2.03 0.408     0.322    0.365
                2.2   0  1   1   1
                2.2   -0.03  0.847    0.106    0.376
                2.2   -0.47  0.118    0.533    0.898
                2.2   -0.34  1    0.757    0.027
                2.2   -0.1  0.    0.302    0.251
                2.2   -0.51  0.984    0.424    0.698
                2.2   -1.52 0.431     0.875    0.812
                2.2   -2.0 0.408     0.322    0.365
                3   0  1   1   1
                3   -0.16  0.847    0.106    0.376
                3   -0.22  0.118    0.533    0.898
                3   -0.18  1    0.757    0.027
                3   -0.59  0.    0.302    0.251
                3   -1.39  0.984    0.424    0.698
                3   -1.31 0.431     0.875    0.812
                3   -1.74 0.408     0.322    0.365
                3.2  0  1   1   1
                3.2   -0.12  0.847    0.106    0.376
                3.2   -0.18  0.118    0.533    0.898
                3.2   -0.13  1    0.757    0.027
                3.2   -0.67  0.    0.302    0.251
                3.2   -1.15  0.984    0.424    0.698
                3.2   -1.10 0.431     0.875    0.812
                3.2   -1.71 0.408     0.322    0.365

                };
            \node[anchor=center] at (1.,0.1) { \footnotesize\textit{32.16}};
            \node[anchor=west] at (1.2,0.1) {\footnotesize \textit{29.15}};
            \node[anchor=center] at (2.,0.1) {\footnotesize\textit{38.69}};
            \node[anchor=west] at (2.2,0.1) {\footnotesize\textit{35.76}};
            \node[anchor=center] at (3.,0.1) {\footnotesize\textit{31.17}};
            \node[anchor=west] at (3.2,0.1) {\footnotesize\textit{27.76}};
            \node[anchor=center] at (1.1,0.3) {\bfseries  Kodak24 \cite{franzen1999kodak}};
            \node[anchor=center] at (2.1,0.3) {\textbf{Bokeh} \cite{ignatov2020rendering}};
            \node[anchor=center] at (3.1,0.3) {\bfseries Urban100 \cite{huang2015single}};
            \end{axis};
            % \node[anchor=center] at (1.7,0.25\textheight) {\bfseries Kodak24};
            % \node[anchor=center] at (7.1,0.25\textheight) {\bfseries Bokeh dB};
            % \node[anchor=center] at (12.5,0.25\textheight) {\bfseries Urban100};
            \end{tikzpicture}
    \end{minipage}
    \begin{minipage}[c]{0.18\linewidth}
        \begin{tikzpicture}
            \centering
            \begin{axis}[%
                xmin=0, xmax=5,
                ymin=0, ymax=2.2,
                width=\linewidth,
                height=0.27\textheight,
                axis lines=none
                ]
                \addplot[
                    scatter,%
                    scatter/@pre marker code/.code={%
                        \edef\temp{\noexpand\definecolor{mapped color}{rgb}{\pgfplotspointmeta}}%
                        \temp
                        \scope[draw=mapped color!80!black,fill=mapped color]%
                    },%
                    scatter/@post marker code/.code={%
                        \endscope
                    },%
                    only marks,     
                    mark=*,
                    mark size=2.5pt,
                    opacity=0.9,
                    point meta={TeX code symbolic={%
                        \edef\pgfplotspointmeta{\thisrow{RED},\thisrow{GREEN},\thisrow{BLUE}}%
                    }},
                ] 
                table {
                x   y   RED   GREEN   BLUE
                0.1   2  1    1    1
                0.1   1.8  0.847    0.106    0.376
                0.1   1.6  0.118    0.533    0.898
                0.1   1.4  1    0.757    0.027
                0.1   1.2  0.    0.302    0.251
                0.1   1  0.984    0.424    0.698
                0.1   0.8 0.431     0.875    0.812
                0.1   0.6 0.408     0.322    0.365
                };
                \node[anchor=west] at (0.1,2) {\scriptsize \bfseries VL};
                \node[anchor=west] at (0.1,1.8) {\scriptsize w/o shapes};
                \node[anchor=west] at (0.1,1.6) {\scriptsize w/o Depth};
                \node[anchor=west] at (0.1,1.4) { \scriptsize w/o micro-T};
                \node[anchor=west] at (0.1,1.2) {\scriptsize w/o periodic-T};
                \node[anchor=west] at (0.1,1.) {\scriptsize w/o Textures};
                \node[anchor=west] at (0.1,0.8) {\scriptsize w/o Text. \& Depth};
                \node[anchor=west] at (0.1,0.6) {\scriptsize Dead Leaves \cite{achddou_cviu}};
            \end{axis}
            % \node[anchor=center] at (1.7,0.25\textheight) {\bfseries Kodak24};
            % \node[anchor=center] at (7.1,0.25\textheight) {\bfseries Bokeh dB};
            % \node[anchor=center] at (12.5,0.25\textheight) {\bfseries Urban100};
            \end{tikzpicture}
        
    \end{minipage}

        \caption{Ablation study - Numerical results. We report the PSNR gap of the ablated models with respect to the results of DRUNet trained on \DLplus. We test each model on two different noise values ($\sigma \in \{25,50\}$) and different natural image datasets with different properties: Kodak24 \cite{franzen1999kodak}, BokehDB\cite{ignatov2020rendering}, Urban100\cite{huang2015single}. The score on top of each column refers to the PSNR of DRUNet trained on \DLplus.}
        \label{fig:ablation_numbers}
\end{figure*}

\section{Conclusion and perspectives}

In a previous work, \cite{achddou2021synthetic}, we had shown that Dead Leaves images were a promising replacement of natural images for the training of image restoration NNs, despite a significant performance gap. 
In this paper, we introduce \DLplus (VL), a principled parametric image model based on the Dead Leaves framework which further incorporates three key image properties: complex geometry, texture modeling and physical depth. On two restoration tasks, networks trained on VL images perform close to the same architectures trained on natural images: the average PSNR gap is 0.72\,dB for denoising and 0.75\,dB ($\times 2$) / 0.60\,dB ($\times 4$) for single-image super-resolution, substantially narrowing the gap left by the original Dead Leaves model. 
We further evaluate the VL model on deblurring and inpainting in Section~\ref{sec:supp_restoration}, where it again approaches natural-image training.
We also study the contribution of each component of the model in a detailed ablation study, each property contributing to restoration performance. 
We believe that such controlled experiments help identify which image properties make a good prior for natural images, offering some insight into deep restoration methods. 
Last, we report experiments suggesting that synthetic training can improve robustness, in particular to geometric distortions, by leveraging the invariance properties built into the training set. 
Further demonstrating and exploiting such generalization ability would be a natural follow-up of this work. 

Another asset of the proposed approach is that the biases of the trained model are directly tied to the explicitly defined properties of the synthetic training set. 

Indeed, models trained on large, weakly curated natural-image datasets may hallucinate content that is not supported by the input, a failure mode of particular concern in satellite or biomedical imaging.
As a preliminary step toward investigating such biases, one can probe the model outside its training range: testing at very high noise levels forces the network to rely entirely on its learned prior, making training-induced hallucinations explicit, as we illustrate in Section~\ref{sec:synthetic_bias}.
A complementary perspective would be to draw samples from the model's implicit prior \cite{kadkhodaie2021stochastic}, making the learned image statistics directly observable.
We also expect that the proposed approach can offer additional insight into how deep restoration architectures operate. 
Since the VL model depends on only a few parameters, one could further improve interpretability by back-propagating through them, adapting attribution methods to recover which parameter settings best account for a given restoration.

\putbib[bibliography]

\clearpage

\setcounter{equation}{0}
\setcounter{figure}{0}
\setcounter{table}{0}
\setcounter{page}{1}
\makeatletter
\renewcommand{\theequation}{S\arabic{equation}}
\renewcommand{\thefigure}{S\arabic{figure}}
\renewcommand{\thetable}{S\arabic{table}}
\renewcommand{\thesection}{S\arabic{section}}
\renewcommand{\thepage}{S\arabic{page}}
% \renewcommand{\bibnumfmt}[1]{[S#1]} % This line requires natbib
% \renewcommand{\citenumfont}[1]{S#1} % This line requires natbib

% \begin{bibunit}[IEEEtran]

\begin{center}
    { \large \bfseries  Supplementary Material: VibrantLeaves}
\end{center}

\section{Background on the Dead Leaves Image model}\label{sec:background}

The dead leaves model is a random field obtained as the sequential superimposition of random shapes. 
It is defined (see ~\cite{matheron1968modele, bordenave2006dead}) from a set of random positions, times and shapes $\{(x_i,t_i,X_i)_{i\in\mathbb{N}}$, where $\mathcal{P}=\sum \delta_{x_i,t_i}$ is a homogeneous Poisson process on $\mathbb{R}^2\times (-\infty,0]$ and the $X_i$ are random sets of $\mathbb{R}^2$ that are independent of $\mathcal{P}$. 
The sets $x_i+X_i$ are called {\it leaves} and for each $i$, the {\it visible part} of the leaf is defined as 
$$V_i=(x_i+X_i)\setminus\bigcup_{t_j\in(t_i,0)}(x_j+X_j),$$
that is, the visible part of the leaf $(x_i,t_i,X_i)$ is obtained by removing from this leaf all leaves that are indexed by a time greater than $t_i$ (that falls after it). 
The dead leaves model is then defined as the collection of all visible parts. 
A random image can be obtained by assigning a random gray level (or color) to each visible part. 
% An example of a dead leaves model where the leaves are disks with a constant radius can be seen in Figure \ref{fig:r_fix}.

A particular type of dead leaves model, where the leaves have a size distribution with scaling properties, has been shown to reproduce important statistical properties of natural images~\cite{alvarez1999size,lee2001occlusion}. 
Such models are obtained by considering random leaves $R.X$, where $X$ is a given shape and $R$ is a real random variable with density $f(r)=C.r^{-\gamma}$, with $C$ a normalizing constant. 
The case $\gamma=3$ corresponds to a scale invariant model~\cite{lee2001occlusion}. 
In order for such models to be well defined, values of $R$ have to be restricted to values in $(r_{\min},r_{\max})$, see \cite{gousseau2007modeling} for a detailed mathematical analysis. The resulting model therefore depends on 3 parameters: $r_{\min}$, $r_{\max}$ and $\gamma$. 
This model is especially appealing for natural images, because it incorporates in a very simple setting two of their most fundamental property: non Gaussianity (as a result of edges) and scaling properties~\cite{mumford2001stochastic}. 
Because this model contains details and edges at all scales, potentially of arbitrary contrast, it has been proposed as a tool for the evaluation of the ability of imaging devices to faithfully capture textures\cite{cao2010dead}  and was thereafter retained as a standard for image quality evaluation\cite{ISO}. 

Dead leaves images were first employed for deep image restoration in \cite{achddou2021synthetic}, where the authors identify several critical factors contributing to the success of their method, including the size distribution, a downscaling operation as a basic image acquisition model, and a color sampling algorithm. For each generated dead leaves image, colors are sampled from the color histogram of a randomly selected image from a large set of natural images. We stress here that is is important to sample colors from a single natural image each time, and not from a global color model, which would results in a too strong homogeneity and little variety (see \cite{achddou2021synthetic}). It was later shown that these histograms could be approximated using a sophisticated parametric model \cite{achddou2023learning} without compromising performance. For simplicity, we sample colors directly from natural images for the remainder of this work.

\section{VibrantLeaves Implementation Details}\label{sec:sup_implem}

\subsection{VibrantLeaves function}

The {\bfseries\ttfamily{VL}} function first calls the histogram sampling function, which picks a random color image in the Waterloo database \cite{ma2017waterloo}.
It then calls the {\bfseries \ttfamily{LeavesStack}} function three times to generate a background, a middle-ground and a foreground stack.
Each of these calls are generated with the same parameters except the coverage percentage $p$ which we chose to be $p_b = 100\%$ for the background, $p_m = 50\%$ for the middleground and $p_f = 25\%$ for the foreground.
These are then merged according to the depth-of-field method presented in Section~\ref{sec:dof}. The standard deviations of each blur kernels are chosen equal ($\sigma_1 = \sigma_2$) and are sampled from a power law  $p(\sigma_1) = C.\mathbf{1}_{[0,10]}(\sigma_1).\sigma_1^{0.5}$ of exponent $0.5$ so that larger blur values appear more scarcely than average or small blur values. The image is then downscaled to ensure no aliasing artifacts caused by the artificially sharp edges of the synthetized shapes. This operation is performed with a bilinear downscaling of $2$ and respects Shannon sampling conditions, by first blurring with an appropriate Gaussian kernel. 
The parameters of the function {\bfseries\ttfamily{VL}} are summarized in Table~\ref{tab:paramsVL}.

\begin{table}[h]
    \centering
    \begin{tabular}{gcccgggc}
        \toprule
        Size & \multicolumn{3}{c}{Coverage}&\multicolumn{3}{g}{DoF}& Post-Proc.
        \\
         $w$ &$p_b$&$p_m$&$p_f$&$\sigma_{\min}$&$\sigma_{\max}$ & $\alpha_{\text{blur}}$  & $\downarrow$ \\
         \midrule
         1024& 100 & 50 & 25 & 0 & 10 & 0.5 & 2\\
         \bottomrule
    \end{tabular}
    \caption{Parameters of \ttfamily{VibrantLeaves}} 
    \label{tab:paramsVL}
\end{table}

\subsection{The {\ttfamily{LeavesStack}} function}
This function performs the same operations iteratively until a portion $p$ of the image plane is covered.
\textit{First,} the $(x,y)$ positions of the shapes are sampled uniformly at random in the image plane.

\textit{Second,} we generate a random shape with the {\ttfamily{SampleShape}} function of radius $r$  sampled with a clipped power law density $f$ which follows$$f(r) \sim C.\mathds{1}_{[r_{\min},r_{\max}]}(r).r^{-\alpha},$$ where $C$ is a normalizing constant. The radius parameters are the following: $r_{\min} = 10$ and $r_{\max} = 512$.
%, in order to obtain a reasonable amount of smaller and larger shapes.

% To maintain scale invariance, we choose $\alpha = 3$. We also choose $r_{\min} = 10$ and $r_{\max} = w$, where $w = 500$ is the generated image's width, so we get a reasonable amount of smaller and larger shapes.
\textit{Third,} we generate a texture map with the {\ttfamily{SampleTexture}} function. 
Finally, we multiply the shape mask with this texture map to obtain our colored shape, and update the corresponding pixels of the current image.
The parameters of the function {\ttfamily{LeavesStack}} are summarized in Table~\ref{tab:paramsLS}.
\begin{table}[h]
    \centering
    \begin{tabular}{cgc}
        \toprule
         $r_{\min}$&$r_{\max}$&$\alpha$\\
         \midrule
         10&512&3\\
         \bottomrule
    \end{tabular}
    \caption{Parameters of \ttfamily{LeavesStack}} 
    \label{tab:paramsLS}
\end{table}

\subsection{The {\ttfamily{SampleShape}} function}

This function samples a random shape following the principles presented in Section~\ref{sec:geometry}. 
To generate a random shape of radius $r$, we sample $n$ points uniformly in a disc $D_r$. For shape diversity, we sample the $\alpha$-shape parameter uniformly in $[0.2,0.6]$ (larger values lead to regular shapes that are close to the convex hull of the random points). Smoothing is applied to half the shapes, with a Gaussian kernel of standard deviation sampled uniformly in $[1,10]$ (values over 10 often cause disconnected shapes, as a result of the rough approximation of the curvature motion employed). In addition to these shapes, we also sample disks and rectangles.
The parameters of the function {\ttfamily{SampleShape}} are summarized in Table~\ref{tab:paramshapes}.

\begin{table}[h]
    \centering
 {\setlength{\tabcolsep}{3pt}
    \begin{tabular}{gcggccc}
        \toprule
        Nb of points & Concavity & \multicolumn{2}{g}{Smoothing} & \multicolumn{3}{c}{Proportions}\\ 
        $n$ & $\alpha$ & $\sigma_s$ & $p_{\text{smooth}}$& $p_{\text{poly}}$& $p_{\text{rectangle}}$& $p_{\text{disks}}$\\
         \midrule
        $\mathcal{U}([10,100])$& $\mathcal{U}([0.2,0.6])$ & $\mathcal{U}([1,10])$ & 1/2 &2/3&1/6&1/6\\
         \bottomrule
    \end{tabular}}
    \caption{Parameters of \ttfamily{SampleShape}} 
    \label{tab:paramshapes}
\end{table}

\subsection{The {\ttfamily{SampleTexture}} function}

The proportion of pseudo-periodic texture is chosen to be $1/6$, the one of micro-texture is $2/3$ and the one of 2-scale texture is $1/6$.
The reason for this is that having too many sinusoidal textures in the data set leads to the reproduction of sinusoidal artifacts in the denoised images.
For micro-textures, the slope of the power spectrum is sampled uniformly at random in $[0.5,2.5]$ to get a similar amount of smooth and harsh micro-textures.
For pseudo-periodic patterns and interpolation masks, we chose $T_{\min} = 5$ and $T_{\max} = 100$, to get short and long oscillations. The sharpening parameter of the logit function is sampled with: $\lambda \sim \mathcal{U}([1,10])$. As mentioned in the manuscript, we also apply a thresholding operations to all pseudo-periodic textures, with a threshold $\tau_{\text{texture}} \sim \mathcal{U}([-1,1])$. The atmospheric disturbance presented in Section~\ref{sec:macrotext} is applied to half of these textures, and so is the perspective transform presented in Section~\ref{sec:perspective}. The parameters of the function {\ttfamily{SampleTexture}} are summarized in Table~\ref{tab:paramsTextures}.
Note that the ranges and proportions chosen throughout this paragraph were not tuned to match any evaluation benchmark, nor otherwise optimized; they were simply set to reasonable values based on visual inspection of the resulting images. We leave the optimization of these ranges and proportions to future work.

\begin{table}[h]
    \centering
    {\setlength{\tabcolsep}{4pt}
    \begin{tabular}{gggcccg}
        \toprule
        \multicolumn{3}{g}{Proportions} & \multicolumn{3}{c}{Periodic text.}& Colored noise \\
        $p_{\text{p.p.}}$ & $p_{\text{c.n.}}$ & $p_{\text{b.t.}}$ & $T$ & $\lambda$ & $\tau_{\text{texture}}$ & $\gamma$ \\
         \midrule
         1/6&2/3&1/6 & $\mathcal{U}([5,100])$ & $\mathcal{U}([1,10])$ & $\mathcal{U}([-1,1])$ &$\mathcal{U}([0,5,2.5])$\\
         \bottomrule
    \end{tabular}}
    \caption{Parameters of \ttfamily{SampleTexture}} 
    \label{tab:paramsTextures}
\end{table}

\section{Additional Image Restoration Results}\label{sec:supp_restoration}

We extend the image restoration experiments of Section~\ref{sec:experiments} to two additional degradation tasks: inpainting for random masks and motion deblurring. In contrast to the Plug-and-Play setup used in the main manuscript, we here train two dedicated blind DRUNets end-to-end — one per task — without providing an explicit noise-level map at inference.

\subsection{Degradation Models}

\noindent\textbf{Bernoulli Inpainting.}
The observed image is obtained by randomly zeroing out pixels independently:
\begin{equation}
    y = M \odot x, \quad M_i \overset{\text{i.i.d.}}{\sim} \mathrm{Bernoulli}(1-p),
\end{equation}
where $p \in [30\%, 80\%]$ is the masking ratio and $\odot$ denotes the element-wise product.

\noindent\textbf{Motion Deblurring.}
The observed image is formed by convolving the clean image with a motion blur kernel drawn from the set of Levin et al.~\cite{levin2009understanding}, using mirror boundary conditions, combined with a small amount of additive Gaussian noise:
\begin{equation}
    y = k * x + n, \quad n \sim \mathcal{N}\!\left(0,\sigma_{\min}^2 I\right),
\end{equation}
where $k$ is a motion blur kernel from~\cite{levin2009understanding} and $\sigma_{\min} \in  [0,10/255]$ is a minimal noise level.

\subsection{Blind DRUNet}

We train two separate blind DRUNets~\cite{zhang2021plug}, one per task, without providing the degradation level as an explicit input channel; for deblurring, neither the noise level nor the blur kernel itself (including its type) is given as input to the network. The inpainting model is trained on Bernoulli-masked images at various masking ratios $p$, and the deblurring model is trained on images blurred with kernels drawn from~\cite{levin2009understanding} at various noise levels. Both models map a degraded observation $y$ directly to $x$ without any hyperparameter tuning at test time.

\subsection{Results}

All models are evaluated on the Kodak24 dataset~\cite{franzen1999kodak}. 
For deblurring, we report scores averaged over the full set of motion blur kernels from~\cite{levin2009understanding} at two noise levels ($\sigma = 0$ and $\sigma = 5$). 
For inpainting, we consider two masking ratios ($p = 60\%$ and $p = 80\%$). 
In both cases, we report in Table~\ref{tab:blind_restoration} PSNR~($\uparrow$) and LPIPS~($\downarrow$) as complementary measures of reconstruction fidelity and perceptual quality.

\begin{table*}[h]
    \centering
    \caption{Additional image restoration results with a blind DRUNet. Tests are performed on Kodak24 ~\cite{franzen1999kodak}. Best results are in \textbf{bold}, second are \second{underlined}.}
    \begin{tabular}{rccccgggg}
    \toprule
         \textit{Problem} & \multicolumn{4}{c}{Deblurring} & \multicolumn{4}{g}{Inpainting} \\
         \cmidrule(lr){2-5}\cmidrule(lr){6-9}
         & \multicolumn{2}{c}{$\sigma = 0$} & \multicolumn{2}{c}{$\sigma = 5$} & \multicolumn{2}{g}{$p = 60\%$} & \multicolumn{2}{g}{$p = 80\%$} \\
         \textit{Model} & PSNR & LPIPS & PSNR & LPIPS & PSNR & LPIPS & PSNR & LPIPS \\
         \midrule
         $\text{DRUNet}_{\text{DL}}$  &29.82 &0.0651 &25.35 &0.3204& 29.85 & 0.1022 & 26.73 & 0.2169  \\
         $\text{DRUNet}_{\text{VL}}$  &38.40 &0.0086 &28.25 &0.2874 &32.40 & 0.0565 & 28.77 & 0.1350  \\
         $\text{DRUNet}_{\text{Nat}}$ &39.32 &0.0069 &28.83 &0.2376 &33.39 & 0.0455 & 29.65 & 0.1113  \\
         \bottomrule
    \end{tabular}
    \label{tab:blind_restoration}
\end{table*}

\begin{figure*}[tp]
    \centering
    \begin{tabular*}{\linewidth}{@{\extracolsep{\fill}}ccccc}
        Degraded & Clean & $\text{DRUNet}_{\text{DL}}$ & $\text{DRUNet}_{\text{VL}}$ & $\text{DRUNet}_{\text{Nat}}$ \\
        \midrule
        \Plotimagespy[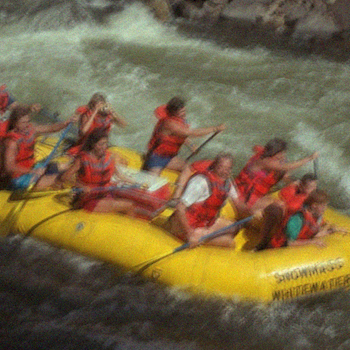]{2.5}{0.5}{0.16\textwidth}{0.4}[3]&
        \Plotimagespy[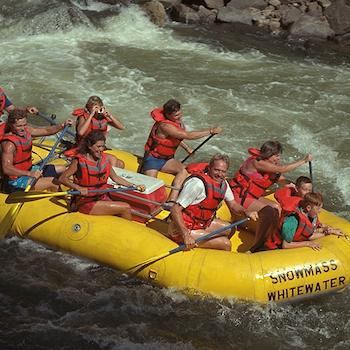]{2.5}{0.5}{0.16\textwidth}{0.4}[3]&
        \Plotimagespy[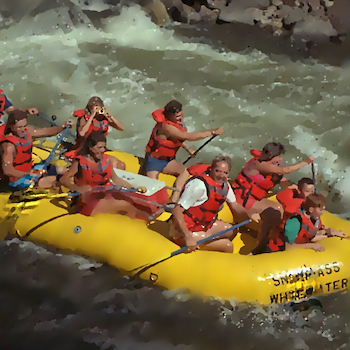]{2.5}{0.5}{0.16\textwidth}{0.4}[3]&
        \Plotimagespy[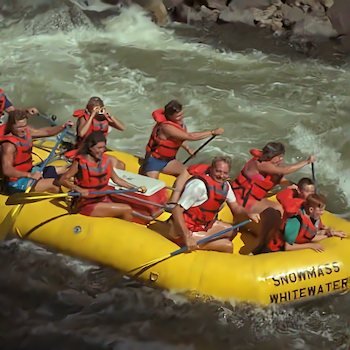]{2.5}{0.5}{0.16\textwidth}{0.4}[3]&
        \Plotimagespy[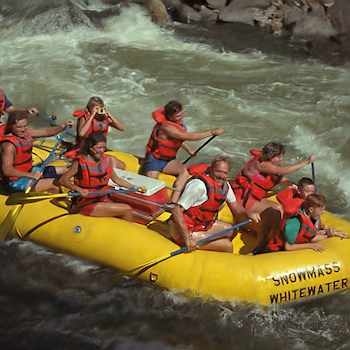]{2.5}{0.5}{0.16\textwidth}{0.4}[3]\\
        &&\small\textbf{Deblurring}&&\\
        \midrule
        \Plotimagespy[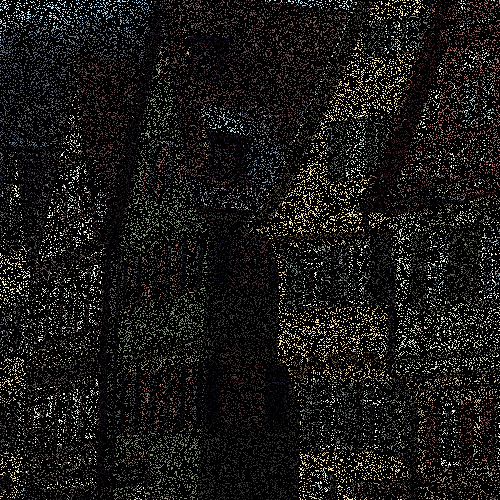]{2}{1.5}{0.16\textwidth}{0.4}[3]&
        \Plotimagespy[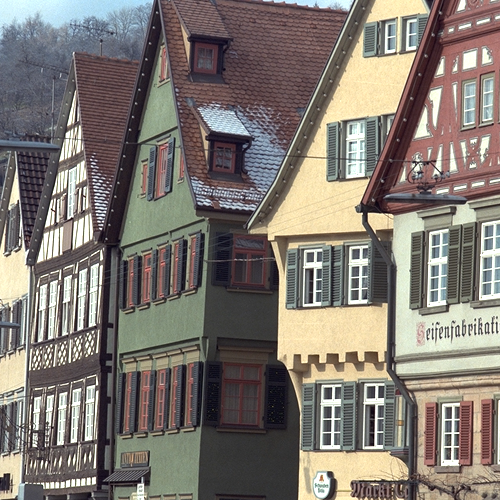]{2}{1.5}{0.16\textwidth}{0.4}[3]&
        \Plotimagespy[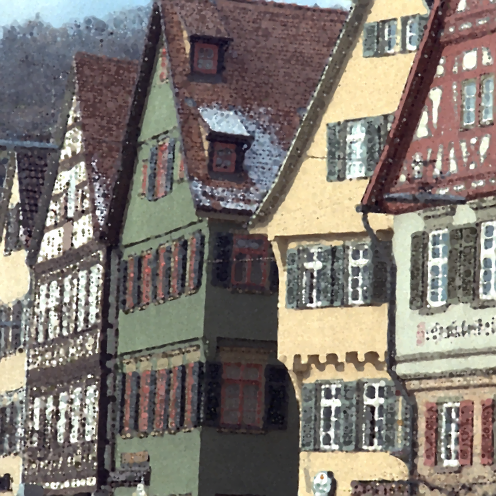]{2}{1.5}{0.16\textwidth}{0.4}[3]&
        \Plotimagespy[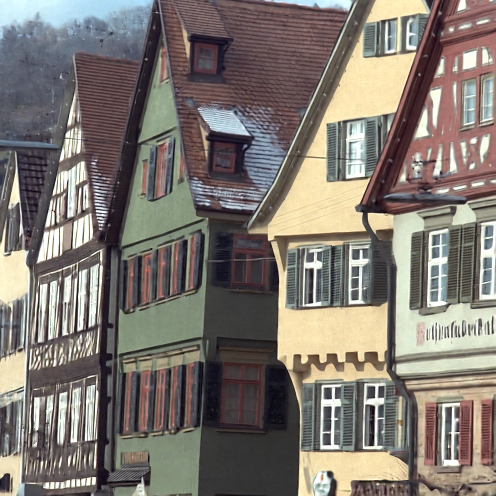]{2}{1.5}{0.16\textwidth}{0.4}[3]&
        \Plotimagespy[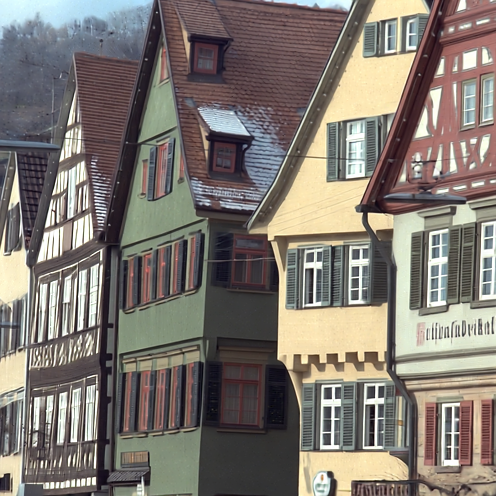]{2}{1.5}{0.16\textwidth}{0.4}[3]\\
        &&\small\textbf{Inpainting}&&\\
        \bottomrule
    \end{tabular*}
    \caption{Blind DRUNet — visual comparison on deblurring (kodim14, kernel $k_0$) and inpainting (kodim08, $p=80\%$). Each row shows the degraded input, the clean reference, and the outputs of the three models trained on Dead Leaves, VibrantLeaves, and natural images respectively.}
    \label{fig:blind_restoration}
\end{figure*}

The numerical results in Table~\ref{tab:blind_restoration} and the visual comparisons in Figure~\ref{fig:blind_restoration} corroborate the results already observed for image denoising and SISR. 
Across both tasks, $\text{DRUNet}_{\text{VL}}$ outperforms $\text{DRUNet}_{\text{DL}}$ by a large margin and approaches $\text{DRUNet}_{\text{Nat}}$ closely. For deblurring, the gap between $\text{DRUNet}_{\text{VL}}$ and $\text{DRUNet}_{\text{Nat}}$ is less than $1$\,dB at both noise levels, while $\text{DRUNet}_{\text{DL}}$ falls behind by more than  $3$\,dB. A similar hierarchy holds for inpainting, with VL again closing most of the gap with the natural-image baseline. Visually, $\text{DRUNet}_{\text{DL}}$ produces over-smoothed outputs that lack fine detail, whereas $\text{DRUNet}_{\text{VL}}$ better preserves textures and edges, consistent with its LPIPS advantage over $\text{DRUNet}_{\text{DL}}$.

\section{Cross-Modality Experiments}\label{sec:supp_cross}

We evaluate the cross-modality generalization of DRUNet models trained on different synthetic datasets by testing them on two out-of-distribution image domains: the FastMRI single-coil knee MRI test set~\cite{zbontar2018fastmri} and the UC Merced Land Use (UCMLD) satellite image dataset~\cite{yang2010bag}. Both datasets lie far outside the natural-image distribution used for training. Results are reported at two noise levels ($\sigma=25$ and $\sigma=50$), and averaged across the two datasets in the last two columns.
\begin{table*}[h]
    \centering

        \setlength{\tabcolsep}{4pt}

\begin{tabular}{lccccccccbbbb}
\toprule
& \multicolumn{4}{c}{UCMLD Satellite}
& \multicolumn{4}{c}{FastMRI Knee (SC)}
& \multicolumn{4}{c}{Average} \\
\cmidrule(lr){2-5}\cmidrule(lr){6-9}\cmidrule(lr){10-13}
& \multicolumn{2}{c}{$\sigma=25$} & \multicolumn{2}{c}{$\sigma=50$}
& \multicolumn{2}{c}{$\sigma=25$} & \multicolumn{2}{c}{$\sigma=50$}
& \multicolumn{2}{c}{$\sigma=25$} & \multicolumn{2}{c}{$\sigma=50$} \\
\cmidrule(lr){2-3}\cmidrule(lr){4-5}\cmidrule(lr){6-7}\cmidrule(lr){8-9}\cmidrule(lr){10-11}\cmidrule(lr){12-13}
Method
& PSNR & {\tiny LPIPS} & PSNR & {\tiny LPIPS}
& PSNR & {\tiny LPIPS} & PSNR & {\tiny LPIPS}
& PSNR & {\tiny LPIPS} & PSNR & {\tiny LPIPS} \\
\midrule
% $\text{DRUNet}_{\text{Fractal}}$ \cite{kataoka2020pre}
% & 16.68 & \tiny 0.5622 & 17.09 & \tiny 0.6279
% & 19.67 & \tiny 0.5506 & 19.52 & \tiny 0.6428
% & \ 18.17 & \tiny \ 0.5564 & \ 18.31 & \tiny \ 0.6354 \\

% $\text{DRUNet}_{\text{ClevR}}$ \cite{johnson2017clevr}
% & 30.08 & \tiny 0.2347 & 27.22 & \tiny 0.3432
% & 34.32 & \tiny 0.4643 & 33.08 & \tiny 0.4943
% & \ 32.20 & \tiny \ 0.3495 & \ 30.15 & \tiny \ 0.4188 \\

$\text{DRUNet}_{\text{DL}}$ \cite{achddou2021synthetic}
& 30.05 & \tiny 0.1954 & 27.35 & \tiny 0.2953
& 34.15 & \tiny 0.4577 & 32.84 & \tiny 0.5000
& \ 32.10 & \tiny \ 0.3266 & \ 30.09 & \tiny \ 0.3977 \\

% $\text{DRUNet}_{\text{DL-Text}}$ \cite{baradad2021learning}
% & 30.65 & \tiny 0.1428 & 27.62 & \tiny 0.2364
% & 34.04 & \tiny 0.4455 & 32.33 & \tiny 0.5324
% & \ 32.34 & \tiny \ 0.2942 & \ 29.98 & \tiny \ 0.3844 \\

% $\text{DRUNet}_{\text{GTAV}}$ \cite{richter2016playing}
% & \second{31.61} & \tiny \second{0.1214} & \second{28.78} & \tiny \second{0.2066}
% & 35.13 & \tiny 0.3224 & \second{33.72} & \tiny \second{0.4198}
% & \ \second{33.37} & \tiny \ 0.2219 & \ \second{31.25} & \tiny \ \second{0.3132} \\

$\text{DRUNet}_{\text{VL}}$
& 31.58 & \tiny 0.1228 & 28.67 & \tiny 0.2138
& \second{35.14} & \tiny \second{0.3016} & 33.70 & \tiny 0.4344
& \ 33.36 & \tiny \ \second{0.2122} & \ 31.19 & \tiny \ 0.3241 \\

\rowcolor{gray!20}
$\text{DRUNet}_{\text{Nat}}$
& \best{32.06} & \tiny \best{0.1025} & \best{29.17} & \tiny \best{0.1871}
& \best{35.43} & \tiny \best{0.2160} & \best{33.96} & \tiny \best{0.3551}
& \cellcolor{blue!30!gray!20}\ \best{33.75} & \cellcolor{blue!30!gray!20}\tiny \ \best{0.1593} & \cellcolor{blue!30!gray!20}\ \best{31.56} & \cellcolor{blue!30!gray!20}\tiny \ \best{0.2711} \\
\bottomrule
\end{tabular}

\caption{Cross-modality denoising results on out-of-distribution datasets. DRUNet models trained on various synthetic and natural image datasets are evaluated on the UCMLD satellite image dataset (210 images) and the FastMRI single-coil knee MRI test set (93 images).Best results are in \best{bold}, second in \second{underlined}.}
\label{tab:cross_modality}
\end{table*}

The results in Table~\ref{tab:cross_modality} confirm that the ranking observed on natural-image benchmarks transfers to out-of-distribution domains. $\text{DRUNet}_{\text{Nat}}$ remains the strongest model overall, but $\text{DRUNet}_{\text{VL}}$ consistently closes most of the gap with it, achieving on average less than $0.4$\,dB below $\text{DRUNet}_{\text{Nat}}$ on both datasets. By contrast, $\text{DRUNet}_{\text{DL}}$ lags behind $\text{DRUNet}_{\text{VL}}$ by roughly $1.3$\,dB on average, confirming that the improvements brought by VibrantLeaves are not specific to the natural-image domain. 
% Notably, $\text{DRUNet}_{\text{VL}}$ and $\text{DRUNet}_{\text{GTAV}}$ perform comparably, with VL holding a slight LPIPS advantage on the FastMRI dataset.

\section{Additional Robustness Experiments}\label{sec:additional_robustness}

We extend the robustness analysis of Section~\ref{sec:invariances} to three further distortion types: Gaussian blur, contrast (S-curve) and hue shift. 

The contrast distortion is an S-curve applied channel-wise, defined by
$$g_{\mathrm{contrast}}(x, k) = \frac{\sigma\!\left(k\,(x - 0.5)\right) - \sigma(-k/2)}{\sigma(k/2) - \sigma(-k/2)},$$
where $\sigma(t)=(1+e^{-t})^{-1}$ is the sigmoid function and $k>0$ controls the steepness of the curve. Larger values of $k$ produce stronger contrast enhancement.

Results are reported in Table~\ref{tab:inv_blur}.
Under Gaussian blur, both $\Dnat$ and $\Dvl$ exhibit a positive $\Delta$, since blurring destroys high frequency details and therefore makes denoising easier; the advantage of $\Dvl$ is small but consistent across all blur levels ($\mathbf{\Delta} \in [+0.03, +0.24]$).

For contrast distortions, the two models degrade almost identically ($|\mathbf{\Delta}| \leq 0.01$), confirming that neither benefits from a specific invariance to S-curve contrast changes. This is expected: the S-curve is a nonlinear, pointwise tone-mapping, not a scalar rescaling of intensity, so it falls outside the 1-homogeneity property of bias-free architectures~\cite{kadkhodaie2021stochastic} discussed in Section~\ref{sec:invariances}, and neither VL nor natural training data confers an advantage here.
Hue shifts produce a similarly mild degradation for both models, with a slight but consistent edge for $\Dvl$ ($\mathbf{\Delta} \in [+0.02, +0.04]$) that does not depend strongly on the rotation angle.
Overall, the gains of $\Dvl$ on these three distortions are more modest than those observed for rotation and scale, which is consistent with the fact that the VibrantLeaves model incorporates explicit rotation and scale invariances but not photometric ones.

\begin{table}[t]
  \centering
  \small
  \setlength{\tabcolsep}{5pt}
  \begin{tabular}{l rrrr}
    \toprule
    \multicolumn{5}{c}{Blur (Gaussian $\sigma$)} \\
    \midrule
    $\beta$ & $0.5$ & $1.625$ & $2.75$ & $3.875$ \\
    \midrule
    $\Delta_{\Dnat}(g,\beta)$ & $+1.36$ & $+6.19$ & $+8.43$ & $+9.89$ \\
    $\Delta_{\Dvl}(g,\beta)$  & $+1.39$ & $+6.36$ & $+8.67$ & $+10.04$ \\
    $\mathbf{\Delta}(g,\beta)$& $+0.03$ & $+0.18$ & $+0.24$ & $+0.15$ \\
    \midrule
    \multicolumn{5}{c}{Contrast (S-curve $k$)} \\
    \midrule
    $\beta$ & $2.5$ & $5$ & $7.5$ & $10$ \\
    \midrule
    $\Delta_{\Dnat}(g,\beta)$ & $-0.15$ & $-0.58$ & $-1.03$ & $-1.40$ \\
    $\Delta_{\Dvl}(g,\beta)$  & $-0.16$ & $-0.58$ & $-1.02$ & $-1.39$ \\
    $\mathbf{\Delta}(g,\beta)$& $-0.01$ & $+0.00$ & $+0.01$ & $+0.01$ \\
    \midrule
    \multicolumn{5}{c}{Hue shift} \\
    \midrule
    $\beta$ & $45\,\mathrm{°}$ & $90\,\mathrm{°}$ & $135\,\mathrm{°}$ & $180\,\mathrm{°}$ \\
    \midrule
    $\Delta_{\Dnat}(g,\beta)$ & $-0.10$ & $-0.13$ & $-0.07$ & $-0.11$ \\
    $\Delta_{\Dvl}(g,\beta)$  & $-0.07$ & $-0.10$ & $-0.04$ & $-0.07$ \\
    $\mathbf{\Delta}(g,\beta)$& $+0.02$ & $+0.03$ & $+0.03$ & $+0.04$ \\
    \bottomrule
  \end{tabular}
  \caption{Invariance study: PSNR differences $\Delta_f(g,\beta)$ (see Section~\ref{sec:invariances}) for $\Dnat$ and $\Dvl$ across three distortion types $g$ at levels $\beta$. $\mathbf{\Delta}(g,\beta) = \Delta_{\Dnat}(g,\beta) - \Delta_{\Dvl}(g,\beta)$; a positive value indicates better robustness of $\Dvl$.}
  \label{tab:inv_blur}\label{tab:inv_contrast}\label{tab:inv_hue}
\end{table}

\begin{figure*}[htp]
    \centering
    \resizebox{0.9\textwidth}{!}{
    \setlength{\tabcolsep}{1pt}
    \begin{tabular}{ccccc}
    Ground Truth & Noisy $\sigma = 200$ & $\text{DRUNet}_{\text{DL}}$ & $\Dvl$ & $\Dnat$ \\
    \Plotimagespy[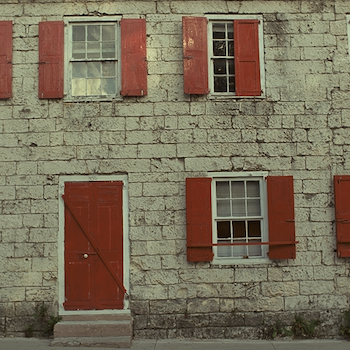]{2.2}{2.2}{0.18\textwidth}{0.4}[4]
    &
    \Plotimagespy[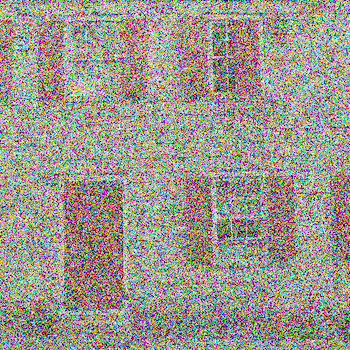]{2.2}{2.2}{0.18\textwidth}{0.4}[4]
    &
    \Plotimagespy[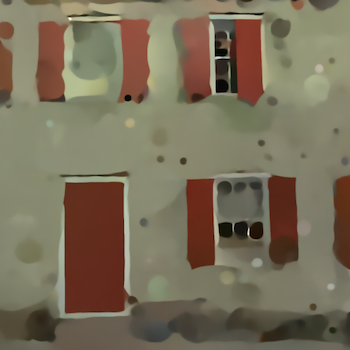]{2.2}{2.2}{0.18\textwidth}{0.4}[4]
    &
    \Plotimagespy[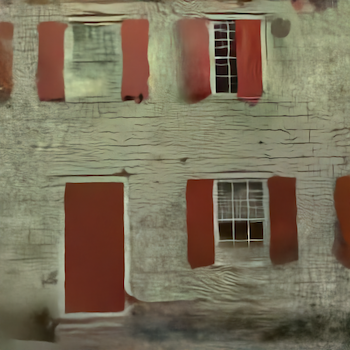]{2.2}{2.2}{0.18\textwidth}{0.4}[4]
    &
    \Plotimagespy[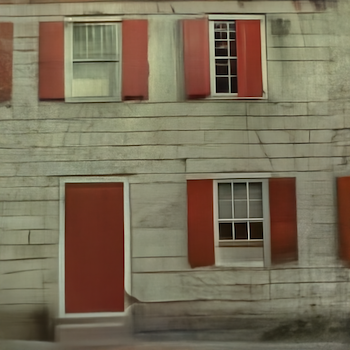]{2.2}{2.2}{0.18\textwidth}{0.4}[4]
    \\
    \Plotimagespy[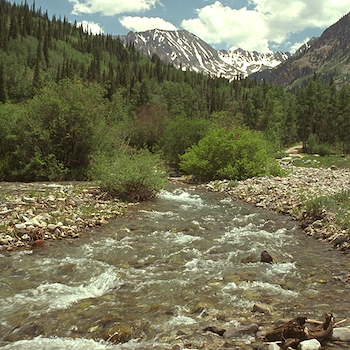]{2.8}{1.5}{0.18\textwidth}{0.4}[4]
    &
    \Plotimagespy[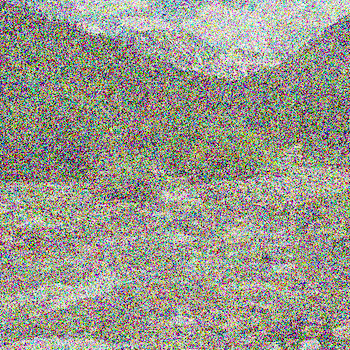]{2.8}{1.5}{0.18\textwidth}{0.4}[4]
    &
    \Plotimagespy[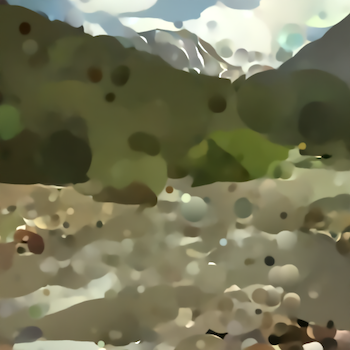]{2.8}{1.5}{0.18\textwidth}{0.4}[4]
    &
    \Plotimagespy[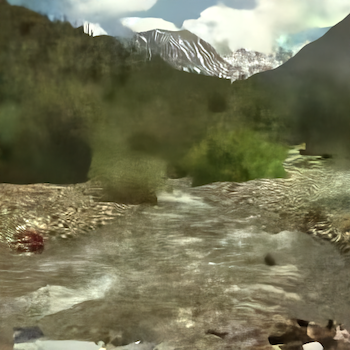]{2.8}{1.5}{0.18\textwidth}{0.4}[4]
    &
    \Plotimagespy[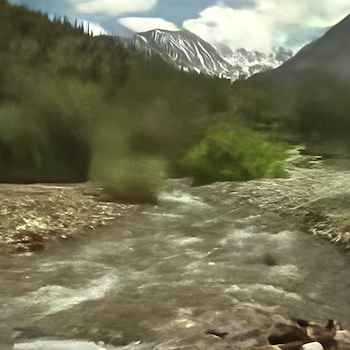]{2.8}{1.5}{0.18\textwidth}{0.4}[4]
    \end{tabular}}
    \caption{Visual comparison of DRUNet denoising at very high noise level ($\sigma=200$, best viewed on a digital screen). We compare the same DRUNet architecture trained on Dead Leaves \cite{achddou2021synthetic} ($\text{DRUNet}_{\text{DL}}$), \DLplus~($\Dvl$), or natural images ($\Dnat$). At this extreme noise level, all models hallucinate details, which explicitly show the biases inherited from the training data.}
    \label{fig:synthetic_bias}
\end{figure*}
\section{Discussion about synthetic bias}\label{sec:synthetic_bias}

As discussed in Section~\ref{sec:denoising}, the choice of training data introduces characteristic biases in the learned image prior.
To make these biases explicit, we probe the three DRUNet variants (DL, VL, Nat) at a noise level of $\sigma = 200$, far outside the training range ($\sigma \in [0, 75]$).
At such extreme degradation, the network can no longer rely on high-frequency cues and must rely entirely on the prior it has learned from its training distribution, causing hallucinations that directly reflect the structure of the training data.

As shown in Figure~\ref{fig:synthetic_bias}, the model trained on Dead Leaves ($\text{DRUNet}_{\text{DL}}$) produces disk-like hallucinations, an artifact directly inherited from the disks of its training images. 
The model trained on \DLplus~($\Dvl$) produces better outputs but struggles to recover regular straight lines. We also notice that the model reproduces the pseudo-periodic textures introduced in VL images. 

Finally, the model trained on natural images ($\Dnat$) yields the most perceptually plausible reconstructions overall, but it hallucinates realistic details that were not present in the original image — most notably, a spurious vertical line in the first example — possibly a consequence of the strong horizontal and vertical bias of the natural image training distribution.

\section{Ablation Study Results}\label{sec:sup_ablation}
\begin{figure*}[tp]
    \centering
    \begin{tabular*}{\linewidth}{@{\extracolsep{\fill}}cccccc}
        Noisy &Clean & Ablation & DRUNet-VL & DRUNet-Nat \\
        \midrule
        
        \Plotimagespy[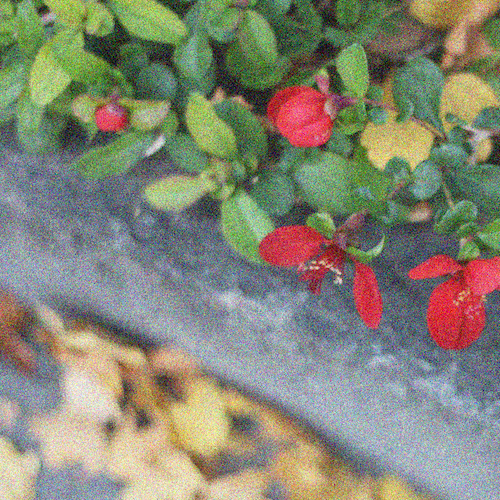]{0.5}{2.3}{0.16\textwidth}{0.4}[3]&
        \Plotimagespy[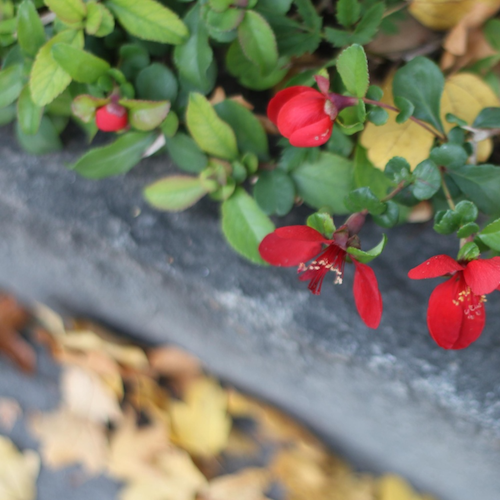]{0.5}{2.3}{0.16\textwidth}{0.4}[3]&
        \Plotimagespy[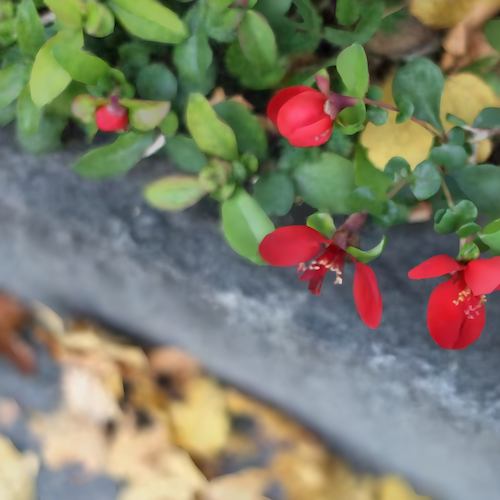]{0.5}{2.3}{0.16\textwidth}{0.4}[3]&
        \Plotimagespy[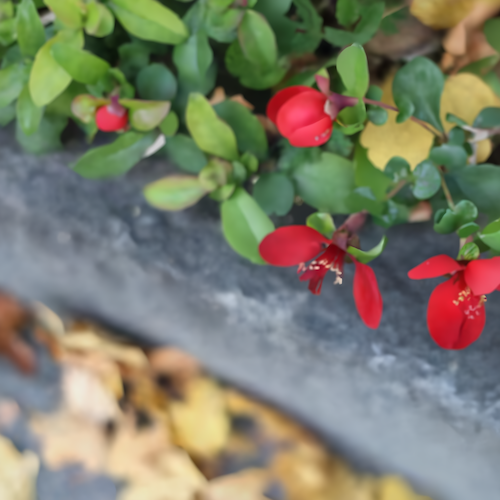]{0.5}{2.3}{0.16\textwidth}{0.4}[3]&
        \Plotimagespy[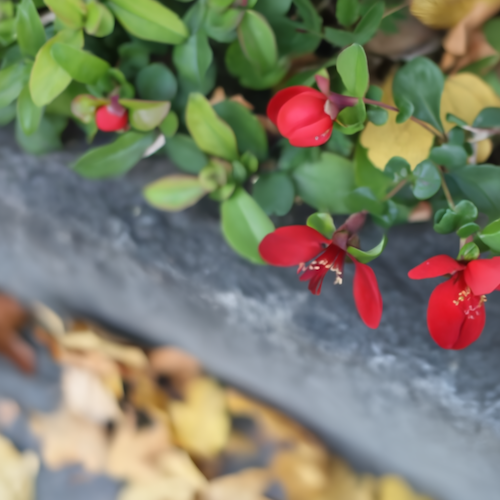]{0.5}{2.3}{0.16\textwidth}{0.4}[3]
        \\
        &&\small \textbf{W/o Depth \Circle{depth}}&& \\
        \midrule
        
        \Plotimagespy[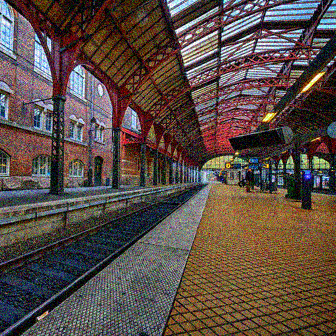]{1.35}{0.6}{0.16\textwidth}{0.4}[3]&
        \Plotimagespy[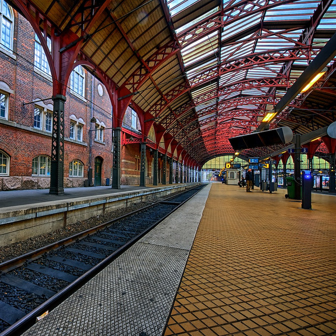]{1.35}{0.6}{0.16\textwidth}{0.4}[3]&
        \Plotimagespy[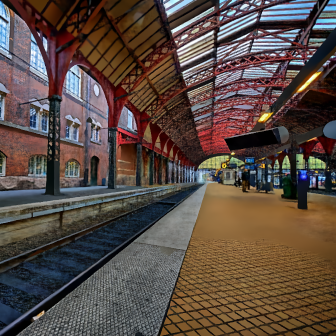]{1.35}{0.6}{0.16\textwidth}{0.4}[3]&
        \Plotimagespy[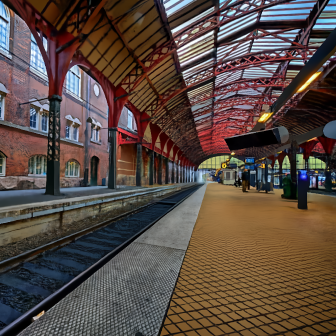]{1.35}{0.6}{0.16\textwidth}{0.4}[3]&
        \Plotimagespy[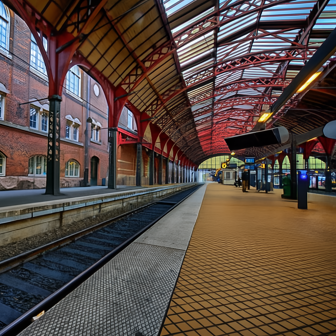]{1.35}{0.6}{0.16\textwidth}{0.4}[3]
        \\
        &&\small \textbf{W/o Periodic-T \Circle{sin}}&& \\
        \midrule
        
        \Plotimagespy[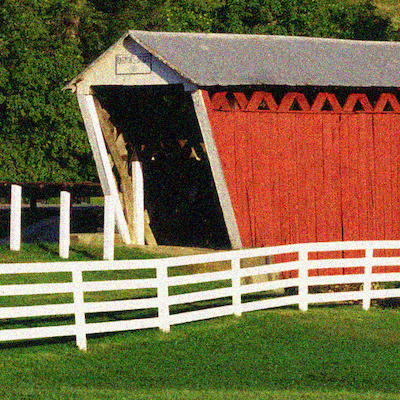]{1.1}{0.3}{0.16\textwidth}{0.4}[4]&
        \Plotimagespy[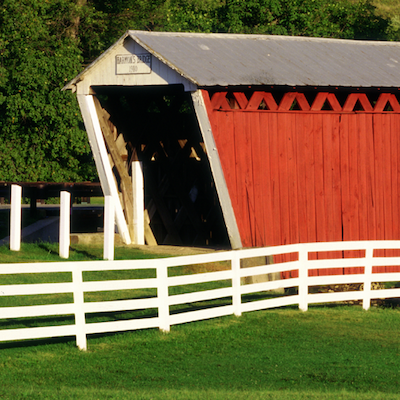]{1.1}{0.3}{0.16\textwidth}{0.4}[4]&
        \Plotimagespy[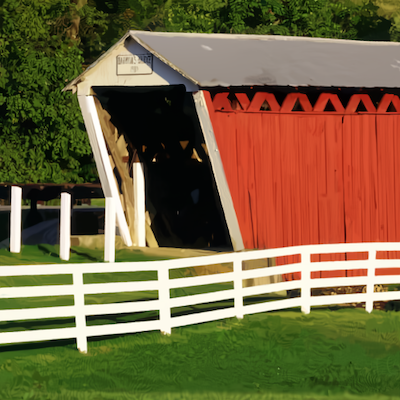]{1.1}{0.3}{0.16\textwidth}{0.4}[4]&
        \Plotimagespy[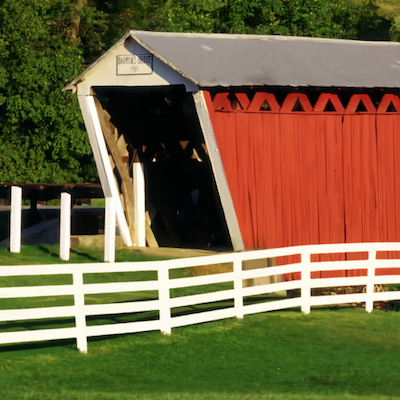]{1.1}{0.3}{0.16\textwidth}{0.4}[4]&
        \Plotimagespy[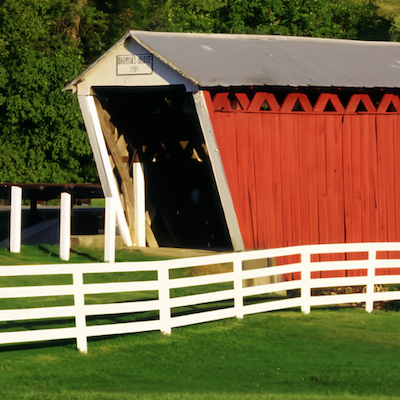]{1.1}{0.3}{0.16\textwidth}{0.4}[4]
        \\
        &&\small \textbf{W/o Micro-T \Circle{microt}}&& \\
        \midrule
        \Plotimagespy[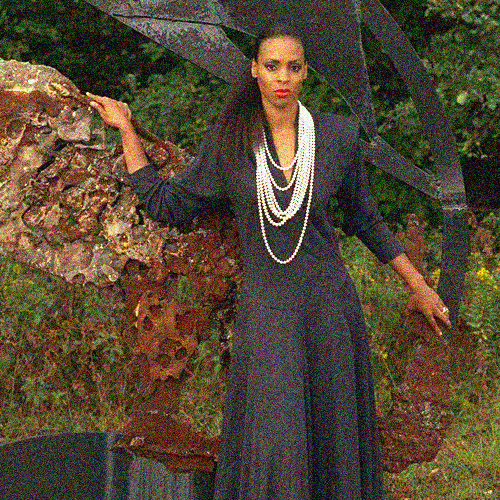]{0.55}{1}{0.16\textwidth}{0.4}[4]&
        \Plotimagespy[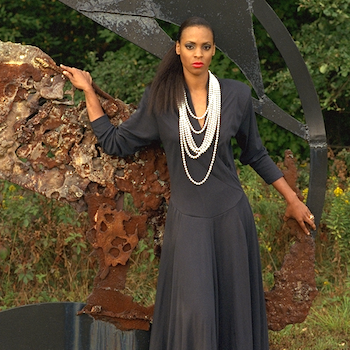]{0.55}{1}{0.16\textwidth}{0.4}[4]&
        \Plotimagespy[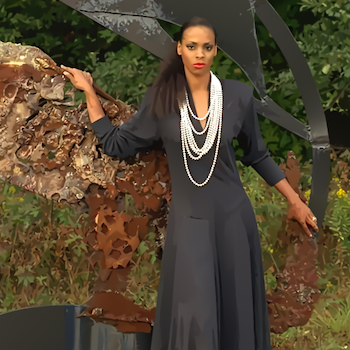]{0.55}{1}{0.16\textwidth}{0.4}[4]&
        \Plotimagespy[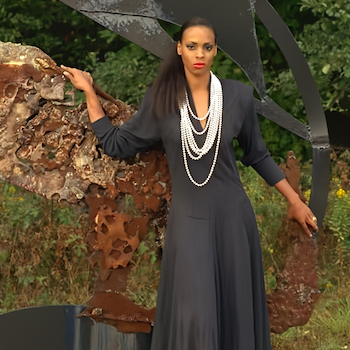]{0.55}{1}{0.16\textwidth}{0.4}[4]&
        \Plotimagespy[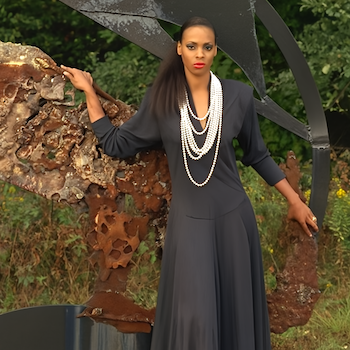]{0.55}{1}{0.16\textwidth}{0.4}[4]\\
        &&\small \textbf{W/o Textures \Circle{text}}&& \\
        \bottomrule
    \end{tabular*}
    \caption{Ablation Study - Visual Comparisons. We compare here the ablated models to the models trained on either VL or Natural Images. Each row correspond to a single ablated model. Visually our version of DRUNet trained on VL images surpasses the ablated models, and is visually close to $\Dnat$.}
    \label{fig:ablation_visual}
\end{figure*}

In Figure~\ref{fig:ablation_visual}, we report some examples of the ablation study results from Section~\ref{sec:ablations}. These visual examples show the importance of the role of texture modeling, depth, and geometry.

\section{Additional sample visualizations of VibrantLeaves images}

\begin{figure*}
    \includegraphics[width = 0.195\textwidth]{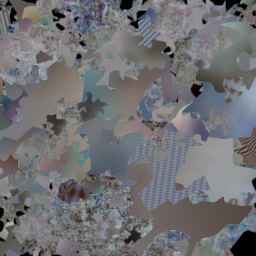}
    \includegraphics[width = 0.195\textwidth]{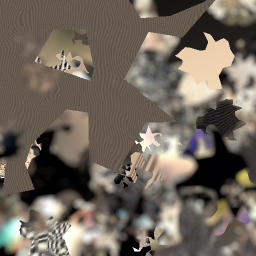}
    \includegraphics[width = 0.195\textwidth]{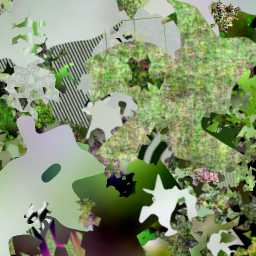}
    \includegraphics[width = 0.195\textwidth]{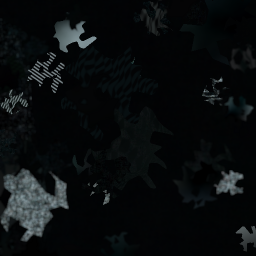}
    \includegraphics[width = 0.195\textwidth]{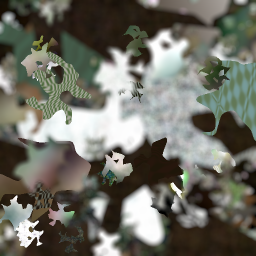}\\
    \includegraphics[width = 0.195\textwidth]{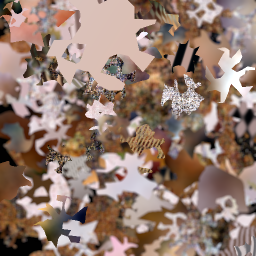}
    \includegraphics[width = 0.195\textwidth]{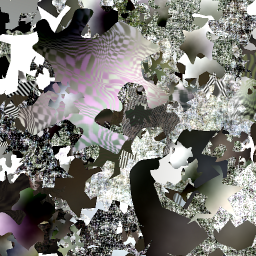}
    \includegraphics[width = 0.195\textwidth]{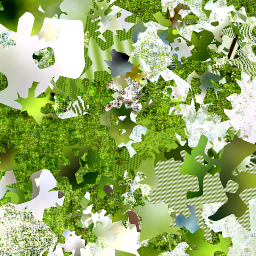}
    \includegraphics[width = 0.195\textwidth]{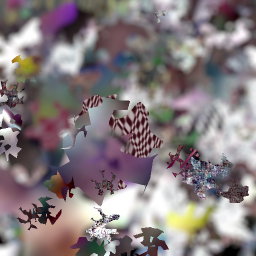}
    \includegraphics[width = 0.195\textwidth]{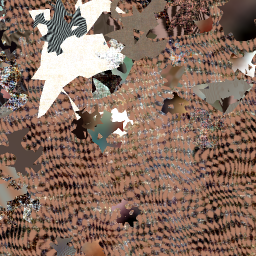}\\
    \includegraphics[width = 0.195\textwidth]{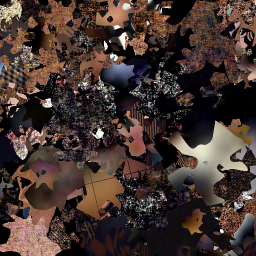}
    \includegraphics[width = 0.195\textwidth]{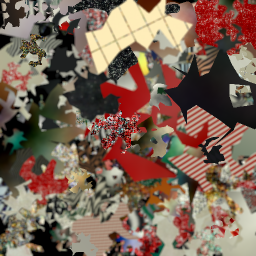}
    \includegraphics[width = 0.195\textwidth]{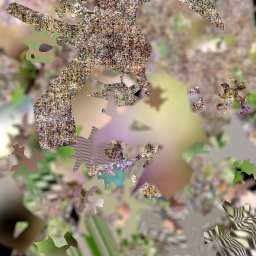}
    \includegraphics[width = 0.195\textwidth]{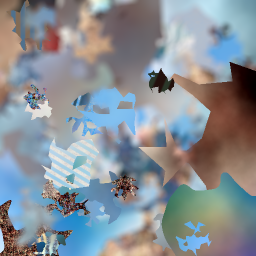}
    \includegraphics[width = 0.195\textwidth]{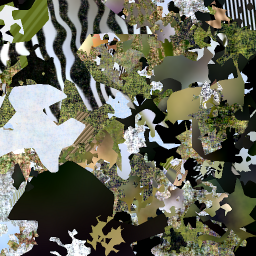}\\
    \includegraphics[width = 0.195\textwidth]{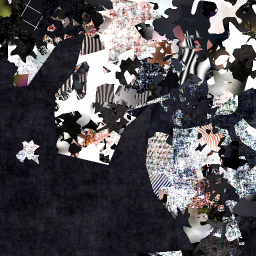}
    \includegraphics[width = 0.195\textwidth]{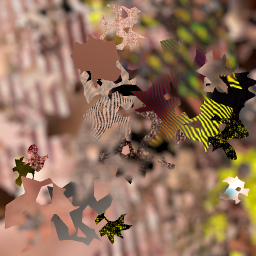}
    \includegraphics[width = 0.195\textwidth]{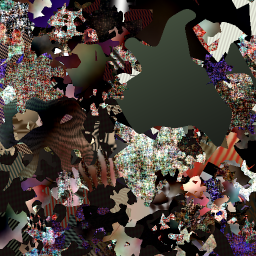}
    \includegraphics[width = 0.195\textwidth]{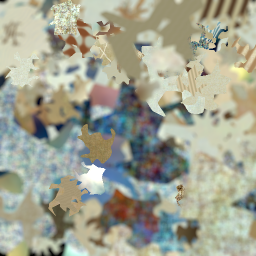}
    \includegraphics[width = 0.195\textwidth]{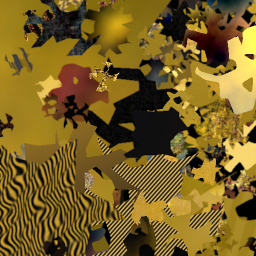}
\caption{Additional samples from the Vibrant Leaves model. The examples display diverse texture patterns, color profiles, object shape distributions, and depth-of field.}
\end{figure*}

% \bibliographystyle{IEEEtran}
% \bibliography{bibliography}
% \end{bibunit}

\end{bibunit}
\end{document}